\documentclass[letterpaper]{article}
\usepackage{uai2018}
\usepackage[margin=1in]{geometry}

\usepackage{times}  
\usepackage{helvet}  
\usepackage{courier}  
\usepackage{url}  

\usepackage[authoryear,round,longnamesfirst]{natbib}

\usepackage[utf8]{inputenc} 
\usepackage{hyperref}       
\usepackage{url}            
\usepackage{booktabs}       
\usepackage{amsfonts}       
\usepackage{nicefrac}       
\usepackage{microtype}      
\usepackage{adjustbox}

\usepackage{xcolor}

\usepackage{amsmath}

\usepackage{graphicx} 
\usepackage{subfigure} 

\graphicspath{{figures/}}

\usepackage{wrapfig}

\usepackage{algorithm}
\usepackage{algorithmic}

\usepackage{sectsty}
\usepackage{titlecaps}

\ifx\proof\undefined
\newenvironment{proof}{\par\noindent{\bf Proof\ }}{\hfill\BlackBox\\[2mm]}
\fi

\ifx\theorem\undefined
\newtheorem{theorem}{Theorem}
\fi
\ifx\example\undefined
\newtheorem{example}{Example}
\fi
\ifx\lemma\undefined
\newtheorem{lemma}[theorem]{Lemma}
\fi

\ifx\corollary\undefined
\newtheorem{corollary}[theorem]{Corollary}
\fi

\ifx\assumption\undefined
\newtheorem{assumption}{Assumption}
\fi

\ifx\definition\undefined
\newtheorem{definition}{Definition}
\fi

\ifx\proposition\undefined
\newtheorem{proposition}[theorem]{Proposition}
\fi

\ifx\remark\undefined
\newtheorem{remark}{Remark}
\fi

\ifx\conjecture\undefined
\newtheorem{conjecture}[theorem]{Conjecture}
\fi

\ifx\factoid\undefined
\newtheorem{factoid}[theorem]{Fact}
\fi

\ifx\axiom\undefined
\newtheorem{axiom}[theorem]{Axiom}
\fi

\newcommand{\RN}[1]{%
	\textup{\lowercase\expandafter{\it \romannumeral#1}}%
}

\def\ReLU{\textsf{ReLU}} 
\def\KL{\textsf{KL}} 

\input{Definitions}

\title{A Unified Particle-Optimization Framework for Scalable Bayesian Sampling}

\author{} 

\author{ {\bf Changyou Chen} \\
University at Buffalo (\href{mailto:cchangyou@gmail.com}{cchangyou@gmail.com}) \\
\And
Ruiyi Zhang, Wenlin Wang, Bai Li, Liqun Chen  \\
Duke University\\
}

\begin{document}
	
	\maketitle
\sectionfont{\MakeUppercase}

\begin{abstract}
	There has been recent interest in developing scalable Bayesian sampling methods such as stochastic gradient MCMC (SG-MCMC) and Stein variational gradient descent (SVGD) for big-data analysis. A standard SG-MCMC algorithm simulates samples from a discrete-time Markov chain to approximate a target distribution, thus samples could be highly correlated, an undesired property for SG-MCMC. In contrary, SVGD directly optimizes a set of particles to approximate a target distribution, and thus is able to obtain good approximations with relatively much fewer samples. In this paper, we propose a principle particle-optimization framework based on Wasserstein gradient flows to unify SG-MCMC and SVGD, and to allow new algorithms to be developed. Our framework interprets SG-MCMC as particle optimization on the space of probability measures, revealing a strong connection between SG-MCMC and SVGD. The key component of our framework is several particle-approximate techniques to efficiently solve the original partial differential equations on the space of probability measures. Extensive experiments on both synthetic data and deep neural networks demonstrate the effectiveness and efficiency of our framework for scalable Bayesian sampling.
\end{abstract}
\vspace{-0.7cm}
\section{Introduction}
\vspace{-0.4cm}
Bayesian methods have been playing an important role in modern machine learning, especially in unsupervised learning \citep{KingmaW:ICLR14,LiLCPCHC:NIPS17}, and recently in deep reinforcement learning \citep{houthooft2016vime,liu2017stein}. When dealing with big data, two lines of research directions have been developed to scale up Bayesian methods, {\it e.g.}, variational-Bayes-based and sampling-based methods. Stochastic gradient Markov chain Monte Carlo (SG-MCMC) is a family of scalable Bayesian learning algorithms designed to efficiently sample from a target distribution such as a posterior distribution \citep{WellingT:ICML11,ChenFG:ICML14,DingFBCSN:NIPS14,ChenDC:NIPS15}. In principle, SG-MCMC generates samples from a Markov chain, which are used to approximate a target distribution. Under a standard setting, samples from SG-MCMC are able to match a target distribution exactly with an infinite number of samples \citep{TehTV:arxiv14,ChenDC:NIPS15}. However, this is practically infeasible, as only a finite number of samples are obtained. Although nonasymptotic approximation bounds w.r.t.\! the number of samples have been investigated \citep{TehTV:arxiv14,VollmerZT:arxiv15,ChenDC:NIPS15}, there are no theory/algorithms to guide learning an optimal set of fixed-size samples/particles. This is an undesirable property of SG-MCMC, because in practice one often seeks to learn the optimal samples of a finite size that best approximate a target distribution.

A remedy for this issue is to adopt the idea of particle-based sampling methods, where a set of particles (or samples) are initialized from some simple distribution, followed by iterative updates to better approximate a target distribution. The updating procedure is usually done by optimizing some metrics such as a distance measure between the target distribution and the current approximation. There is not much work in this direction for large-scale Bayesian sampling, with an outstanding representative being the Stein variational gradient descent (SVGD) \citep{LiuW:NIPS16}. In SVGD, the update of particles are done by optimizing the KL-divergence between the empirical particle distribution and a target distribution, thus the samples are designed to be updated optimally to reduce the KL-divergence in each iteration. Because of this property, SVGD is found to perform better than SG-MCMC when the number of samples used to approximate a target distribution is limited, and has been applied to other problems such as deep generative models \citep{FengWL:UAI17} and deep reinforcement learning \citep{liu2017stein,HaarnojaTAL:ICML17,ZhangLCC:AISTATS18}.

Though often achieving comparable performance in practice, little work has been done on investigating connections between SG-MCMC and SVGD, and on developing particle-optimization schemes for SG-MCMC. In this paper, adopting ideas from Waserstein-gradient-flow literature, we propose a unified particle-optimization framework for scalable Bayesian sampling. The idea of our framework is to work directly on the evolution of a density functions on the space of probability measures, {\it e.g.}, the Fokker-Planck equation in SG-MCMC. To make the evolution solution computationally feasible, particle approximations are adopted for densities, where particles can be optimized during the evolution process. Both SG-MCMC and SVGD are special cases of our framework, and are shown to be highly related. Notably, sampling with SG-MCMC becomes a deterministic particle-optimization problem as SVGD on the space of probability measures, overcoming the aforementioned correlated-sample issue. Furthermore, we are able to develop new {\em unified} particle-optimization algorithms by combing SG-MCMC and SVGD, which is less prone to high-dimension space and thus obtains better performance for large-scale Bayesian sampling. We conduct extensive experiments on both synthetic data and Bayesian learning of deep neural networks, verifying the effectiveness and efficiency of our proposed framework.

\section{Preliminaries}
\vspace{-0.3cm}
In this section, we review related concepts and algorithms for SG-MCMC, SVGD, and Wasserstein gradient flows (WGF) on the space of probability measures.
\vspace{-0.3cm}
\subsection{Stochastic gradient MCMC}
\vspace{-0.1cm}
\paragraph{Diffusion-based sampling methods}
Generating random samples from a distribution ({\it e.g.}, a posterior distribution) is one of the fundamental problems in Bayesian statistics, which has many important applications in machine learning. Traditional Markov Chain Monte Carlo methods (MCMC), such as the Metropolis–Hastings algorithm \citep{Metropolis:53} produces unbiased samples from a desired distribution when the density function is known up to a normalizing constant. However, most of these methods are based on random walk proposals which suffer from high dimensionality and often lead to highly correlated samples. On the other hand, dynamics-based sampling methods such as the Metropolis adjusted Langevin algorithm (MALA) \citep{XifaraSLBG:SPL14} avoid this high degree of correlation by combining dynamical systems with the Metropolis step. In fact, these dynamical systems are derived from a more general mathematical technique called diffusion process, or more specifically, It\'{o} diffusion \citep{Oksendal:85}.

Specifically, our objective is to generate random samples from a posterior distribution $p(\thetab|\Xb)\propto p(\Xb|\thetab)p(\thetab)$, where $\thetab\in\mathbb{R}^r$ represents the model parameter, and $\Xb\triangleq \{\xb_i\}_{i=1}^N$ represents the data. The canonical form is $p(\thetab|\Xb)=(1/Z)\exp(U(\thetab))$, where $U(\thetab)=$
\vspace{-0.2cm}
{\small\begin{align*}
\log p(\Xb|\thetab)+\log p(\thetab)
\triangleq \sum_{i=1}^N\log p(\xb_i | \thetab) + \log p(\thetab)
\end{align*}}
is referred to as the potential energy based on an i.i.d.\! assumption of the model, and $Z$ is the normalizing constant. 
In Bayesian sampling, the posterior distribution corresponds to the (marginal) stationary distribution of a (continuous-time) It\'{o} diffusion, defined as a stochastic differential equation of the form:
\begin{align}\label{eq:itodif}
\mathrm{d}\Thetab_t = F(\Thetab_t)\mathrm{d}t + g(\Thetab_t)\mathrm{d}\mathcal{W}_t~,
\end{align}
where $t$ is the time index; $\Thetab_t \in \mathbb{R}^{\pr}$ represents the full variables in a dynamical system, and
$\Thetab_t \supseteq \thetab_t$ (thus $\pr \geq r$) is potentially an augmentation of model parameter $\thetab$; $\mathcal{W}_t \in \mathbb{R}^{\pr}$ is $\pr$-dimensional Brownian motion. Functions $F: \mathbb{R}^{\pr} \to \mathbb{R}^{\pr}$ and $g: \mathbb{R}^{\pr} \rightarrow \mathbb{R}^{\pr\times \pr}$ are assumed to satisfy the Lipschitz continuity condition \citep{Ghosh:book11}. By Fokker-Planck equation (or the forward Kolmogorov equation) \citep{Kolmogoroff:MA31,Risken:FPE89}, when appropriately designing the diffusion-coefficient functions $F(\cdot)$ and $g(\cdot)$, the stationary distribution of the corresponding It\'{o} diffusion equals the posterior distribution of interest, $p(\thetab|\Xb)$. For example, the 1st-order Langevin dynamic defines $\Thetab = \thetab$, and $F(\Thetab_t) = \frac{1}{2}\nabla_{\thetab} U(\thetab),~~ g(\Thetab_t) = \Ib_r$; the 2nd-order Langevin diffusion defines $\Thetab = (\thetab, \qb)$, and 
$F(\Thetab_t)= \Big( \begin{array}{c}
\qb \\
-B \qb-\nabla_\thetab U(\thetab) \end{array} \Big),\hspace{0.1cm}
g(\Thetab_t) = \sqrt{2B}\Big( \begin{array}{cc}
{\bf 0} & {\bf 0} \\
{\bf 0} & \Ib_r \end{array} \Big)$
for a scalar $B > 0$; $\qb$ is an auxiliary variable known as the momentum \citep{ChenFG:ICML14,DingFBCSN:NIPS14}.

Let the density of $\Thetab_t$ be $\mu_t$, it is known $\mu_t$ is characterized by the Fokker-Planck (FP) equation \citep{Risken:FPE89}:
\begin{align}\label{eq:FPE}
\frac{\partial \mu_t}{\partial t} = -\nabla_{\Thetab}\cdot \left(\mu_tF(\Thetab_t)\right) + \nabla_{\Thetab}\nabla_{\Thetab}\!:\!\left(\mu_t\Sigma(\Thetab_t)\right)
\end{align}
where $\Sigma(\Thetab_t) \triangleq g(\Thetab_t)g^{\top}(\Thetab_t)$, $\ab\cdot\bb \triangleq \ab^{\top} \bb$ for vectors $\ab$ and $\bb$, $\Ab\!:\!\Bb\triangleq \mbox{trace}(\Ab^{\top}\Bb)$ for matrices $\Ab$ and $\Bb$. The FP equation is the key to develop our particle-optimization framework for SG-MCMC. In the following, we focus on the simplest case of 1st-order Langevin dynamics if not stated explicitly, though the derivations apply to other variants.
\vspace{-0.3cm}
\paragraph{Stochastic gradient MCMC}
SG-MCMC algorithms are discretized numerical approximations of the It\'{o} diffusion \eqref{eq:itodif}. They mitigate the slow mixing and non-scalability issues encountered in traditional MCMC algorithms by $\RN{1})$ adopting gradient information of the posterior distribution, $\RN{2})$ using minibatches of the data in each iteration of the algorithm to generate samples, and $\RN{3})$ ignoring the rejection step as in standard MCMC. To make the algorithms scalable in a big-data setting, three developments will be implemented based on the It\'{o} diffusion: $\RN{1})$ define appropriate functions $F$ and $g$ in the It\'{o}-diffusion formula so that the (marginal) stationary distributions coincide with the target posterior distribution $p(\thetab|\Xb)$; $\RN{2})$ replace $F$ or $g$ with unbiased stochastic approximations to reduce the computational complexity, {\it e.g.}, approximating $F$ with a random subset of the data instead of using the full data. For example, in the 1st-order Langevin dynamics, $\nabla_{\thetab}U(\thetab)$ could be {\em approximated} by an unbiased estimator with a subset of data:
\vspace{-0.2cm}
\begin{align}\label{eq:U}
\nabla_{\thetab} \tilde{U}(\thetab) \triangleq \nabla\log p(\thetab)+ \frac{N}{n}\sum_{i=1}^n \nabla_{\thetab}\log p(\xb_{\pi_i}|\thetab)
\end{align}
where $\pi$ is a size-$n$ random subset of $\{1, 2, \cdots, N\}$, leading to the first SG-MCMC algorithm in machine learning -- stochastic gradient Langevin dynamics (SGLD) \citep{WellingT:ICML11}; and $\RN{3})$ solve the generally intractable continuous-time It\^{o} diffusions with a numerical method, {\it e.g.}, the Euler method \citep{ChenDC:NIPS15}. For example, this leads to the following update in SGLD:
\begin{align*}
\thetab_{\ell} = \thetab_{\ell-1} + \nabla_{\thetab}\tilde{U}(\thetab_{\ell-1})h + \sqrt{2h}\deltab_{\ell}~,
\end{align*} 
where $h$ means the stepsize, $\ell$ indexes the samples, $\deltab_{\ell}\sim\mathcal{N}(\mathbf{0}, \Ib)$ is a random sample from an isotropic normal distribution. After running the algorithm for $L$ steps, the collection of samples $\{\thetab_{\ell}\}_{\ell=1}^L$ are used to approximate the unknown posterior distribution $\frac{1}{Z}e^{U(\thetab)}$.
\vspace{-0.1cm}
\subsection{Stein variational gradient descent}
\vspace{-0.2cm}
Different from SG-MCMC, SVGD initializes a set of particles which are iteratively updated so that the empirical particle distribution approximates the posterior distribution. Specifically, we consider a set of particles $\{\thetab^{(i)}\}_{i=1}^M$ drawn from some distribution $q$. SVGD tries to update these particles by doing gradient descent on the interactive particle system via
\begin{align*}
\thetab^{(i)} \leftarrow \thetab^{(i)} + h \phi(\thetab^{(i)}),~~\phi = \arg\max_{\phi\in \mathcal{F}} \{\dfrac{\partial}{\partial h} \KL(q_{[h\phi]}||p)\}
\end{align*}
where $\phi$ is a function perturbation direction chosen to minimize the KL divergence between the updated density $q_{[h\phi]}$ estimated by the particles and the posterior $p(\thetab|\Xb)$ ($p$ for short). Since $\KL(q\|p)$ is convex in $q$, global optimum of $q = p$ can be guaranteed.
SVGD considers $\mathcal{F}$ as the unit ball of a vector-valued reproducing kernel Hilbert space (RKHS) $\mathcal{H}$ associated with a kernel $\kappa(\thetab,\thetab^{\prime})$. In such as setting, \cite{liu2016stein} shown:
\begin{align}\label{eq:ksd}
-\frac{\partial}{\partial h} &\KL(q_{[h \phi]}\|p)|_{h=0} = \mathbb{E}_{\thetab \sim q}[\text{trace}(\Gamma_p \phi(\thetab))],\\
&\text{with}
~~\Gamma_p \phi(\thetab) \triangleq \nabla_{\thetab} U( \thetab) ^{\top} \phi(\thetab) + \nabla_{\thetab} \cdot \phi(\thetab), \nonumber
\end{align}
where $\Gamma_p$ is called the Stein operator. 
Assuming that the update function $\phi(\thetab)$ is in a RKHS with kernel $\kappa(\cdot,\cdot)$, it was shown in \citep{liu2016stein} that (\ref{eq:ksd}) is maximized with:
\vspace{-0.2cm}
\begin{align}
\label{equ:close}
\phi(\thetab) = \mathbb{E}_{\thetab\sim q}[\kappa(\thetab, \thetab^\prime) \nabla_{\thetab} U(\thetab)
+ \nabla_{\thetab} \kappa(\thetab, \thetab^\prime)].
\end{align}
When approximating the expectation $\mathbb{E}_{\thetab\sim q}[\cdot]$ with empirical particle distribution and adopting stochastic gradients, we arrive at the following updates for the particles ($\ell$ denotes the iteration number): $\thetab_{\ell+1}^{(i)} = \thetab_{\ell}^{(i)} +$
\vspace{-0.2cm}
{\small\begin{align}  \label{eq:svgd_update}
	\hspace{-0.3cm}\dfrac{h}{M} \sum_{j=1}^M \left[ \kappa(\thetab_{\ell}^{(j)}, \thetab_{\ell}^{(i)}) \nabla_{\thetab_{\ell}^{(j)}} \tilde{U}( \thetab_{\ell}^{(j)}) + \nabla_{\thetab_{\ell}^{(j)}} \kappa(\thetab_{\ell}^{(j)}, \thetab_{\ell}^{(i)}) \right]
	\end{align}}
SVGD applies updates \eqref{eq:svgd_update} repeatedly, moving the samples to a target distribution $p$.

\vspace{-0.1cm}
\subsection{Wasserstein Gradient Flows}
\vspace{-0.3cm}
For a better motivation of WGF, we start from gradient flows defined on the Euclidean space.
\vspace{-0.3cm}
\paragraph{Gradient flows on the Euclidean space}\label{sec:gf_euc}
For a smooth function $E: \mathbb{R}^r \rightarrow \mathbb{R}$, and a starting point $\thetab_0 \in \mathbb{R}^r$, the gradient flow of $E(\thetab)$ is defined as the solution of the differential equation: $\frac{\mathrm{d}\thetab}{\mathrm{d}t} = -\nabla E(\thetab(t))$, s.t.\! $\thetab(0) = \thetab_0$. This is a standard Cauchy problem \citep{Rulla:NA96}, endowed with a unique solution if $\nabla E$ is Lipschitz continuous. When $E$ is non-differentiable, the gradient is replaced with its subgradient, defined as $\partial E(\thetab) \triangleq \{\pb \in \mathbb{R}^r: F(\thetab^\prime) \geq F(\thetab) + \pb \cdot (\thetab^\prime - \thetab), \forall \thetab^\prime \in \mathbb{R}^r\}$. Note $\partial E(\thetab) = \{\nabla E(\thetab)\}$ if $E$ is differentiable at $\thetab$. In this case, the gradient flow formula above is replaced with: $\frac{\mathrm{d}\thetab}{\mathrm{d}t} \in -\partial E(\thetab(t))$.
\vspace{-0.4cm}
\paragraph{Wasserstein gradient flows}
Let $\mathcal{P}(\Omega)$ denote the space of probability measures on $\Omega \subset \mathbb{R}^r$. WGF is an extension of gradient flows in Euclidean space by lifting the definition onto the space of probability measures. Formally, let $\mathcal{P}(\Omega)$ be endowed with a Riemannian geometry induced by the 2nd-order Wasserstein distance, {\it i.e.}, the curve length between two elements (two distributions) is defined as:
{\small\begin{align*}
	W_2^2(\mu, \nu) \triangleq \inf_{\gamma}\left\{\int_{\Omega \times \Omega}\|\thetab - \thetab^\prime\|_2^2\mathrm{d}\gamma(\thetab, \thetab^\prime): \gamma \in \Gamma(\mu, \nu)\right\}
	\end{align*}}
where $\Gamma(\mu, \nu)$ is the set of joint distributions over $(\thetab, \thetab^\prime)$ such that the two marginals equal $\mu$ and $\nu$, respectively. The Wasserstein distance can be explained as an optimal-transport problem, where one wants to transform elements in the domain of $\mu$ to $\nu$ with a minimum cost \citep{Villani:08}. The term $\|\thetab - \thetab^\prime\|_2^2$ represents the cost to transport $\thetab$ in $\mu$ to $\thetab^\prime$ in $\nu$, and can be replaced by a general metric $c(\thetab, \thetab^\prime)$ in a metric space. If $\mu$ is absolutely continuous w.r.t.\! the Lebesgue measure, there is a unique optimal transport plan from $\mu$ to $\nu$, {\it i.e.}, a mapping $\mathcal{T}: \mathbb{R}^r \rightarrow\mathbb{R}^r$ pushing elements in the domain of $\mu$ onto $\nu$ satisfying $\mathcal{T}_{\#}\mu = \nu$. Here $\mathcal{T}_{\#}\mu$ denotes the pushforward measure of $\mu$. The Wasserstein distance is equivalently reformulated as: 
$W_2^2(\mu, \nu) \triangleq \inf_{\mathcal{T}}\left\{\int_{\Omega}c(\thetab, \mathcal{T}(\thetab))\mathrm{d}\mu(\thetab)\right\}$. 

Let $\{\mu_{t}\}_{t\in[0,1]}$ be an absolutely continuous curve in $\mathcal{P}(\Omega)$ with finite second-order moments. We consider to define the change of $\mu_t$'s by investigating $W_2^2(\mu_{t}, \mu_{t+h})$. Motivated by the Euclidean-space case, this is reflected by a vector field, $\vb_{t}(\thetab) \triangleq \lim_{h\rightarrow 0}\frac{\mathcal{T}(\thetab_{t}) - \thetab_{t}}{h}$ called the {\em velocity of the particle}. A gradient flow can be defined on $\mathcal{P}(\Omega)$ correspondingly \citep{Ambrosio:book05}.
\vspace{-0.2cm}
\begin{lemma}\label{theo:gf_w_exist}
	Let $\{\mu_{t}\}_{t\in[0,1]}$ be an absolutely-continuous curve in $\mathcal{P}(\Omega)$ with finite second-order 
	moments. Then for a.e.\! $t\in [0, 1]$, the above vector field $\vb_{t}$ defines a gradient flow on $\mathcal{P}(\Omega)$ as: $\partial_{t} \mu_{t} + \nabla \cdot (\vb_{t} \mu_{t}) = 0$.
\end{lemma}
The gradient flow describes the evolution of a functional $E$, which is a lifted version of the function in the case of Euclidean space in Section~\ref{sec:gf_euc} to the space of probability measures. $E$ maps a probability measure $\mu$ to a real value, {\it i.e.}, $E: \mathcal{P}(\Omega) \rightarrow \mathbb{R}$. We will focus on the case where $E$ is convex in this paper, which is enough considering gradient flows for SG-MCMC and SVGD, though the theory applies to a more general $\lambda$-convex energy functional setting \citep{Ambrosio:book05}. It can be shown that $\vb_{t}$ in Lemma~\ref{theo:gf_w_exist} has the form $\vb_{t} = -\nabla \frac{\delta E}{\delta \mu_{t}}(\mu_{t})$ \citep{Ambrosio:book05}, where $\frac{\delta E}{\delta \mu_{t}}$ is called the {\em first variation} of $E$ at $\mu_{t}$. Based on this, gradient flows on $\mathcal{P}(\Omega)$ can be written
\vspace{-0.0cm}
\begin{align}\label{eq:gf_dis1}
\partial_{t} \mu_{t} = -\nabla \cdot (\vb_{t} \mu_{t}) = \nabla \cdot \left(\mu_{t} \nabla(\frac{\delta E}{\delta \mu_{t}}(\mu_{t}))\right)~.
\end{align}
\vspace{-0.4cm}
\begin{remark}
	Intuitively, an energy functional $E$ characterizes the landscape structure (appearance) of the corresponding manifold in $\mathcal{P}(\Omega)$, and the gradient flow \eqref{eq:gf_dis1} defines a geodesic path on this manifold. Usually, by choosing appropriate $E$, the landscape is convex, {\it e.g.}, for the cases of both SG-MCMC and SVGD described below. This provides a theoretical guarantee on the optimal convergence of a gradient flow.
\end{remark}
\vspace{-0.3cm}
\section{Particle-Optimization-based Sampling}
\vspace{-0.3cm}
In this section, we interpret the continuous versions of both SG-MCMC and SVGD as WGFs, followed by several techniques for particle optimization in the next section. In the following, $\mu_t$ denotes the distribution of $\thetab_{t}$.

\subsection{SVGD as WGF}\label{sec:svgd_wgf}

The continuous-time and infinite-particle limit of SVGD with full gradients, denoted as SVGD$^\infty$, is known to be a special instance of the Vlasov equation in nonlinear partial-differential-equation literature \citep{liu2017stein_flow}:
\vspace{-0.1cm}
\begin{align}\label{eq:vlasov}
\partial_{t} \mu_{t} = \nabla\cdot \left((\Wb*\mu_{t})\mu_{t}\right)~,
\end{align}
where $(\Wb*\mu_{t})(\thetab) \triangleq \int \Wb(\thetab - \thetab^\prime)\mu_{t}(\thetab^\prime)\mathrm{d}\thetab^\prime$ is the convolutional operator applied for some function $\Wb:\mathbb{R}^r\rightarrow\mathbb{R}$. To specify SVGD$^\infty$, we generalize the convolutional operator, and consider $\Wb$ as a function with two input arguments, {\it i.e.}, 
\vspace{-0.3cm}
{\small\begin{align*}
(\Wb*\mu_{t})(\thetab) \triangleq \int \Wb(\thetab, \thetab^\prime)\mu_{t}(\thetab^\prime)\mathrm{d}\thetab^\prime~.
\end{align*}}
Under this setting, we can specify the function $\Wb(\cdot, \cdot)$ for SVGD$^\infty$ as
\vspace{-0.2cm}
\begin{align}\label{eq:W}
\Wb(\thetab, \thetab^\prime) &\triangleq \nabla_{\thetab^\prime} \log p(\thetab^\prime|\Xb)\kappa(\thetab^\prime, \thetab) + \nabla_{\thetab^\prime} \kappa(\thetab^\prime, \thetab) \nonumber\\
&= \nabla_{\thetab^\prime}\left[p(\thetab^\prime|\Xb)\kappa(\thetab, \thetab^\prime)\right] / p(\thetab^\prime)~.
\end{align}
As will be shown in Section~\ref{sec:particle}, $\Wb$ in \eqref{eq:W} naturally leads to the SVGD algorithm, without the need to derive from an RKHS perspective.
\vspace{-0.2cm}
\begin{proposition}\label{prop:stationary_svgd}
	The stationary distribution of \eqref{eq:vlasov} is $\lim_{t\rightarrow\infty}\mu_t \triangleq \mu = p(\thetab|\Xb)$.
\end{proposition}
To interpret SVGD$^\infty$ as a WGF, we need to specify two quantities, the energy functional and an underlying metric to measure distances between density functions.
\vspace{-0.2cm}
\paragraph{Energy functional and distance metric of SVGD$^\infty$}
There are two ways to derive energy functionals for SVGD$^\infty$, depending on the underlying metrics for probability distributions. When adopting the WGF framework where $W_2$ is used as the underlying metric, according to \eqref{eq:gf_dis1}, the energy functional $E_{s}$ must satisfy
\vspace{-0.2cm}
{\small\begin{align}\label{eq:ef_svgd}
\nabla_{\thetab}&\left(\frac{\delta E_{s}}{\delta \mu_{t}}(\mu_{t})\right) = \Wb(\thetab, \thetab^\prime)*\mu_{t} \\
&= \mathbb{E}_{\thetab^\prime \sim \mu_{t}}\left[\nabla_{\thetab^\prime}\left[p(\thetab^\prime|\Xb)K(\thetab, \thetab^\prime)\right] / p(\thetab^\prime|\Xb)\right]~. \nonumber
\end{align}}
In general, there is no close-form solution for the above equation. Alternatively, \citet{liu2017stein_flow} proved another form of the energy functional by defining a different distance metric on the space of probability measures, called $\mathcal{H}$-Wasserstein distance:
\vspace{-0.2cm}
{\small\begin{align}\label{eq:wh}
W_{\mathcal{H}}(q_1, q_2) \triangleq \inf_{\phi_{t}, \mu_{t}}&\left\{\int_{0}^1\|\phi_{t}\|_{\mathcal{H}}\mathrm{d}t, \mbox{ s.t. } \mu_{t} = -\nabla_{\thetab}\cdot (\phi_{t}\mu_{t}), \right.\nonumber\\
&\left.\mu_0 = q_1, \mu_1 = q_2\|\right\}~,
\end{align}}
where $\phi_{t} \triangleq \Wb*\mu_t$, and $\|\cdot\|_{\mathcal{H}}$ is the norm in the Hilbert space induced by $\kappa(\cdot, \cdot)$. Under this metric, the underlying energy functional is proved to be the standard KL-divergence between $\mu_{t}$ and $p$, {\it e.g.}, 
\begin{align*}
E_s = \KL(\mu_{t}, p(\cdot|\Xb)). 
\end{align*}	
As can be seen in Section~\ref{sec:particle}, this interpretation allows one to derive SVGD, a particle-optimization-based algorithm to approximate the continuous-time equation \eqref{eq:vlasov}.

\subsection{SG-MCMC as WGF}
The continuous-time limit of SG-MCMC, when considering gradients to be exact, corresponds to standard It\'{o} diffusions. We consider the It\'{o} diffusion of SGLD for simplicity, {\it e.g.},
\vspace{-0.3cm}
\begin{align}\label{eq:diffusion}
\mathrm{d}\thetab_{t} = \frac{1}{2}\nabla U(\thetab_{t}) \mathrm{d}t + \mathrm{d}\mathcal{W}~.
\end{align}

\vspace{-0.4cm}
\paragraph{Energy functional}
The energy functional for SG-MCMC is easily seen by noting that the corresponding FP equation \eqref{eq:FPE} is in the gradient-flow form of \eqref{eq:gf_dis1}. Specifically, the energy functional $E$ is defined as:
{\small\begin{align}\label{eq:ito_energy}
	E(\mu)\triangleq \underbrace{-\int U(\thetab)\mu(\thetab)\mathrm{d}\thetab}_{E_1} + \underbrace{\int \mu(\thetab)\log\mu(\thetab)\mathrm{d}\thetab}_{E_2}
	\end{align}}
Note $E_2$ is the energy functional of a pure Brownian motion ({\it e.g.}, $U(\thetab) = 0$ in \eqref{eq:diffusion}). We can verify \eqref{eq:ito_energy} by showing that it satisfies that FP equation. According to \eqref{eq:gf_dis1}, the first variation of $E_1$ and $E_2$ is calculated as
{\small\begin{align}\label{eq:firstvariation}
\frac{\delta E_1}{\delta \mu} = -U,~~~~\frac{\delta E_2}{\delta \mu} = \log \mu + 1~.
\end{align}}
Substituting \eqref{eq:firstvariation} into \eqref{eq:gf_dis1} recovers the FP equation \eqref{eq:FPE} for the It\'{o} diffusion \eqref{eq:diffusion}.
\vspace{-0.2cm}
\section{Particle Optimization}\label{sec:particle}
\vspace{-0.3cm}
An efficient way to solve the generally infeasible WGF formula \eqref{eq:gf_dis1} is to adopt numerical methods with particle approximation. With a little abuse of notation but for conciseness, we do not distinguish subscripts $t$ and $\ell$ for the particle $\thetab$, {\it i.e.}, $\thetab_t$ denotes the continuous-time version of the particle, while $\thetab_\ell$ denotes the discrete-time version. We develop several techniques to approximate different types of WGF for SG-MCMC and SVGD. In particle approximation, the continuous density $\mu_{t}$ is approximated by a set of $M$ particles $(\thetab_{t}^{(i)})_{i=1}^M$ that evolve over time $t$ with weights $(m_i)_{i=1}^M$ such that $\sum_{i=1}^Mm_i = 1$, {\it i.e.}, 
$\mu_{t} \approx \sum_{i=1}^Mm_i \delta(\thetab_{t}^{(i)})$, 
where $\delta(\thetab_t^{(i)}) = 1$ when $\thetab = \thetab_t^{(i)}$ and 0 otherwise. Typically $m_i's$ are chosen at the beginning and fixed over time, thus we assume $m_i = \frac{1}{M}$ and rewrite $\mu_{t} \approx\frac{1}{M}\sum_{i=1}^M\delta(\thetab_{t}^{(i)})$ in the following for simplicity. We investigate two types of particle-approximation methods in the following, discrete gradient flows and by blob methods.

\paragraph{Particle approximation by discrete gradient flows}
Denote $\mathcal{P}_s(\mathbb{R}^r)$ be the space of probability measures with finite 2nd-order moments. Define the following optimization problem with stepsize $h$:
\begin{align}\label{eq:proximal}
J_h(\mu) \triangleq \arg\min_{\nu\in \mathcal{P}_s(\mathbb{R}^d)}\left\{\frac{1}{2h}W_2^2(\mu, \nu) + E(\nu)\right\}~.
\end{align}
A discrete gradient flow of the continuous one in \eqref{eq:gf_dis1} up to time $T$ is the composition of a sequence of the solutions $(\tilde{\mu}_{\ell})_{\ell=1}^{T/h}$ of \eqref{eq:proximal}, {\it i.e.},
\vspace{-0.1cm}
\begin{align}\label{eq:discretegf}
\tilde{\mu}_{\ell} \triangleq J_h(\tilde{\mu}_{\ell-1}) = J_h(J_h(\cdots \mu_0)) \triangleq J_h^{\ell}\mu_0~.
\end{align}
One can show that when $h\rightarrow 0$, the discrete gradient flow \eqref{eq:discretegf} converges to the true flow \eqref{eq:gf_dis1} for all $\ell$. Specifically, let $\partial E(\mu)$ be the set of Wasserstein subdifferential of $E$ at $\mu$, {\it i.e.}, $\xi \in \partial E(\mu)$ if $\partial_t \mu = \xi$ is satisfied. Define $|\partial E|(\mu) = \min\{\|\xi\|_{L^2(\mu)}: \xi\in\partial E(\mu)\}$ to be the minimum norm of the elements in $\partial E(\mu)$. We have
\begin{lemma}[\cite{Craig:thesis14}]\label{lem:discrete_error}
	Assume $E$ is proper, coercive and lower semicontinuous (specify in Section~\ref{app:lemma3} of the Supplementary Material (SM)). For an $\mu_0$ and $t\geq 0$, as $\frac{T}{h}\rightarrow \infty$, the discrete gradient sequence $\tilde{\mu}_{T/h} \triangleq J_{h}^{T/h}\mu_0$ converge uniformly in $t$ to a compact subset of $[0, +\infty)$, and $W_2^2(\tilde{\mu}_{T/h}, \mu_T) \leq \sqrt{3}|\partial E|(\mu)\sqrt{Th}$.
\end{lemma}

Lemma~\ref{lem:discrete_error} suggests the discrete gradient flow can approximate the original WGF arbitrarily well if a small enough stepsize $h$ is adopted. Consequently, one solves \eqref{eq:discretegf} through a sequence of optimization procedures to update the particles. We will derive a particle-approximation method for the $W_2$ term in \eqref{eq:proximal}, which allows us to solve SG-MCMC efficiently. However, this technique is not applicable to SVGD, as we neither have an explicit form of the energy functional in \eqref{eq:ef_svgd} when adopting the $W_2$ metric, nor have an explicit form for the metric $W_{\mathcal{H}}$ in \eqref{eq:wh} when adopting the KL-divergence as the energy functional. Fortunately, this can be solved by the second approximation method called blob methods.
\vspace{-0.3cm}
\paragraph{Particle approximation by blob methods}
The name of blob methods comes from  the classical fluids literature, where instead of evolving the density in \eqref{eq:gf_dis1}, one evolves all particles on a grid with time-spacing $h$ \citep{CarrilloCP:arxiv17}. Specifically, note the function $\vb_{t}$ in \eqref{eq:gf_dis1} represents velocity of particles via transportation map $\mathcal{T}$, thus solving a WGF is equivalent to evolving the particles along their velocity in each iteration. Formally, one can prove
\begin{proposition}[\cite{CRAIGB:MC16}]\label{prop:blob}
	Let $\mu_{0} \approx\frac{1}{M}\sum_{i=1}^M\delta(\thetab_{0}^{(i)})$. Assume $\vb_{t}$ in \eqref{eq:gf_dis1} is well-defined and continuous w.r.t.\! each $\thetab_{t}^{(i)}$ at time $t$. Then solving the PDE \eqref{eq:gf_dis1} reduces to solving a system of ordinary differential equations for the locations of the Dirac masses:
	\vspace{-0.3cm}
	\begin{align}\label{eq:par_ode}
	\mathrm{d}\thetab_{t}^{(i)}/\mathrm{d}t = -\vb_{t}(\thetab_{t}^{(i)})~.
	\end{align}
\end{proposition}
\vspace{-0.2cm}
Proposition~\ref{prop:blob} suggests evolving each particle along the directions defined by $\vb_{t}$, eliminating the requirement to know an explicit form of the energy functional. In the following, we apply the above particle-optimization techniques to derive algorithms for SVGD and SG-MCMC.

\subsection{A particle-optimization algorithm for SVGD}
\vspace{-0.2cm}
As mentioned above, discrete-gradient-flow approximation does not apply to SVGD. We thus rely on the blob method. From Section~\ref{sec:svgd_wgf}, $\vb_{t}$ in SVGD is defined as $\vb_{t}(\thetab) = (\Wb*\mu_{t})(\thetab)$. When $\mu_{t}(\thetab)$ is approximated by particles, $\vb_{t}(\thetab_{t}^{(i)})$ is simplified as:
\vspace{-0.2cm}
{\small\begin{align*}
\vb_{t}(\thetab_{t}^{(i)}) = \frac{1}{M}\sum_{j=1}^M\Wb(\thetab_{t}^{(i)}, \thetab_{t}^{(j)})~.
\end{align*}}
As a result, with the definition of $\Wb$ in \eqref{eq:W}, updating $\{\thetab_{t}^{(i)}\}$ by time discretizing \eqref{eq:par_ode} recovers the update equations for standard SVGD in \eqref{eq:svgd_update}.
\vspace{-0.3cm}
\subsection{Particle-optimization algorithms for SG-MCMC}
\vspace{-0.2cm}
Both the discrete-gradient-flow and the blob methods can be applied for SG-MCMC, which are detailed below.
\vspace{-0.4cm}
\paragraph{Particle optimization with discrete gradient flows}
We first specify Lemma~\ref{lem:discrete_error} in the case of SG-MCMC in Lemma~\ref{lem:variational_fp}, which is known as the Jordan-Kinderlehrer-Otto scheme \citep{JordanKO:MA98}.
\vspace{-0.1cm}
\begin{lemma}[\cite{JordanKO:MA98}]\label{lem:variational_fp}
	Assume that $p(\thetab_t|\Xb)\leq C_1$ is infinitely differentiable, and $\|\nabla_{\thetab}\log p(\thetab|\Xb)\| \leq C_2\left(1 + C_1 - \log p(\thetab|\Xb)\right) (\forall \thetab)$ for some constants $\{C_1, C_2\}$. Let $T = h K$ with $K$ the number of iterations, $\tilde{\mu}_0$ be an arbitrary distribution with same support as $p(\thetab|\Xb)$, and $\{\tilde{\mu}_k\}_{k=1}^K$ be the solution of the functional optimization problem:
	\vspace{-0.2cm}
	{\small\begin{align}\label{eq:variationalFP}
		\tilde{\mu}_k = \arg\min_{\mu \in \mathcal{P}_s(\mathbb{R}^r)}\KL\left(\mu \| p\right) + \frac{1}{2h}W^2_2\left(\tilde{\mu}_{k-1}, \mu\right)~.
		\end{align}}
	Then $\tilde{\mu}_K$ converges to $\mu_T$ in the limit of $h\rightarrow 0$, {\it i.e.}, $\lim_{h\rightarrow 0}\tilde{\mu}_K = \mu_T$, where $\mu_T$ is the solution of the FP equation \eqref{eq:FPE} at time $T$.
\end{lemma}

According to Lemma~\ref{lem:variational_fp}, it is apparent that SG-MCMC can be implemented by iteratively solving the optimization problem in \eqref{eq:variationalFP}. However, particle approximations for both terms in \eqref{eq:variationalFP} are challenging. In the following, we develop efficient techniques to solve the problem.

First, rewrite the optimization problem in \eqref{eq:variationalFP} as
\vspace{-0.1cm}
\begin{align*}
\hspace{-0.2cm}\min_{\mu \in \mathcal{P}_s(\mathbb{R}^r)}\underbrace{-\mathbb{E}_{\mu}[\log p(\thetab|\Xb)]}_{F_1} + \underbrace{\mathbb{E}_{\mu}[\log \mu] + \frac{1}{2h}W^2_2\left(\tilde{\mu}_{k-1}, \mu\right)}_{F_2}
\end{align*}
We aim at deriving gradient formulas for both the $F_1$ and $F_2$ terms under a particle approximation in order to perform gradient descent for the particles. Let $\mu \approx \frac{1}{M}\sum_{i=1}^M\delta(\thetab^{(i)})$. The gradient of $F_1$ is easily approximated as\vspace{-0.4cm}
\begin{align}\label{eq:gradf1}
\frac{\partial F_1}{\partial \thetab^{(i)}} \approx -\nabla_{\thetab^{(i)}}\log p(\thetab^{(i)}|\Xb)~.
\end{align}
To approximate the gradient for $F_2$, let $p_{ij}$ denote the joint distribution of the particle-pair $(\thetab^{(i)}, \thetab_{k-1}^{(j)})$. Note $\mathbb{E}_{\mu}[\log \mu]$ is minimized when the particles $\{\thetab^{(i)}\}$ are uniformly distributed. In other words, the marginal distribution vector $(\sum_j p_{ij})_i$ is a uniform distribution. Combining $\mathbb{E}_{\mu}[\log \mu]$ with the definition of $W_2$, calculating $F_2$ is equivalent to solving the following optimization problem:
{\small\begin{align}\label{eq:F2}
P\triangleq &\{p_{ij}\} = \arg\min_{p_{i,j}}\sum_{i,j}p_{ij}d_{ij} \\
s.t.&~~ \sum_j p_{ij} = \frac{1}{M}, ~~\sum_i p_{ij} = \frac{1}{M}~,\nonumber
\end{align}}\par \vspace{-0.4cm}
where $d_{ij} \triangleq \|\thetab^{(i)} - \thetab_{k-1}^{(j)}\|^2$. We can further enforce the joint distribution $\{p_{ij}\}$ to have maximum entropy by introducing a regularization term $\mathbb{E}_{p_{ij}}[\log p_{ij}]$, which is stronger than the regularizer enforced for the marginal distribution above. After introducing Lagrangian multipliers $\{\alpha_i , \beta_i\}$ to deal with the constraints in \eqref{eq:F2}, we arrive at the dual problem:
\begin{align*}
\max&\mathcal{L}^D(\{p_{ij}\}, \{\alpha_i\}, \{\beta_i\}) = \lambda\sum_{i,j}p_{ij}\log p_{ij} + p_{ij}d_{ij} \\
&+ \sum_i \alpha_i(\sum_j p_{ij} - \frac{1}{M}) + \sum_j \beta_j(\sum_i p_{ij} - \frac{1}{M})~,
\end{align*}
where $\lambda$ is the weight for the regularizer. The optimal $p_{ij}$'s can be obtained by applying KKT conditions to set the derivative w.r.t.\! $p_{ij}$ to be zero, ending up with the following form:
\vspace{-0.4cm}
\begin{align*}
p_{ij}^* = u_i e^{-d_{ij}/\lambda}v_j~,
\end{align*}	
where $u_i \triangleq e^{-\frac{1}{2}-\frac{\alpha_i}{\lambda}}$, $v_j = e^{-\frac{1}{2}-\frac{\beta_j}{\lambda}}$. As a result, the particle gradients on $F_2$ can be approximated as
\begin{align}\label{eq:gradf2}
\frac{\partial F_2}{\partial \thetab^{(i)}} &\approx -\frac{\sum_ju_iv_jd_{ij}e^{-d_{ij}/\lambda}}{\partial \thetab^{(i)}} \\
=& \sum_j 2u_iv_j(\frac{d_{ij}}{\lambda} - 1)e^{-d_{ij}/\lambda}(\thetab^{(i)} - \thetab_{k-1}^{(j)})~.\nonumber
\end{align}\par \vspace{-0.3cm}
Theoretically, we need to adaptively update $\{u_i, v_j\}$ as well to ensure the constraints in \eqref{eq:F2}. In practice, however, we use a fixed scaling factor $\gamma$ to approximate $u_iv_j$ for the sake of simplicity.

Particle gradients are obtained by combining \eqref{eq:gradf1} and \eqref{eq:gradf2}, which are then used to update the particles $\{\thetab^{(i)}\}$ by standard gradient descent. Intuitively, \eqref{eq:gradf1} encourages particles move to local modes while \eqref{eq:gradf2} regularizes particle interactions. Different from SVGD, our scheme imposes both attractive and repulsive forces for the particles. Specifically, by inspecting \eqref{eq:gradf2}, we can conclude that: $\RN{1})$ When $\thetab^{(i)}$ is far from a previous particle $\thetab_k^{(j)}$, {\it i.e.}, $\frac{d_{ij}}{\lambda} > 1$, $\thetab^{(i)}$ is pulled close to $\{\thetab_k^{(j)}\}$ with force proportional to $(\frac{d_{ij}}{\lambda} - 1)e^{-d_{ij}/\lambda}$; $\RN{2})$ when $\thetab^{(i)}$ is close enough to a previous particle $\thetab_k^{(j)}$, {\it i.e.}, $\frac{d_{ij}}{\lambda} < 1$, $\thetab^{(i)}$ is pushed away, preventing it from collapsing to $\thetab_k^{(j)}$.

\paragraph{Particle optimization with blob methods}
The idea of blob methods can also be applied to particle approximation for SG-MCMC, which require the velocity vector field $\vb_{t}$. According to \eqref{eq:ito_energy}, this is calculated as: $\vb_{t}(\thetab) = -\nabla_{\thetab}\frac{\delta (E_1 + E_2)}{\delta \mu} = -\nabla_{\thetab} U - \nabla_{\thetab} \mu/\mu$. Unfortunately, direct application of particle approximation is infeasible because the term $\nabla_{\thetab}\mu$ is undefined with discrete $\mu$. To tackle this problem, we adopt the idea in \cite{CarrilloCP:arxiv17} to approximate the energy functional $E_2$ in \eqref{eq:ito_energy} as: $E_2 \approx \int \mu(\thetab) \log (\mu*K)(\thetab) \mathrm{d}\thetab$, where $K(\cdot, \cdot)$ is another kernel function to smooth out $\mu$. Consequently, based on \cite{CarrilloCP:arxiv17}, the velocity $\vb_{t}$ can be calculated as (details in Section~\ref{supp:particle_derivation} of the SM):
\begin{align}\label{eq:velocity}
\vb_{t}(\thetab) &= -\nabla_{\thetab}U - \sum_{j=1}^n\nabla_{\thetab_{t}^{(j)}}K(\thetab, \thetab_{t}^{(j)})/\sum_kK(\thetab_{t}^{(j)}, \thetab_{t}^{(k)}) \nonumber\\
&- \sum_{j=1}^n\nabla_{\thetab_{t}^{(j)}} K(\thetab, \thetab_{t}^{(j)})/\sum_{k=1}^nK(\thetab, \thetab_{t}^{(k)})
\end{align}
Given $\vb_{t}$, particle updates can be obtained by solving \eqref{eq:par_ode} numerically as in SVGD. By inspecting the formula of $\vb_{t}$ in \eqref{eq:velocity}, the last two terms both act as repulsive forces. Interestingly, the mechanism is similar to SVGD, but with adaptive force between different particle pairs.

\section{The General Recipe}

Based on the above development, a more general particle-optimization framework is proposed by combining  the PDEs of both SG-MCMC and SVGD. As a result, we propose the following PDE to drive evolution of densities
\begin{align}\label{eq:unified}
\frac{\partial \mu_t}{\partial t} = &-\nabla_{\thetab}\cdot \left(\mu_tF(\thetab_t)\right) + \lambda_1\nabla_{\thetab}\cdot\left((\Wb*\mu_{t})\mu_{t}\right) \nonumber\\
&+ \lambda_2\nabla_{\thetab}\nabla_{\thetab}\!:\!\left(\mu_tg(\thetab_t)g^{\top}(\thetab_t)\right)~,
\end{align}
where $\lambda_1$ and $\lambda_2$ are two constants. It is easily seen that to ensure the stationary distribution of \eqref{eq:unified} to be equal to $p(\thetab|\Xb)$, the following condition must be satisfied:
\begin{align}\label{eq:generalFP}
\nabla_{\thetab}\cdot &\left(p(\thetab|\Xb)F(\thetab)\right) = \lambda_1\nabla_{\thetab}\cdot\left((\Wb*p(\thetab|\Xb))p(\thetab|\Xb)\right) \nonumber\\
&+ \lambda_2\nabla_{\thetab}\nabla_{\thetab}\!:\!\left(p(\thetab|\Xb)g(\thetab)g^{\top}(\thetab)\right)
\end{align}
There are many feasible choices for the functions and parameters $\{F(\thetab), \Wb, g(\thetab), \lambda_1, \lambda_2\}$ to satisfy \eqref{eq:generalFP}. However, the verification procedure might be complicated given the present of a convolutional term in \eqref{eq:generalFP}. We recommend the following choices for simplicity:
\vspace{-0.3cm}
\begin{itemize}
	\item $F(\thetab) = \frac{1}{2}U(\thetab)$, $\Wb = 0$, $g(\thetab) = \Ib$ and $\lambda_2 = 1$: this reduces to the Wasserstein-based SGLD with particle optimization. Specifically, when the discrete-gradient-flow approximation is adopted, the algorithm is denoted as $w$-SGLD; whereas when the blob method is adopted, it is denoted as $w$-SGLD-B.
	\item $F(\thetab) = 0$, $g(\thetab) = 0$, $\Wb$ is defined as \eqref{eq:W}: this reduces to standard SVGD.
	\item $F(\thetab) = \frac{1}{2}U(\thetab)$, $g(\thetab) = \Ib$, $\Wb$ is defined as \eqref{eq:W}, and $\lambda_2 = 1$: this is the combination of SGLD and SVGD, and is called particle interactive  SGLD, denoted as PI-SGLD or $\pi$-SGLD.
\end{itemize}
\vspace{-0.3cm}
It is easy to verify that condition \eqref{eq:generalFP} is satisfied for all the above three particle-optimization algorithms. Furthermore, particle updates are readily developed by applying either the discrete-gradient-flow or blob-based methods.

%
\vspace{-0.4cm}
\section{Related Particle-Based MCMC Methods}
\vspace{-0.4cm}
There have been related particle-based MCMC algorithms. Representative methods are sequential Monte Carlo (SMC) \citep{DelMoralDJ:JRSS06}, particle MCMC (PMCMC) \citep{AndrieuDH:JRSS10} and many variants. In SMC, particles are sample from a proposal distribution, and the corresponding weights are updated by a resampling step. PMCMC extends SMC by sampling from an extended distribution interacted with a MH-rejection step. Compared to our framework, their proposal distributions are typically hard to choose; furthermore, optimality of the particles from both methods can not be guaranteed. Furthermore, the methods are typically much more computationally expensive. Recently, \cite{DaiHDS:AISTATS16} proposed a particle-based MCMC algorithm by approximating a target distribution with weighted kernel density estimator, which updates particle weights based on likelihoods of the corresponding particles. This approach is theoretically sound but lacks an underlying geometry interpretation. Finally, we note that $w$-SGLD has been successfully applied to reinforcement learning recently for improved policy optimization \citep{ZhangCLC:ICML18}.

\vspace{-0.4cm}
\section{Experiments}
\vspace{-0.4cm}
We verify our framework on a set of experiments, including a number of toy experiments and applications to Bayesian sampling of deep neural networks (DNNs).
\vspace{-0.4cm}
\subsection{Demonstrations}
\begin{figure}[h!] \centering
	\begin{tabular}{ccccc}
		\hspace{-4mm}
		\includegraphics[width=2cm]{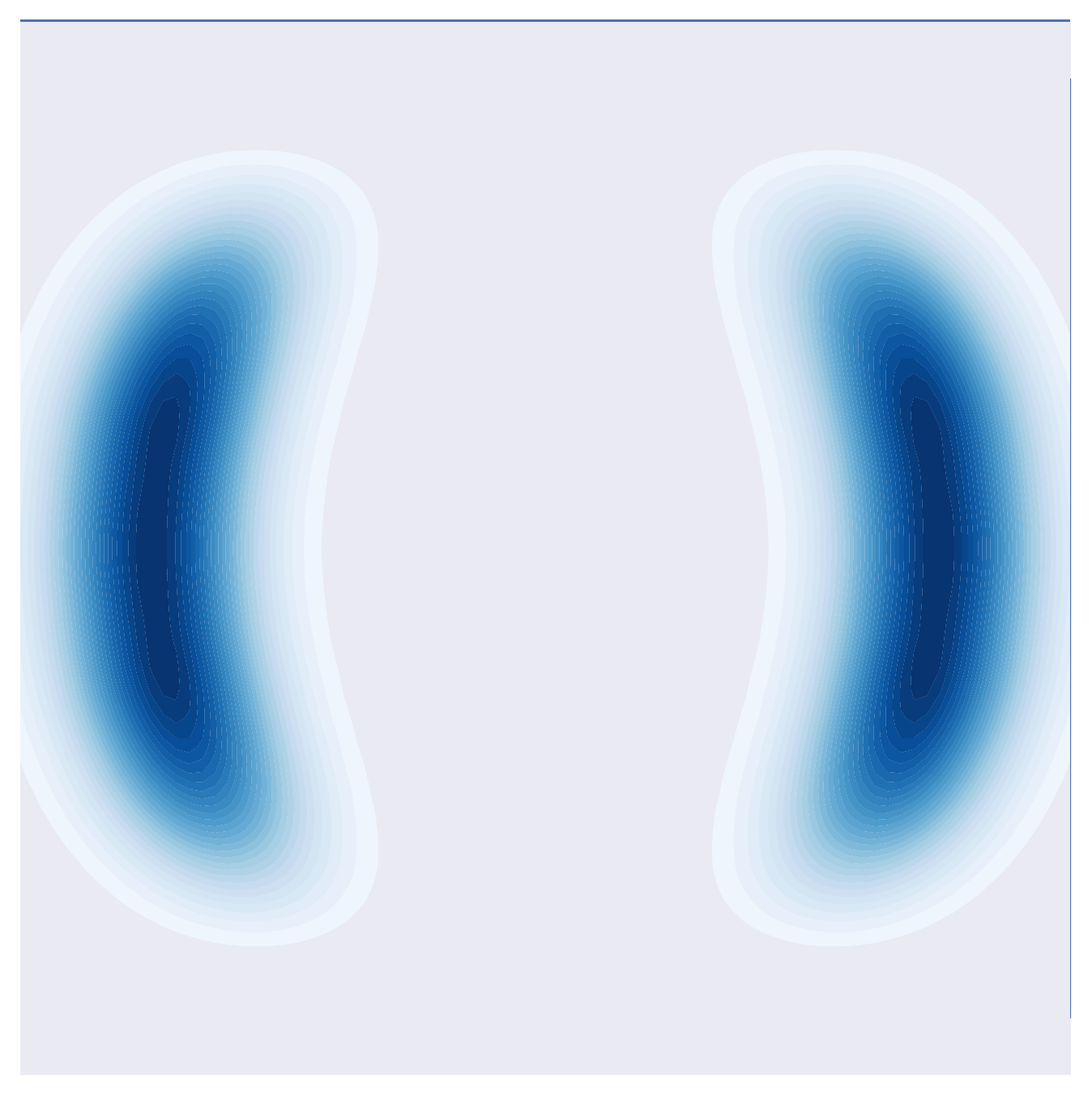}  
		&   \hspace{-5mm}
		\includegraphics[width=2cm]{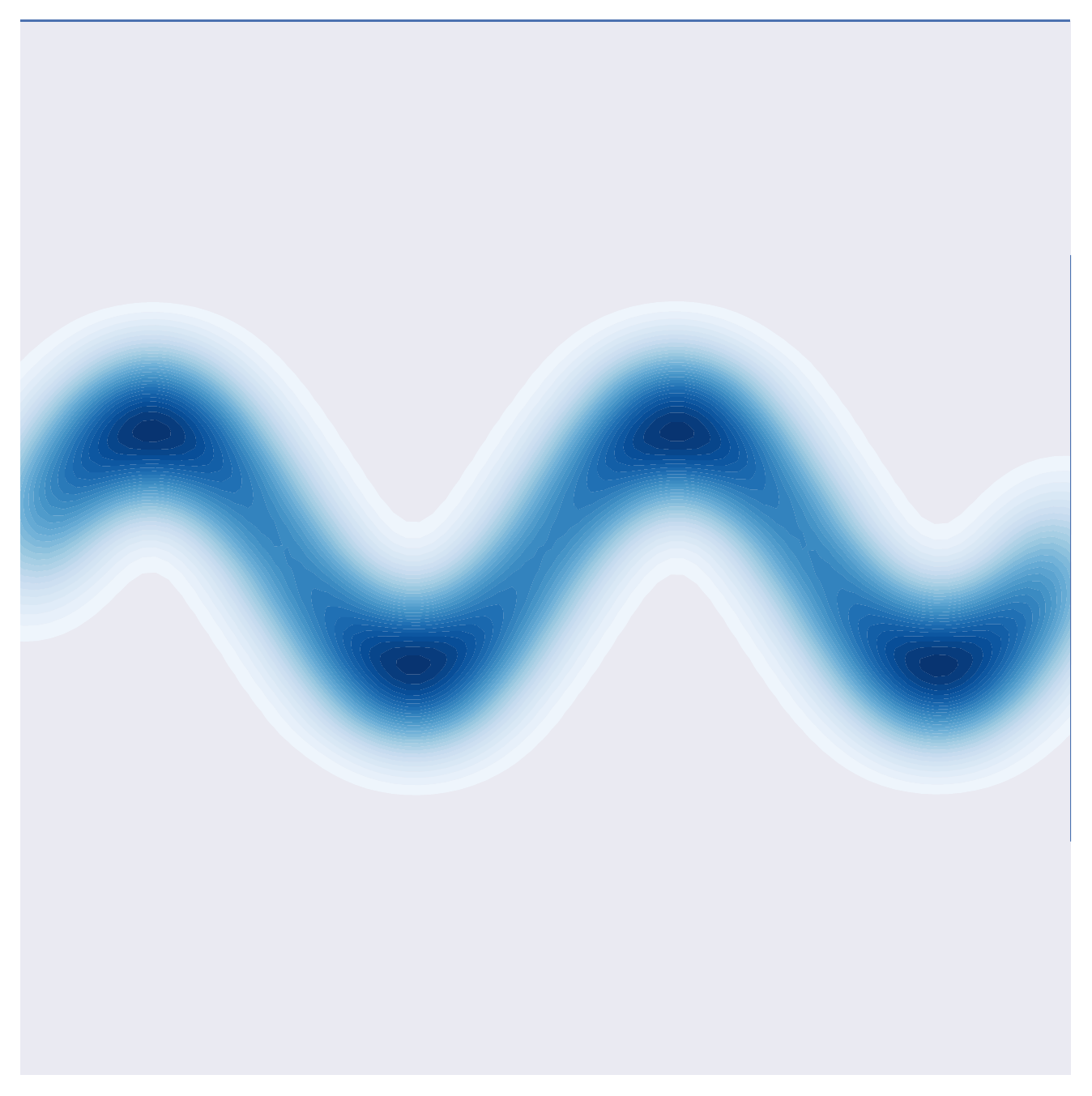} 
		& 		\hspace{-5mm}
		\includegraphics[width=2cm]{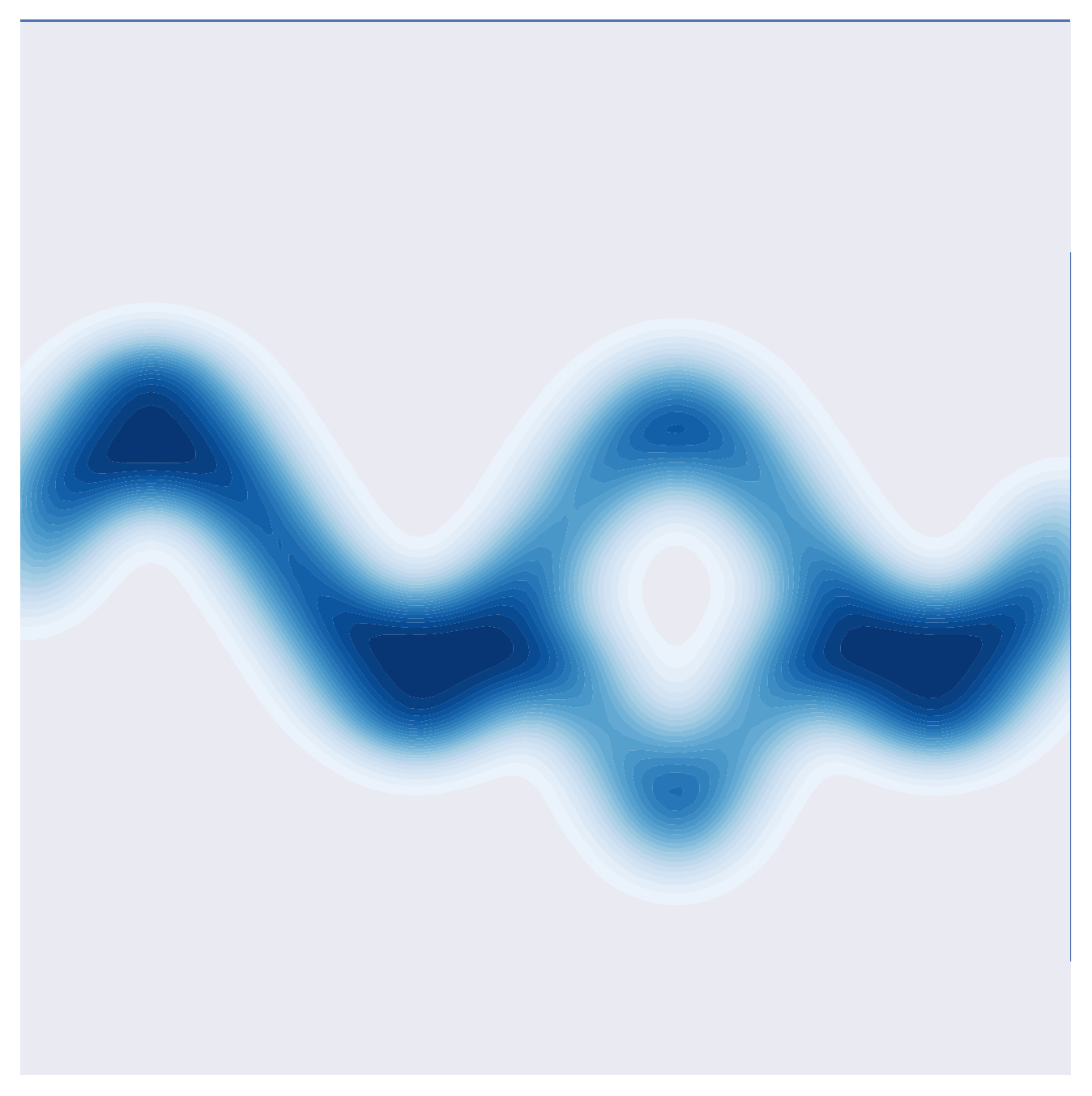}   
		&		\hspace{-5mm}
		\includegraphics[width=2cm]{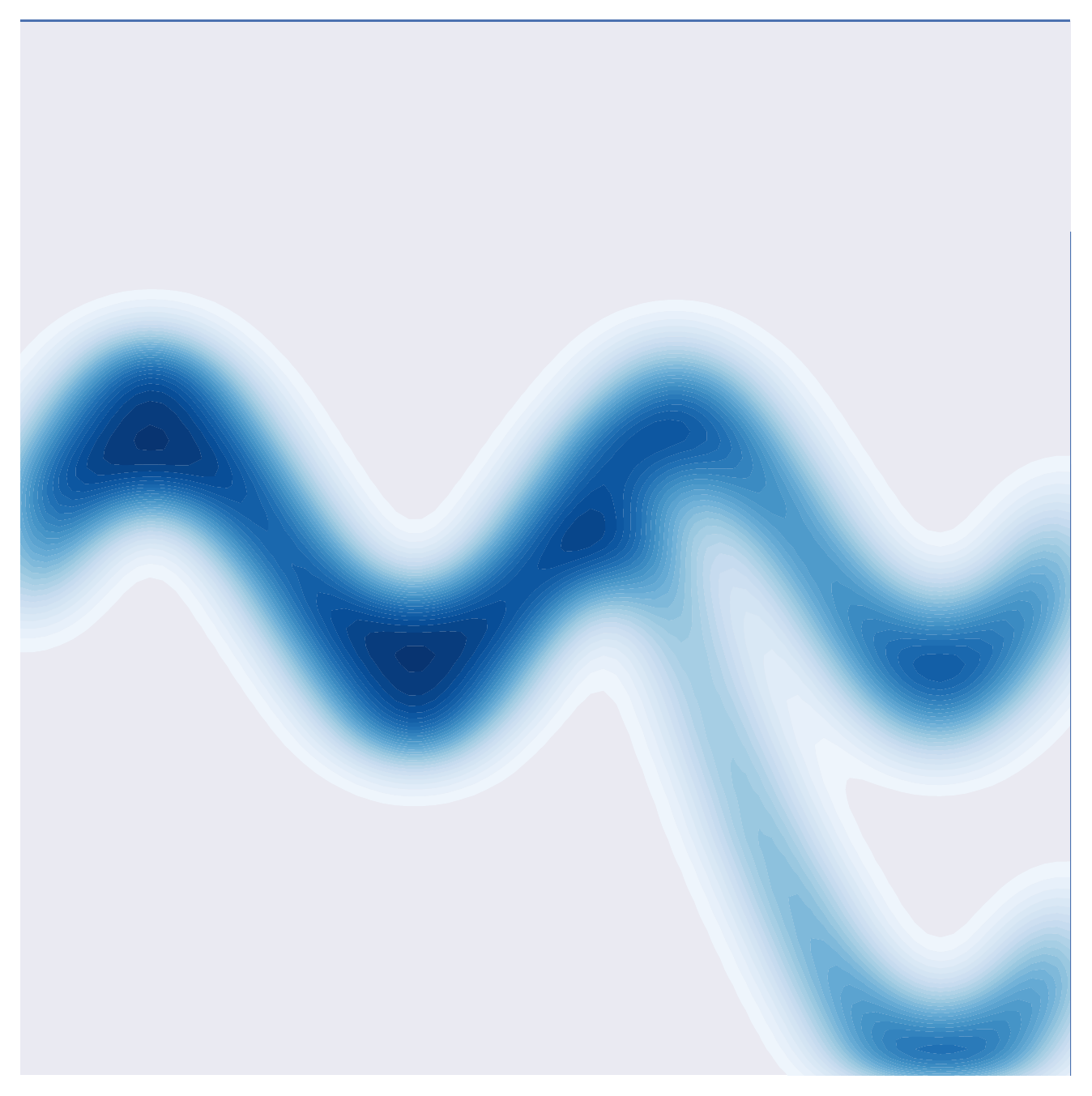}   
		\vspace{-1mm}
		\\
		\hspace{-4mm}
		\includegraphics[width=2cm]{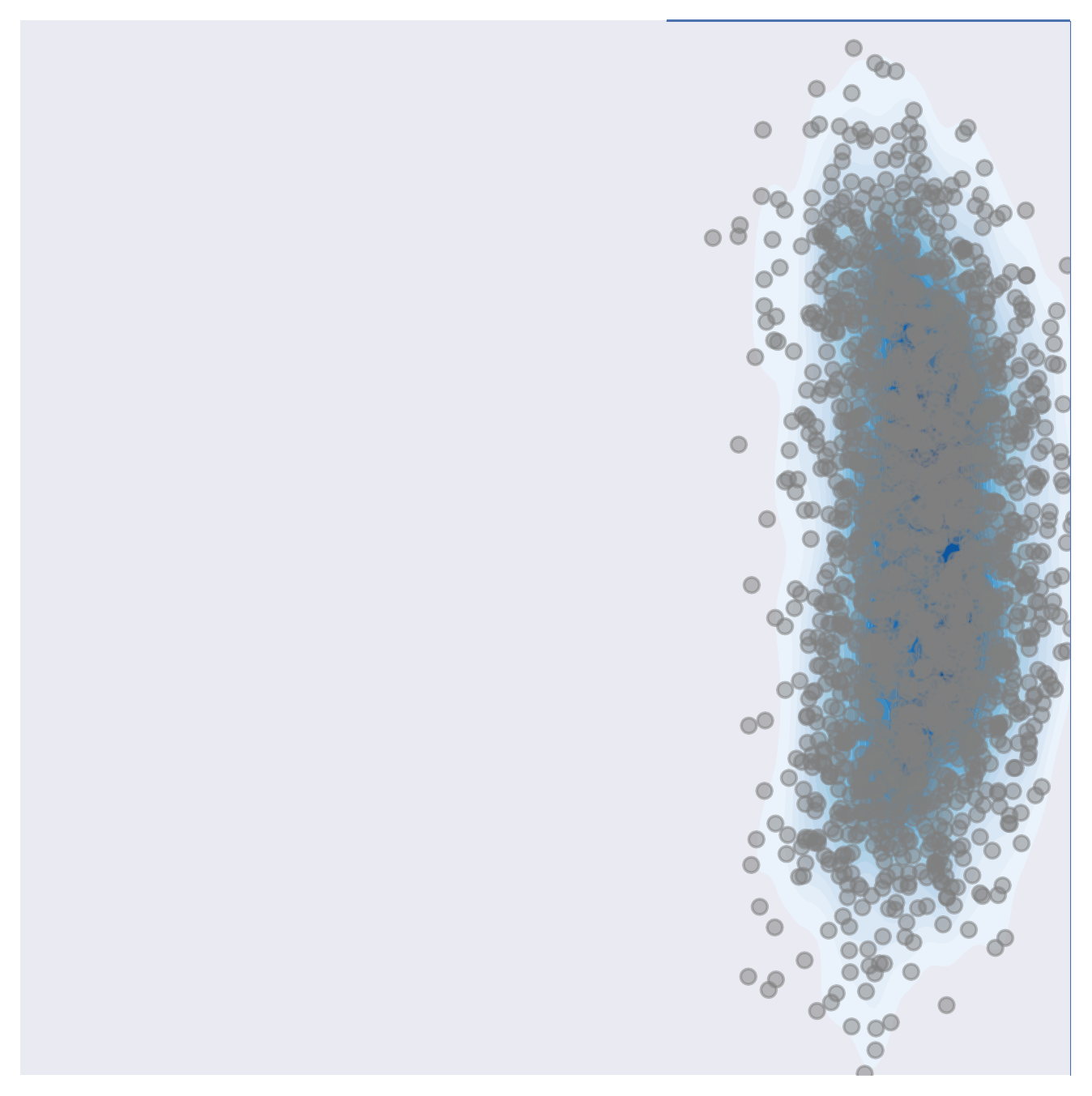}  
		&   \hspace{-5mm}
		\includegraphics[width=2cm]{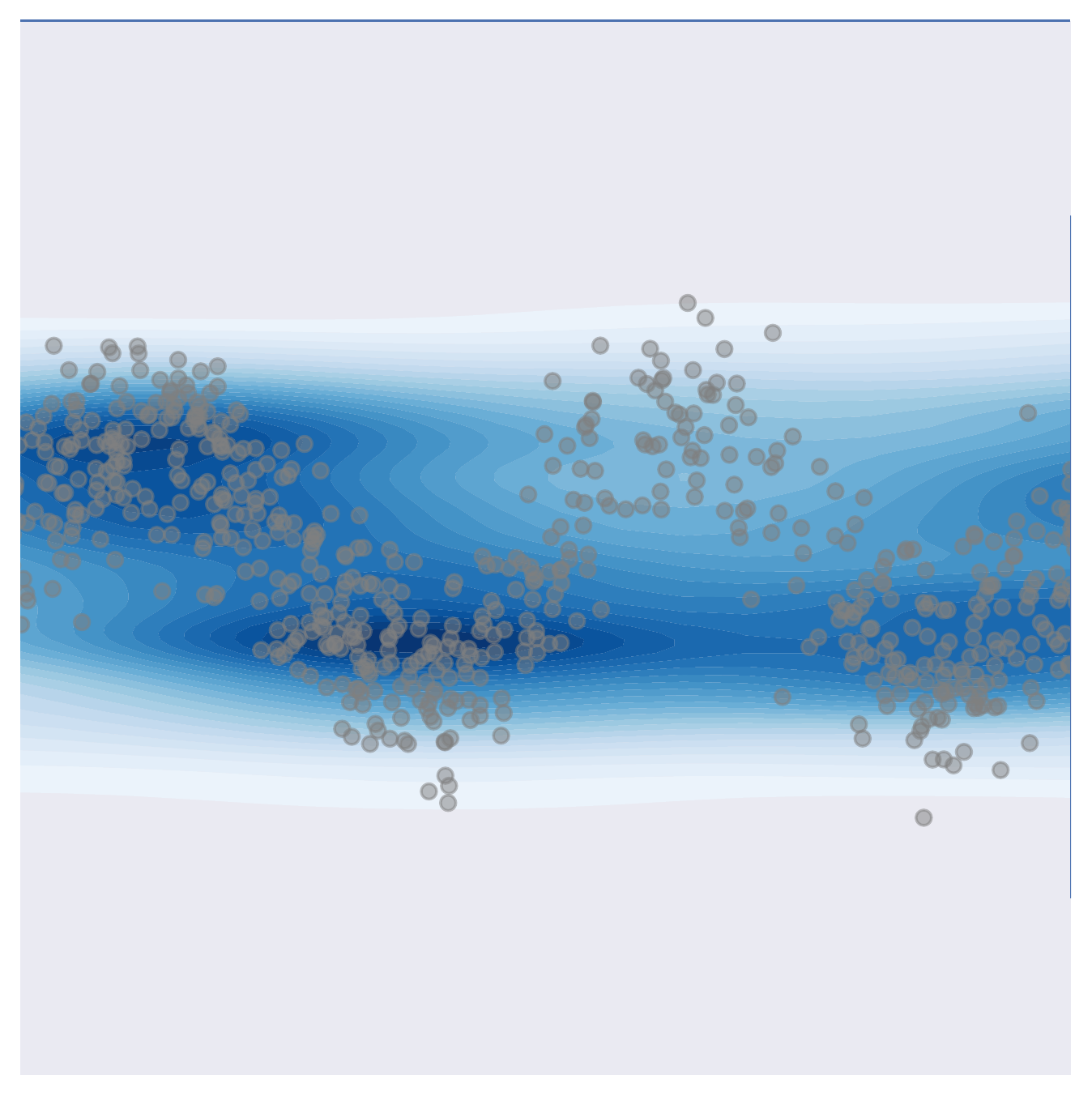} 
		& 		\hspace{-5mm}
		\includegraphics[width=2cm]{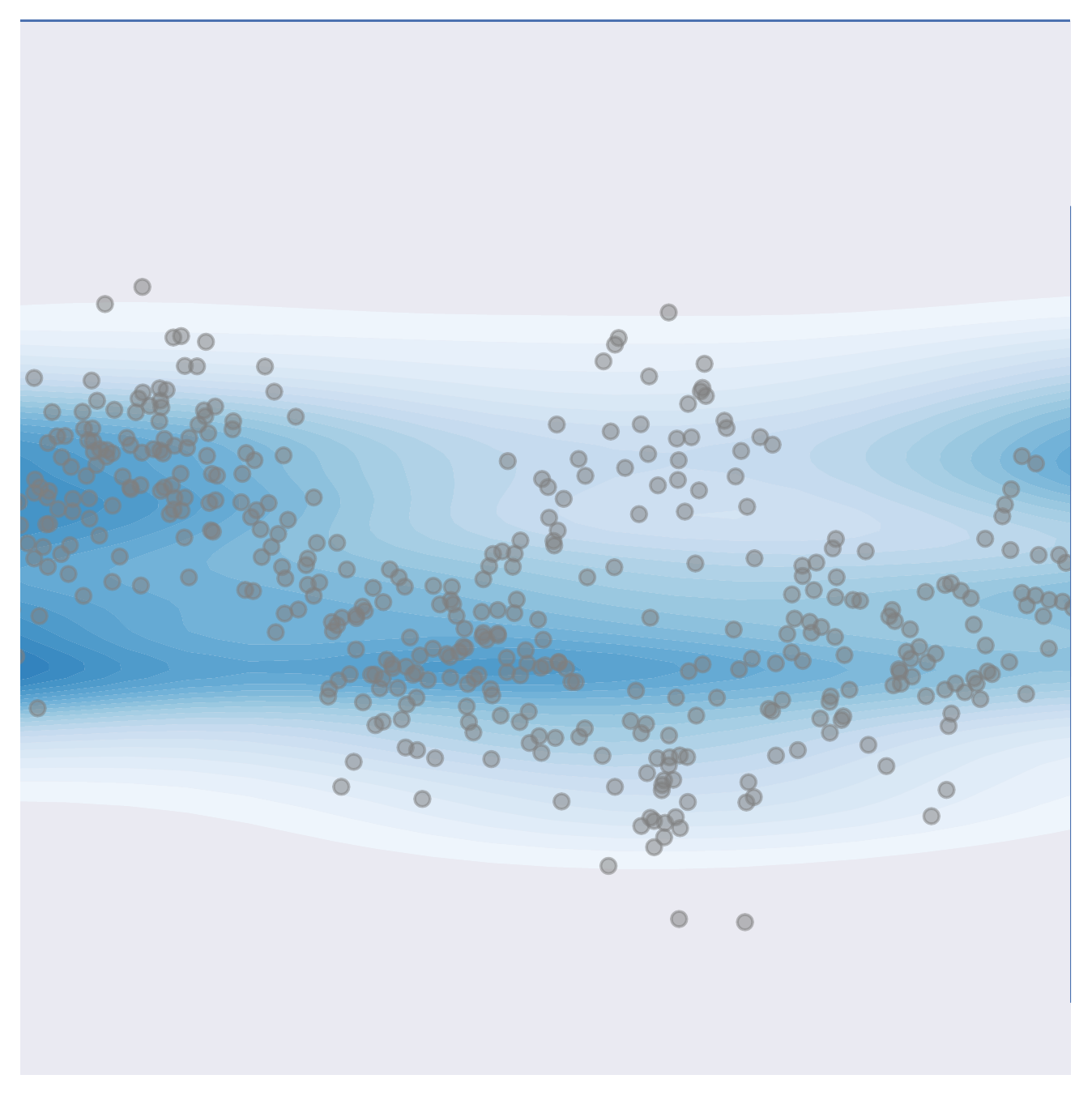}   
		&		\hspace{-5mm}
		\includegraphics[width=2cm]{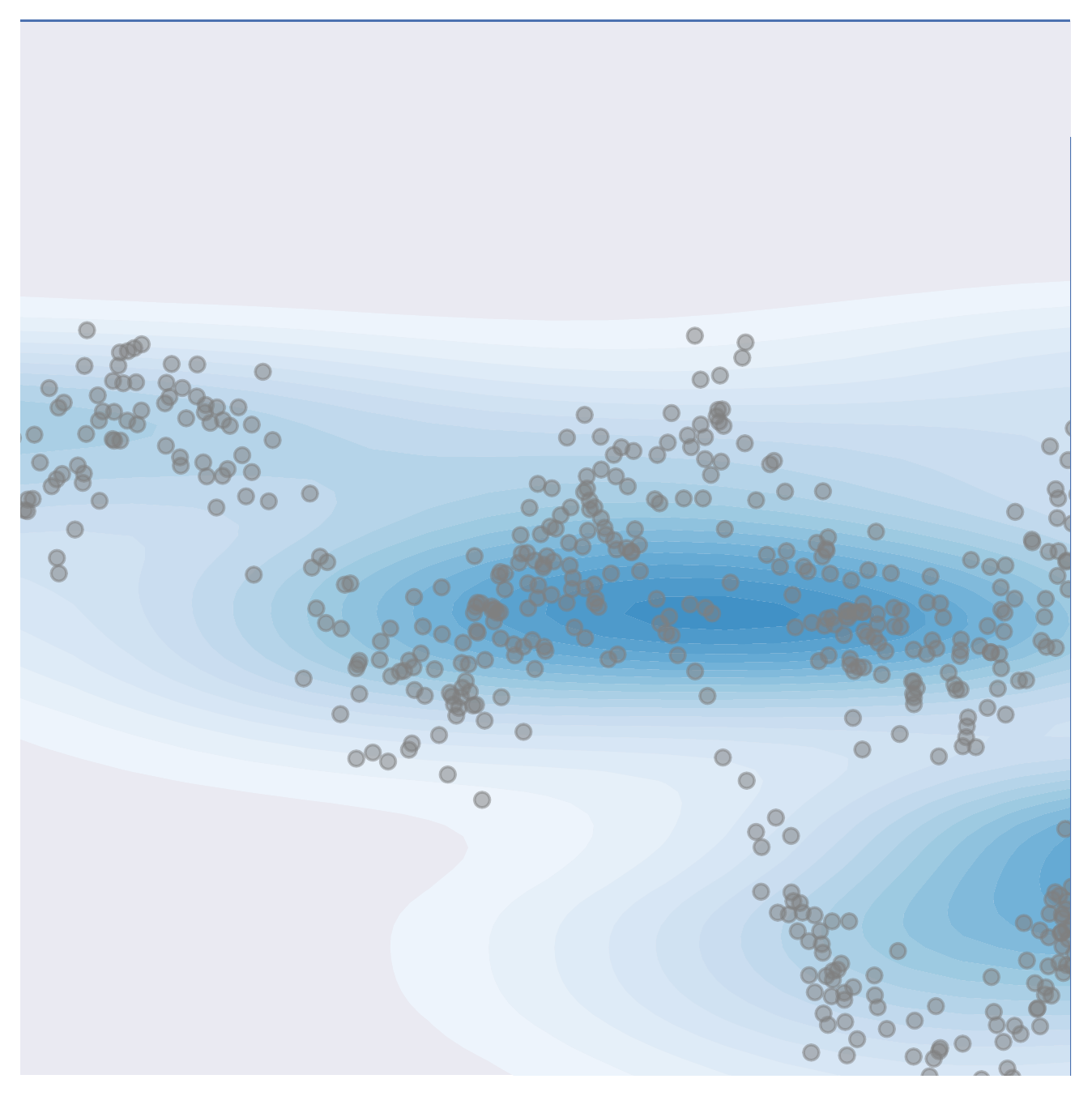}
		\vspace{-1mm}
		\\
		\hspace{-4mm}
		\includegraphics[width=2cm]{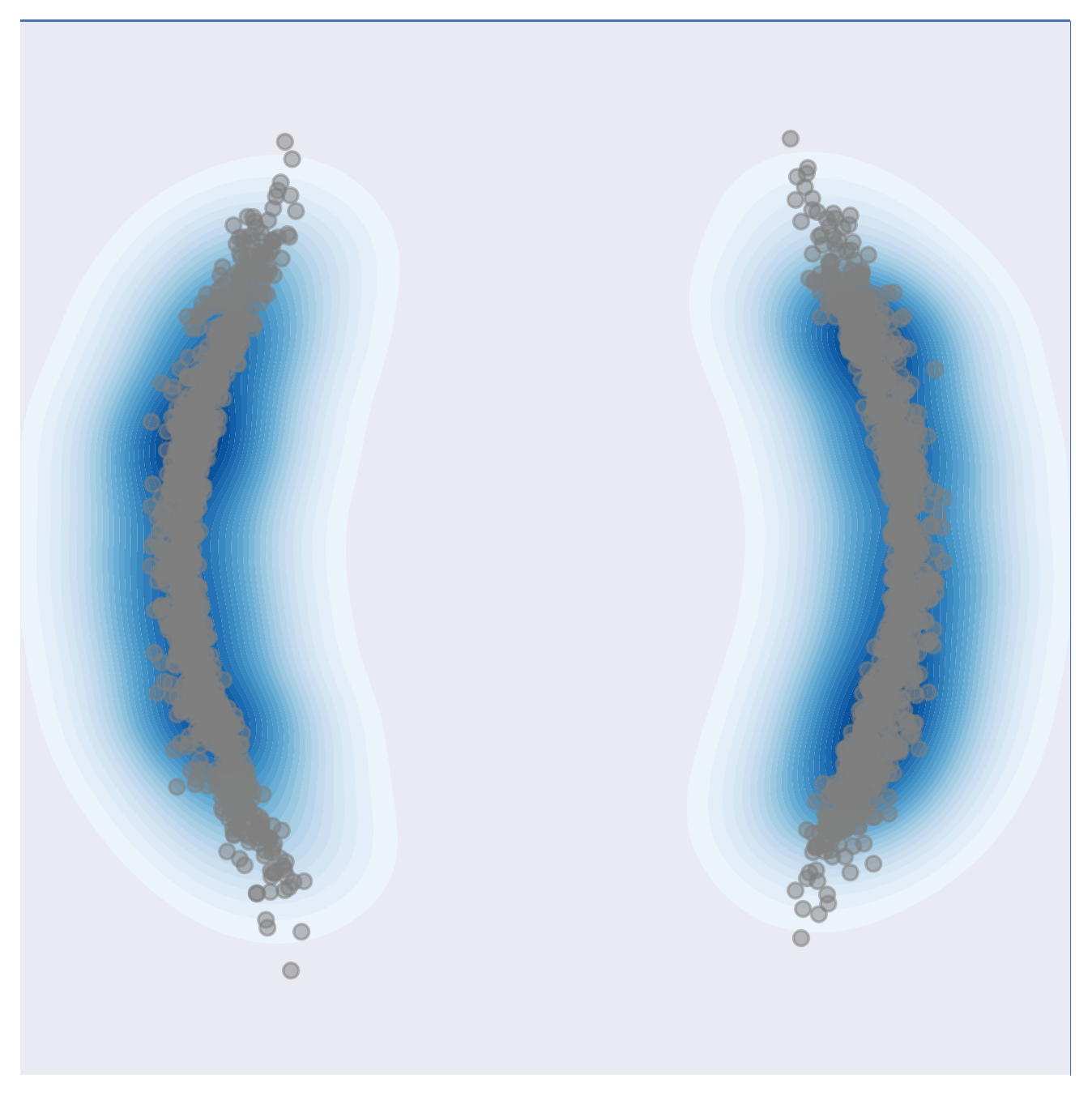}  
		&   \hspace{-5mm}
		\includegraphics[width=2cm]{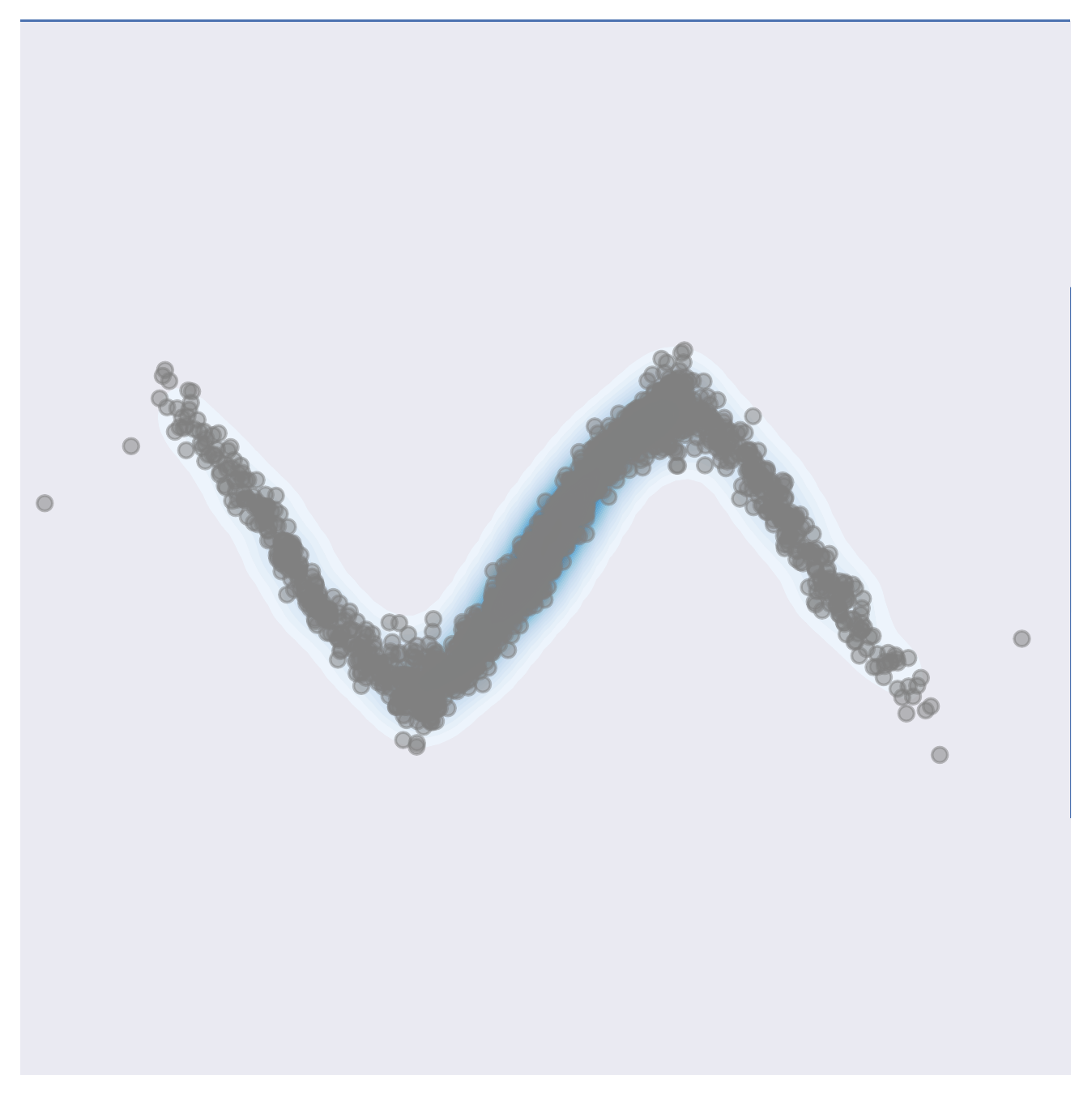} 
		& 		\hspace{-5mm}
		\includegraphics[width=2cm]{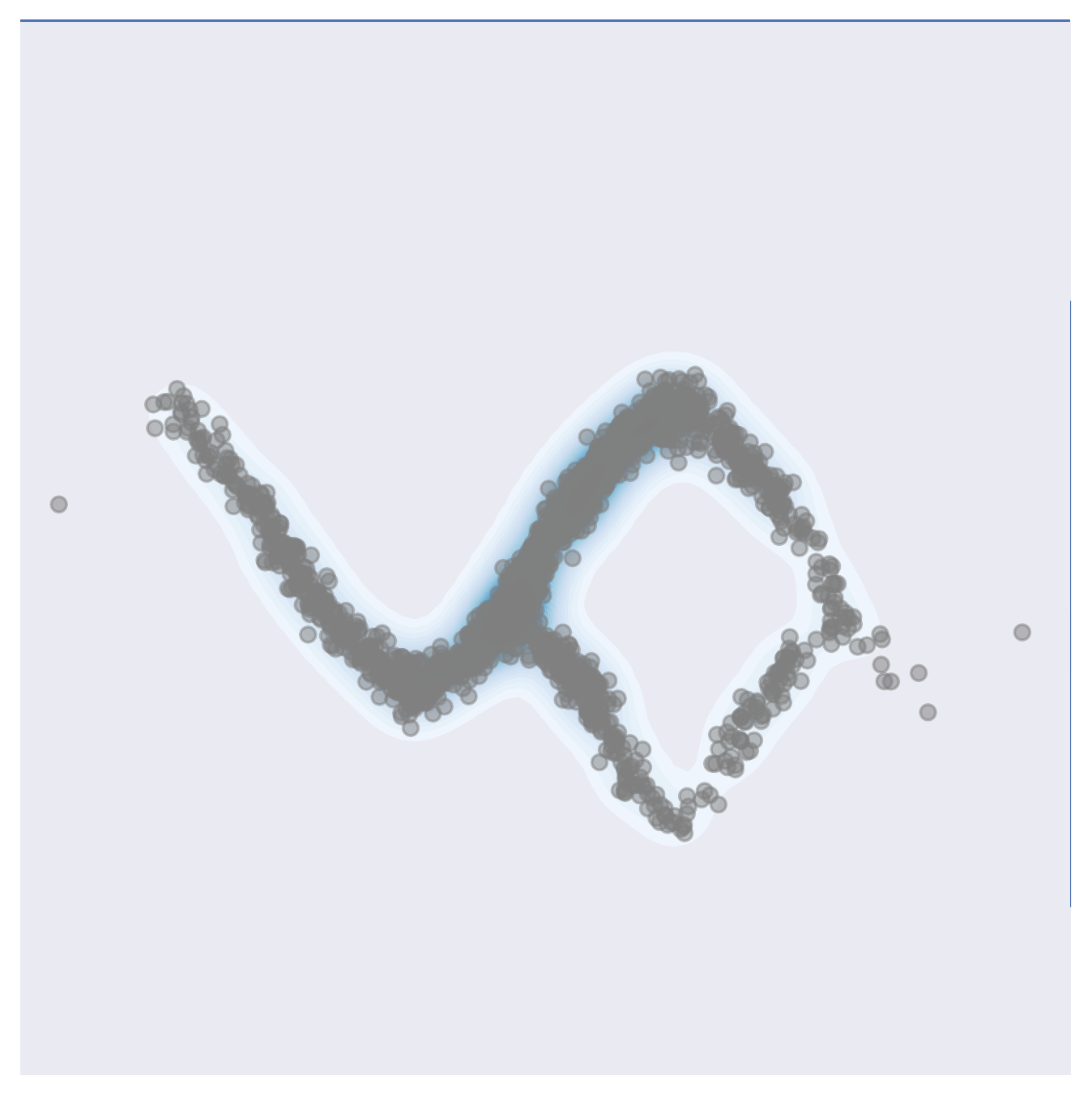}   
		&		\hspace{-5mm}
		\includegraphics[width=2cm]{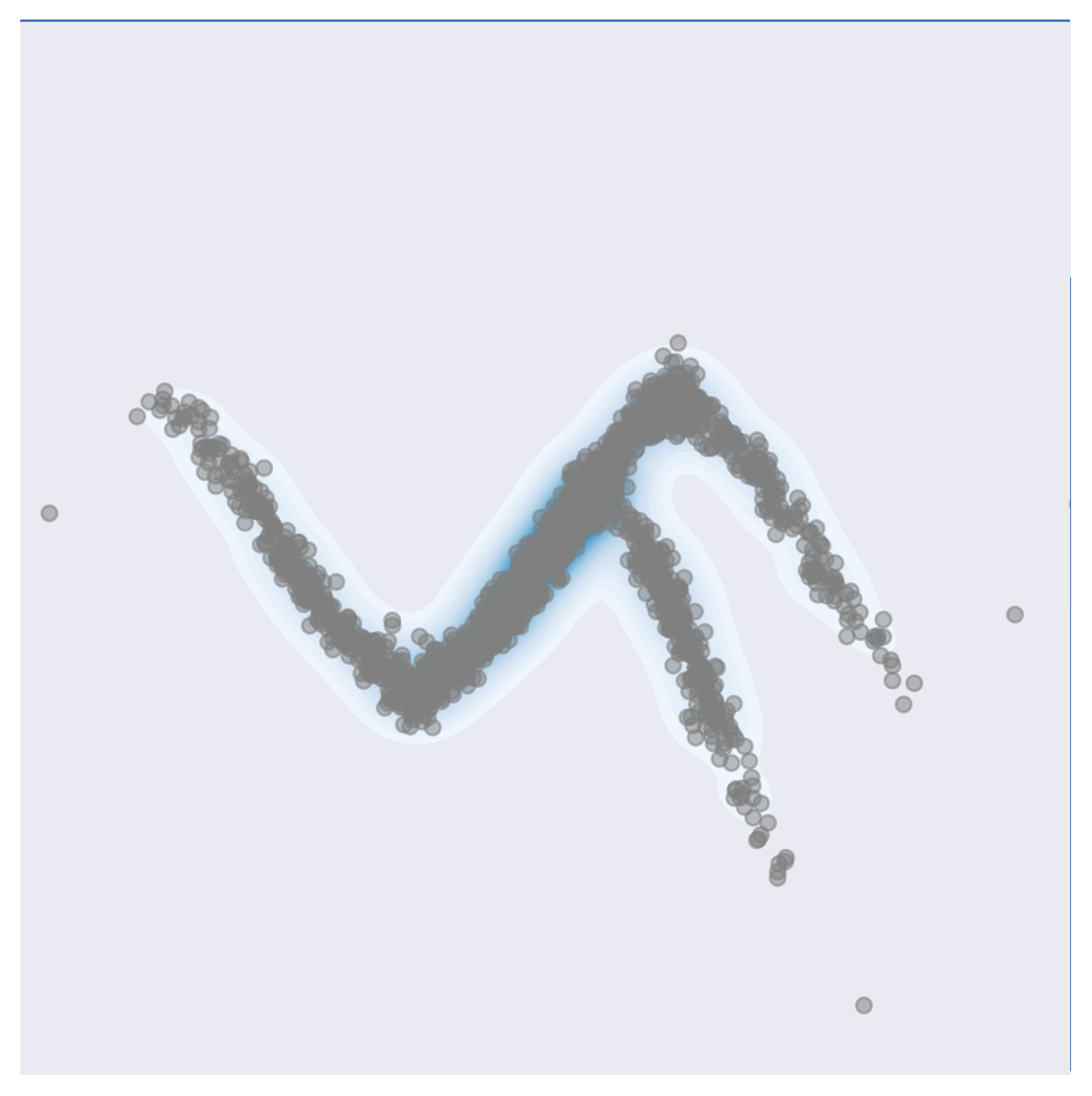}
		\vspace{-1mm}
		\\
		\hspace{-4mm}
		\includegraphics[width=2cm]{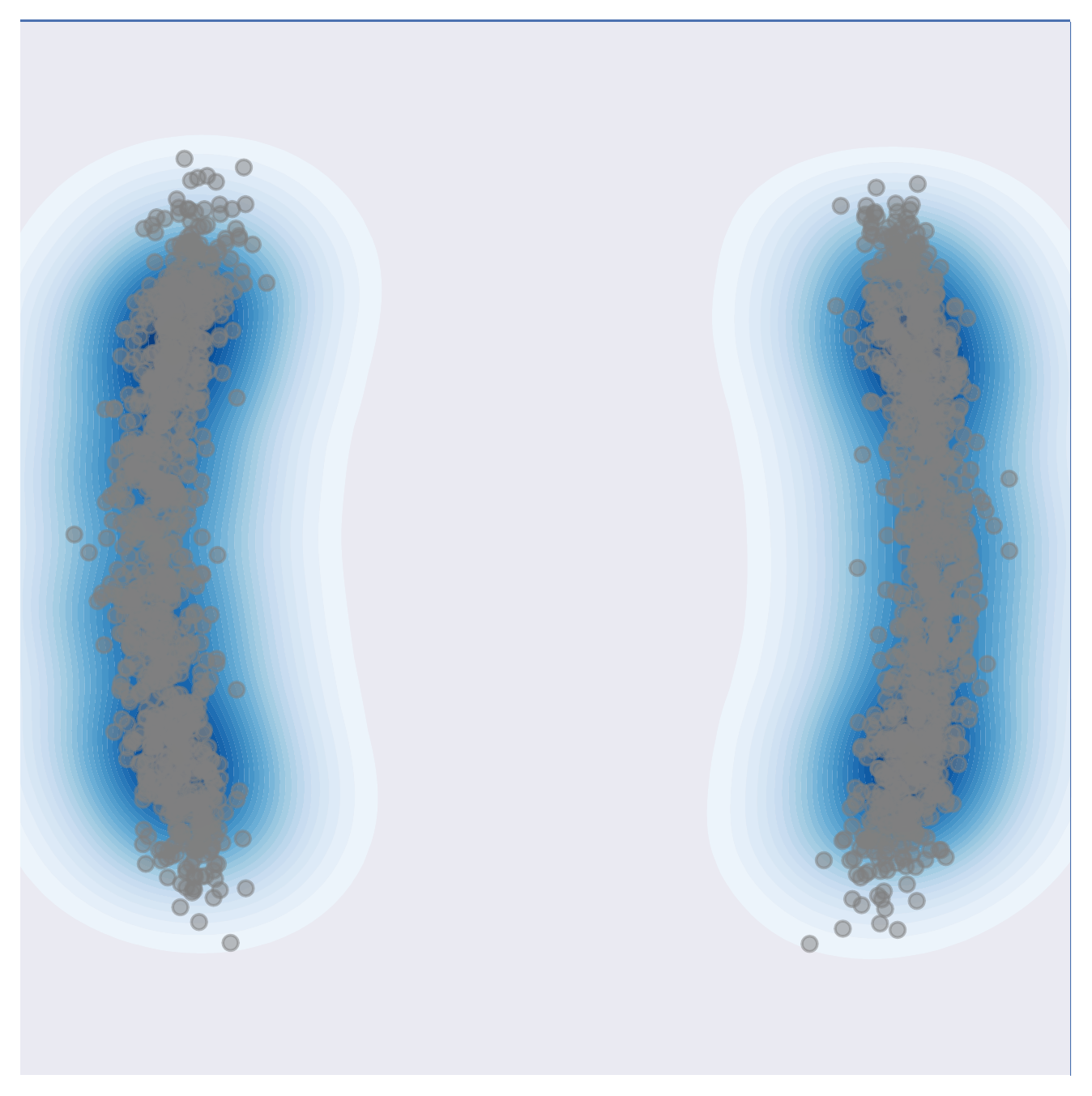}  
		&   \hspace{-5mm}
		\includegraphics[width=2cm]{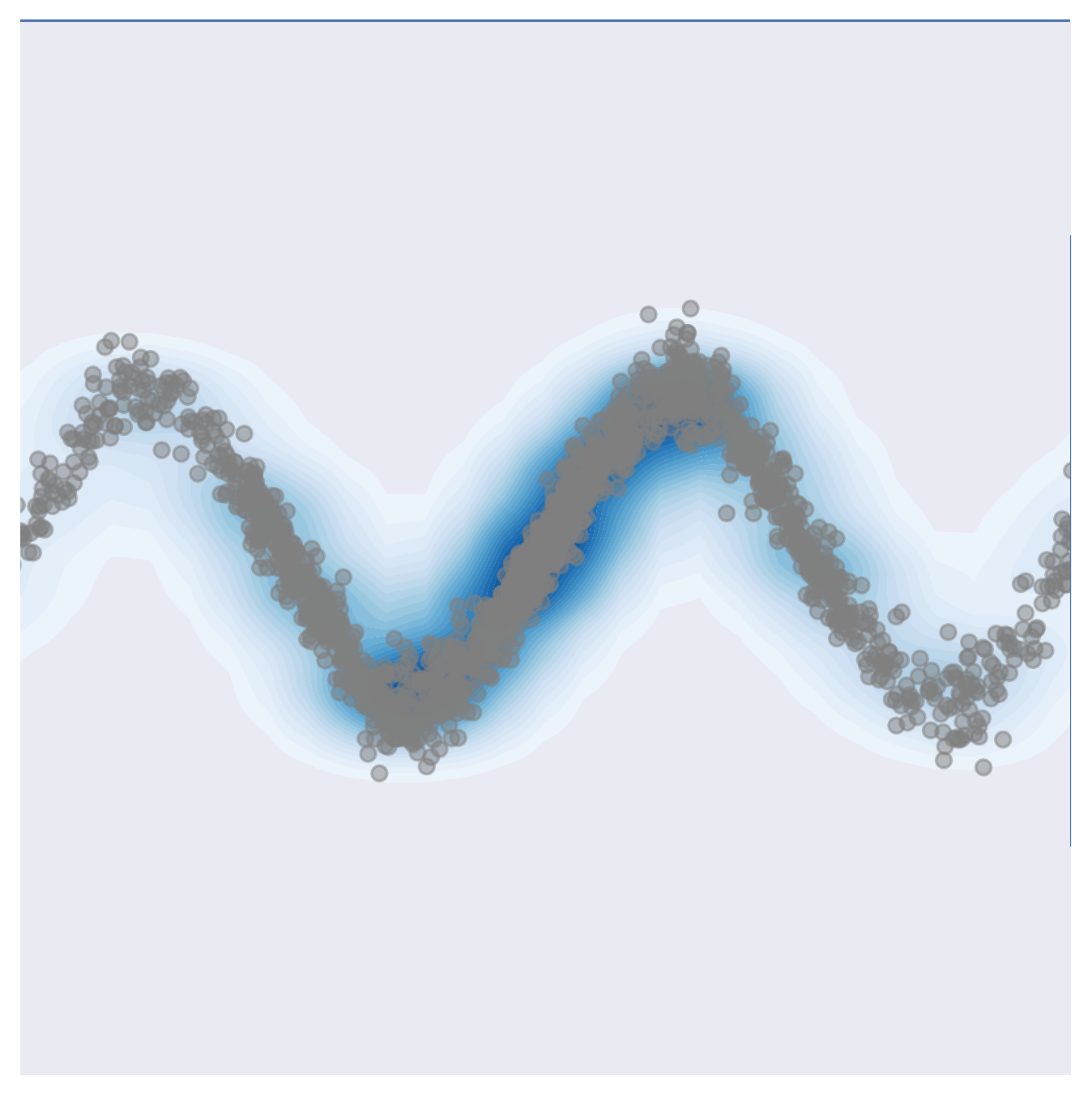} 
		& 		\hspace{-5mm}
		\includegraphics[width=2cm]{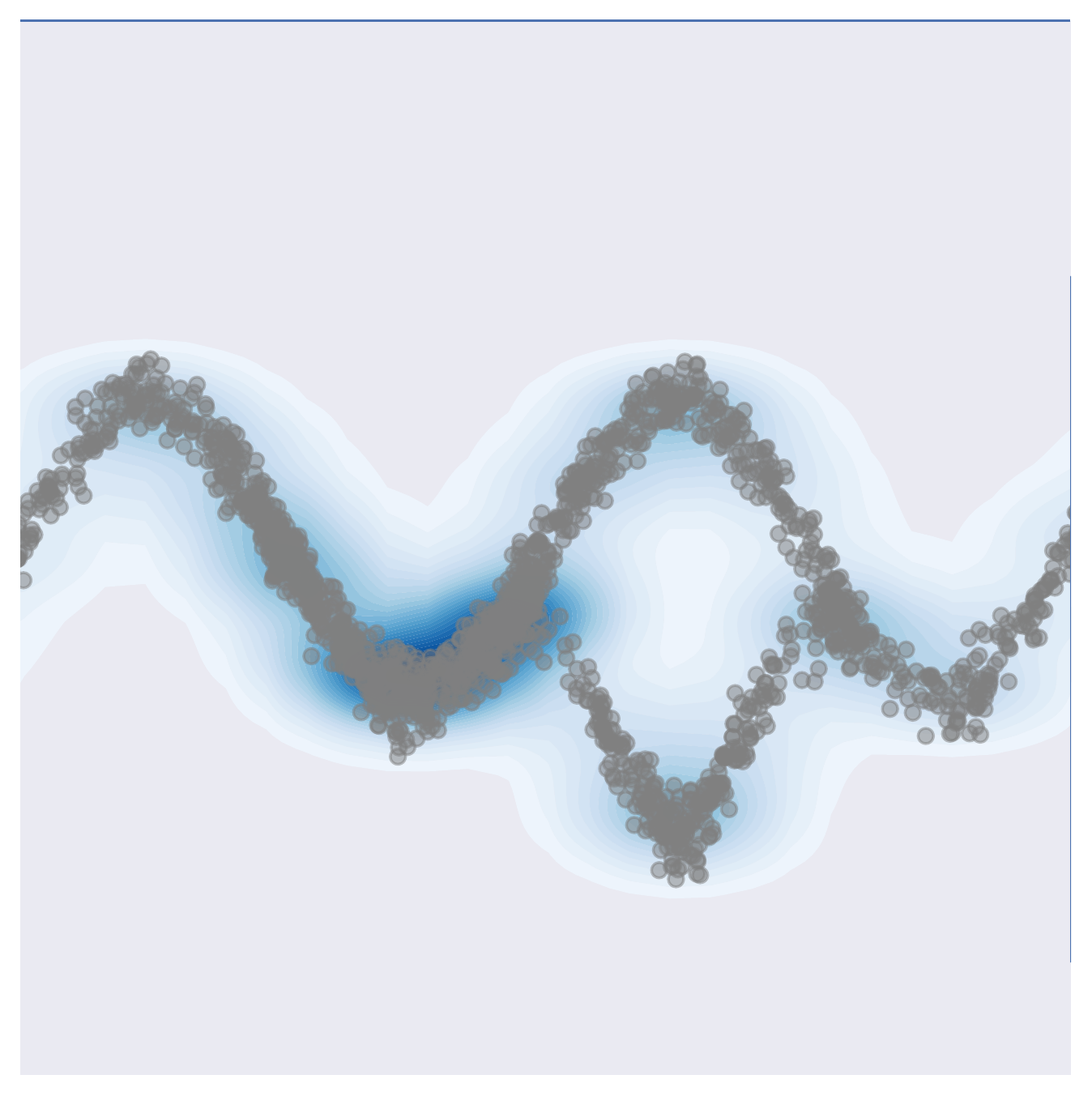}   
		&		\hspace{-5mm}
		\includegraphics[width=2cm]{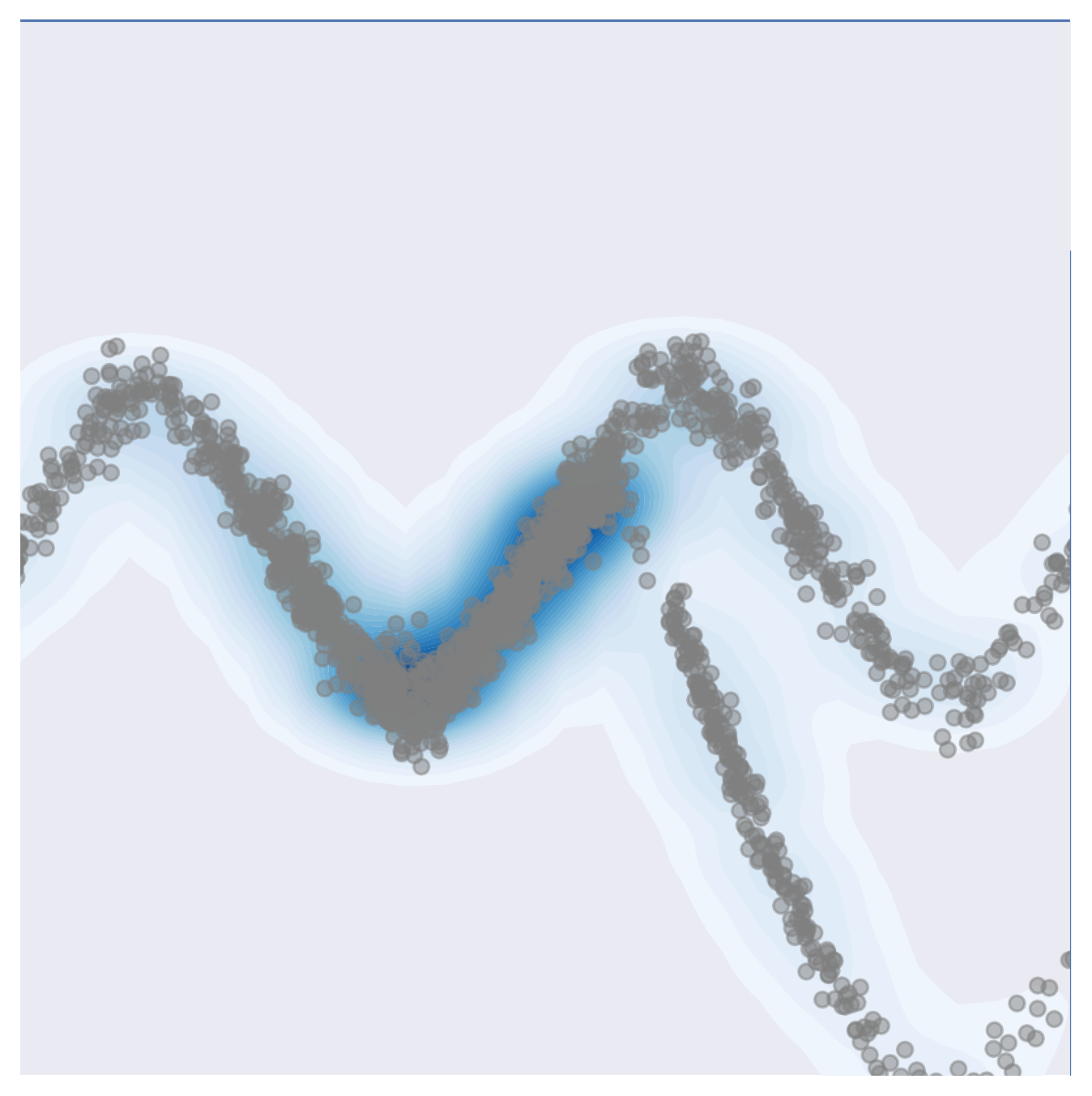}
		\vspace{-1mm}
		\\
		\hspace{-4mm}
		\includegraphics[width=2cm]{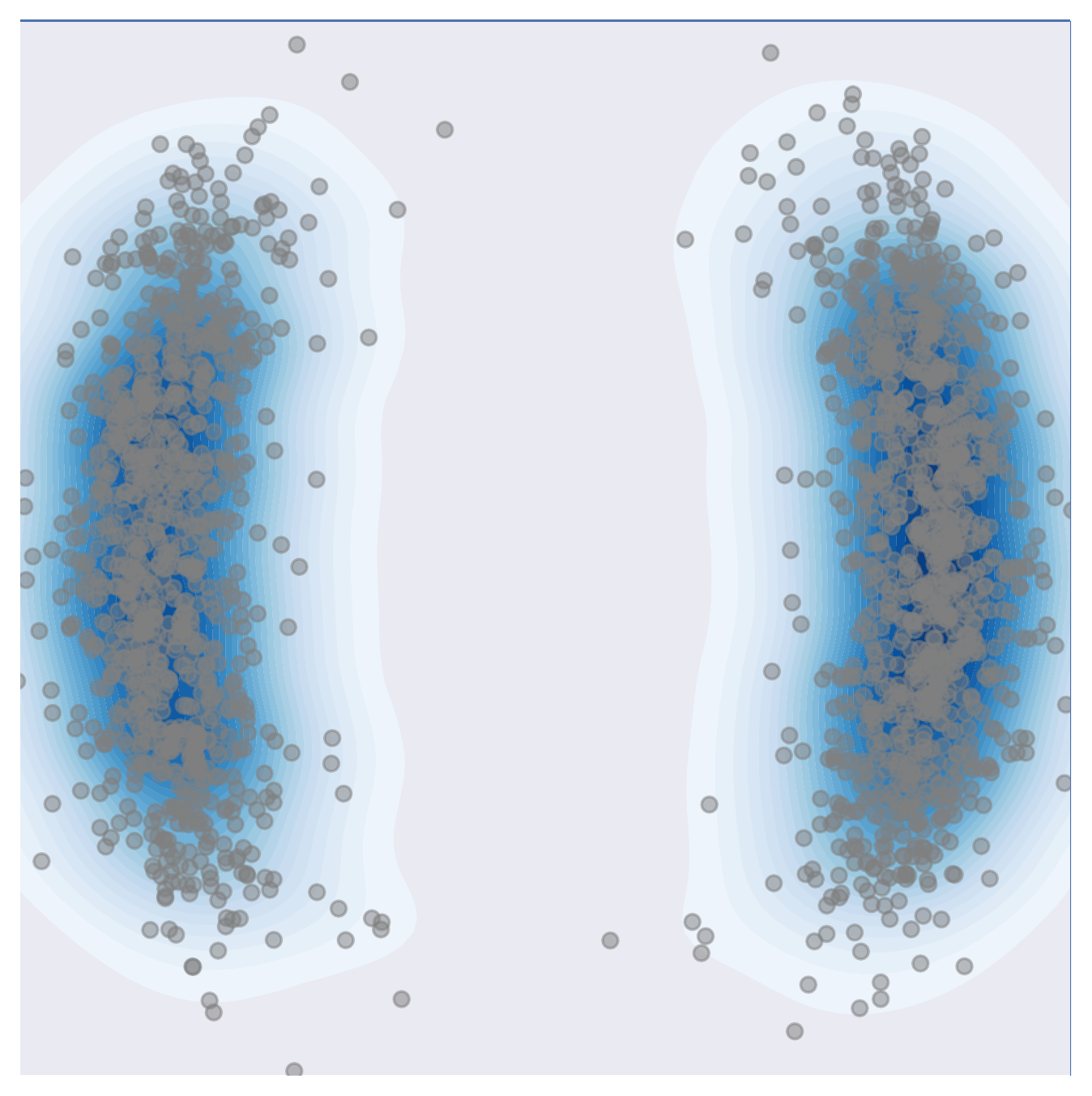}  
		&   \hspace{-5mm}
		\includegraphics[width=2cm]{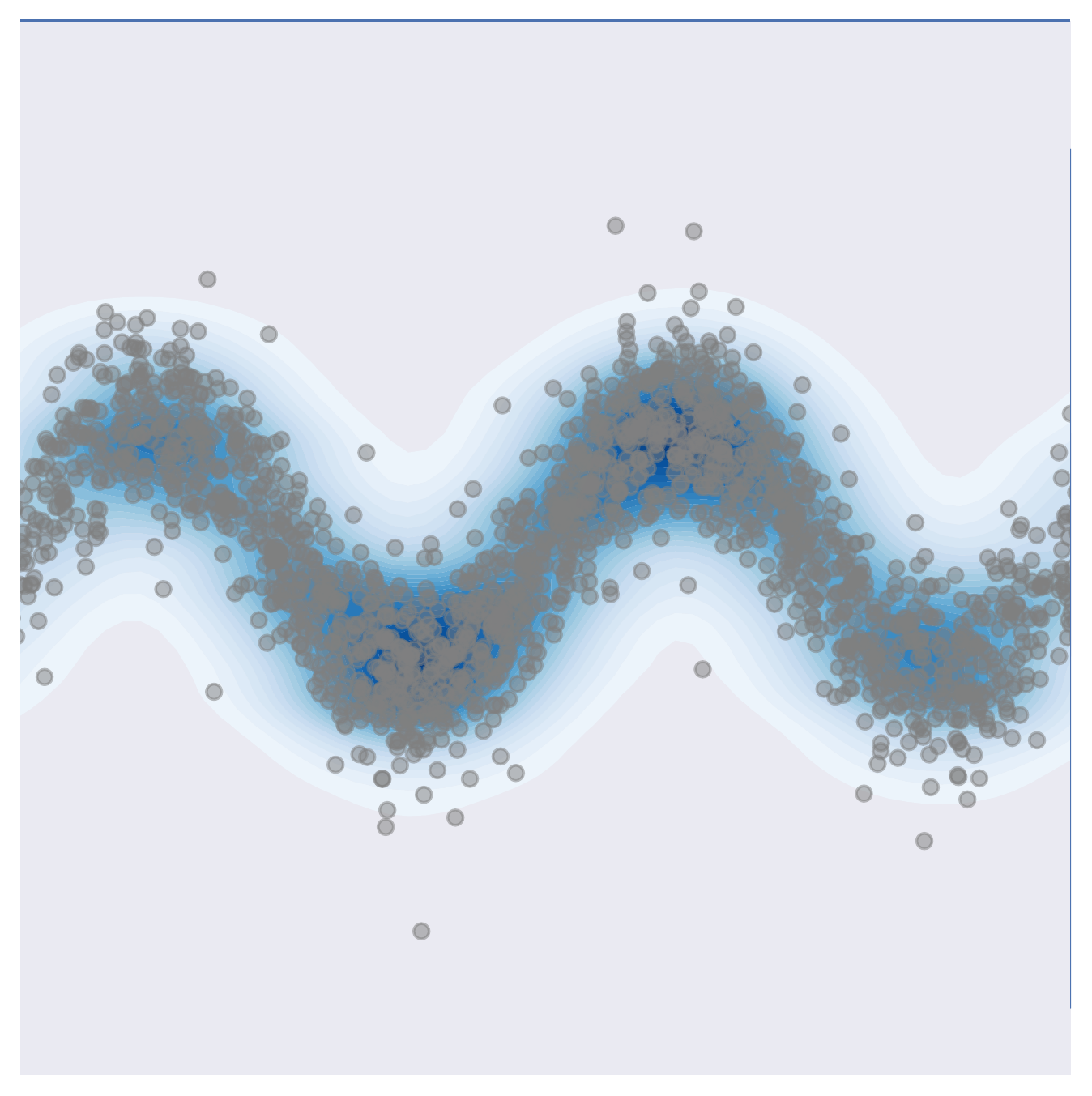} 
		& 		\hspace{-5mm}
		\includegraphics[width=2cm]{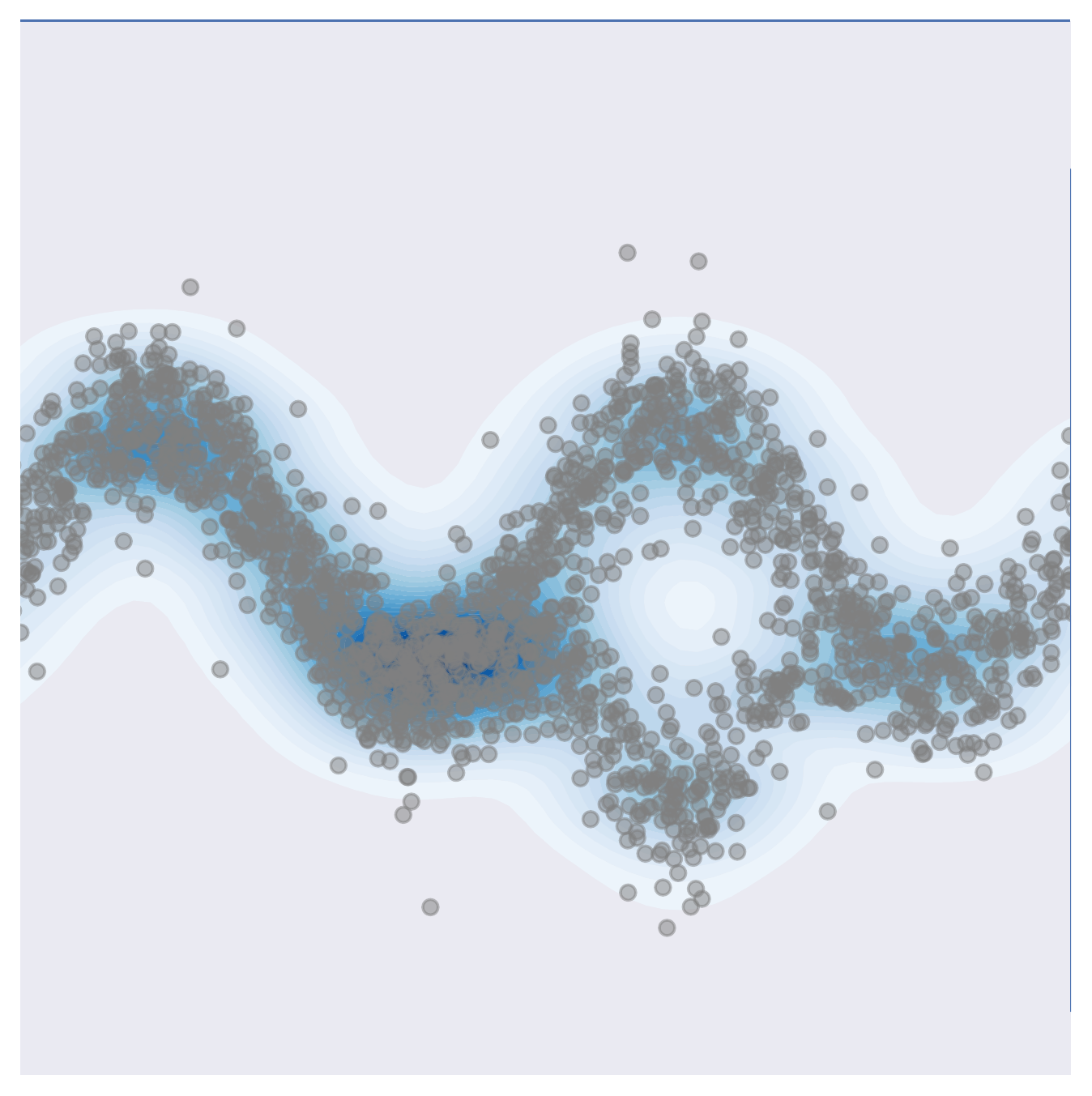}   
		&		\hspace{-5mm}
		\includegraphics[width=2cm]{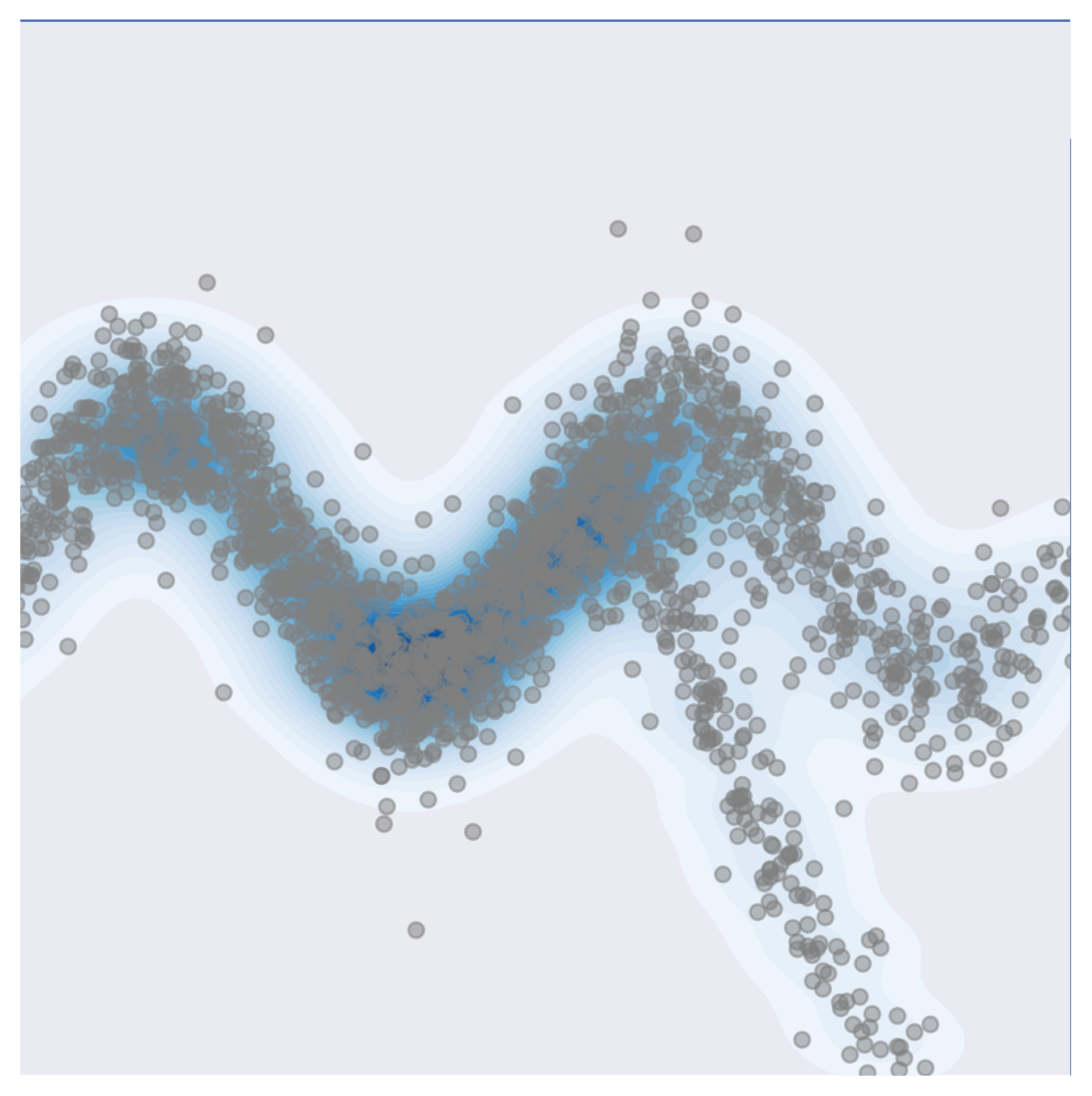}
		\vspace{-1mm}
		\\
		\hspace{-4mm}
		\includegraphics[width=2cm]{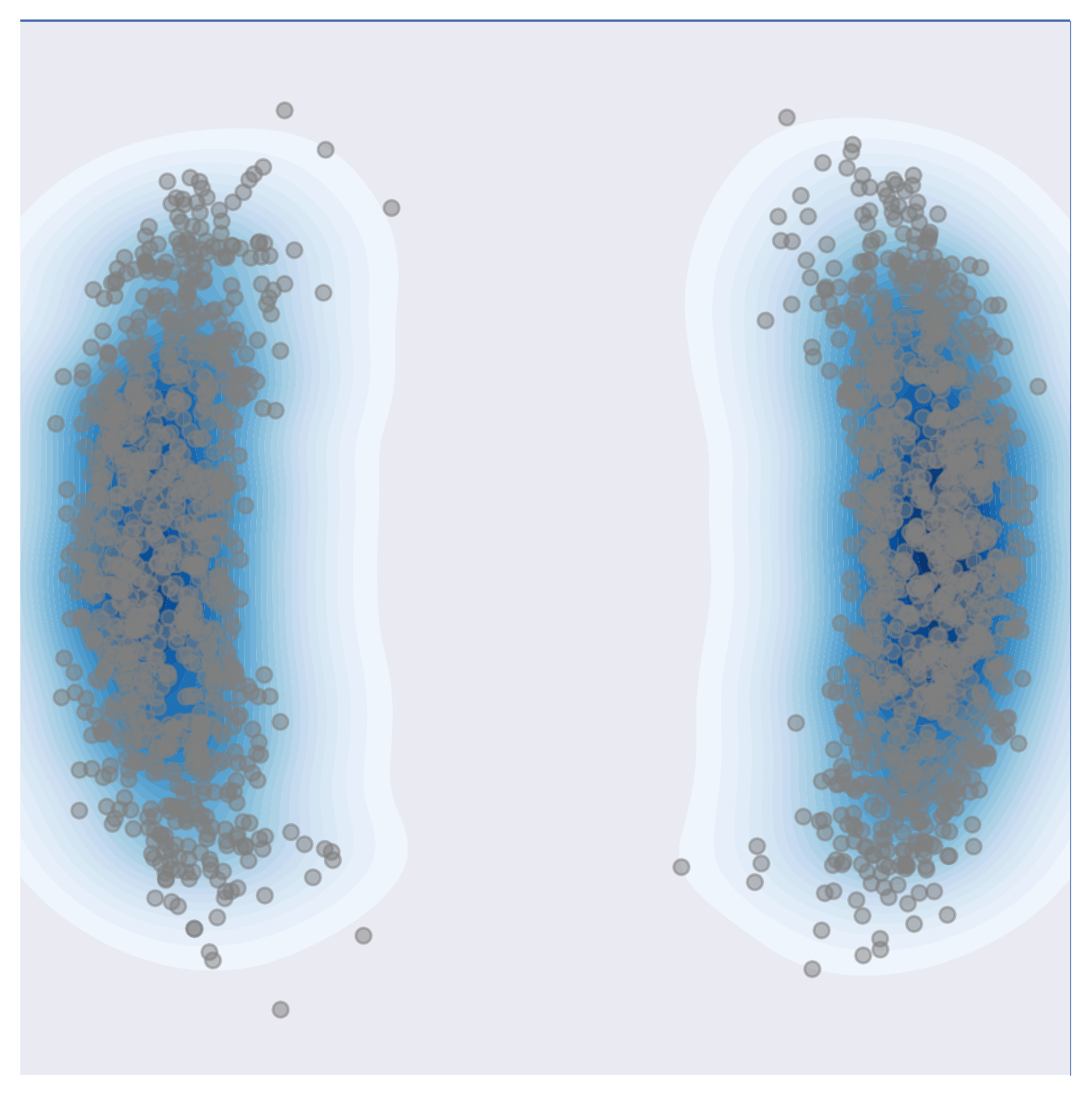}  
		&   \hspace{-5mm}
		\includegraphics[width=2cm]{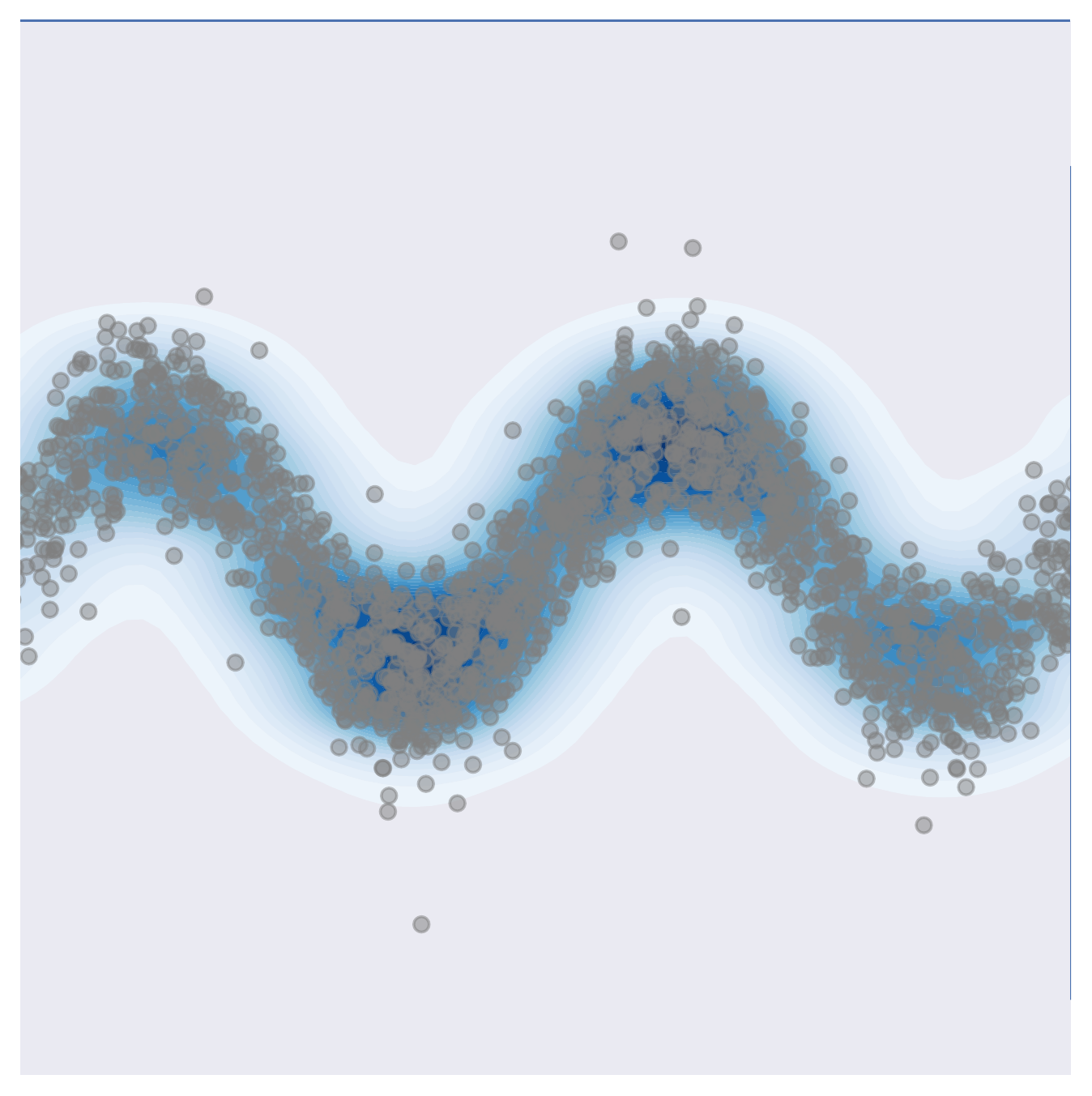} 
		& 		\hspace{-5mm}
		\includegraphics[width=2cm]{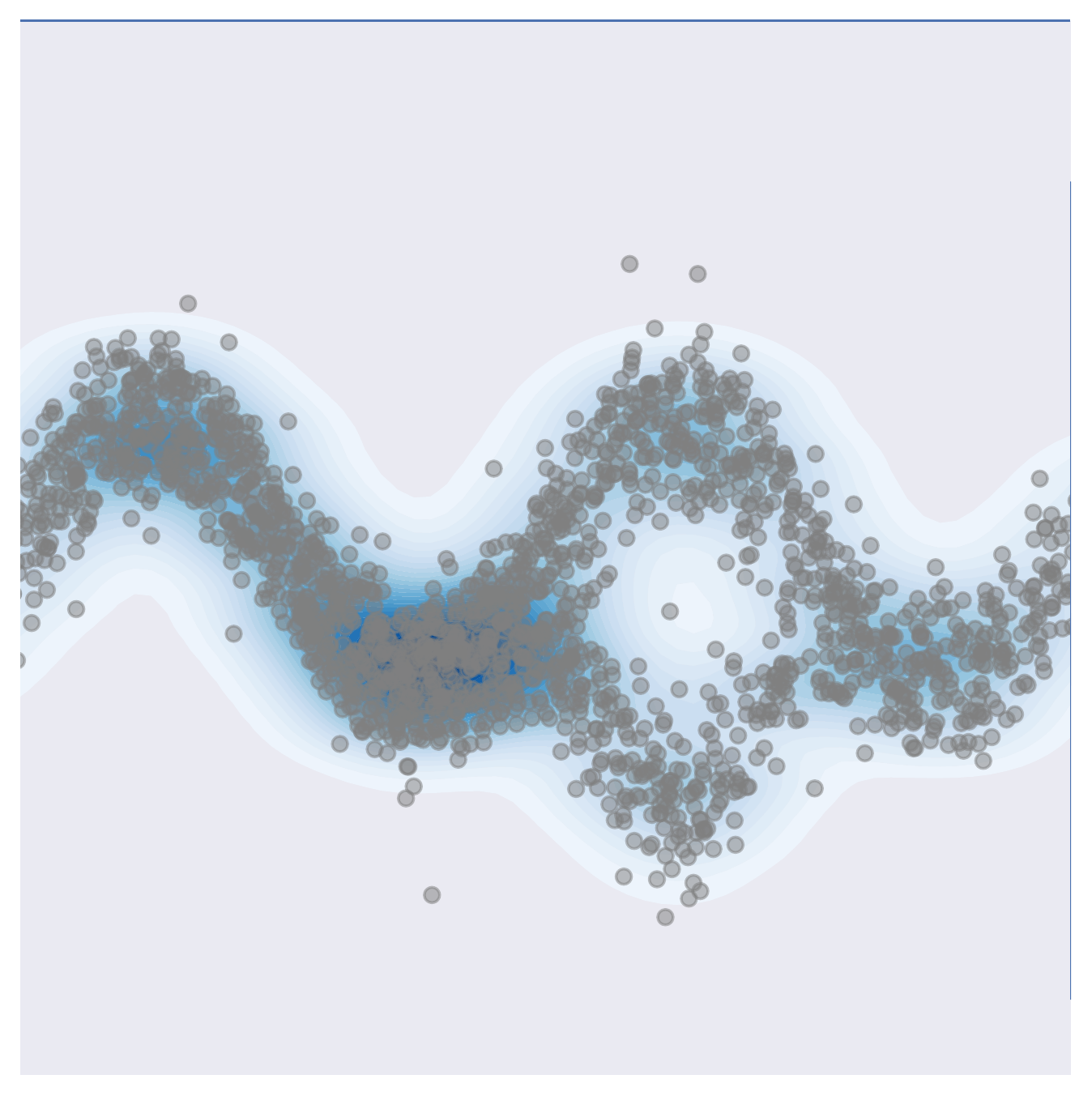}   
		&		\hspace{-5mm}
		\includegraphics[width=2cm]{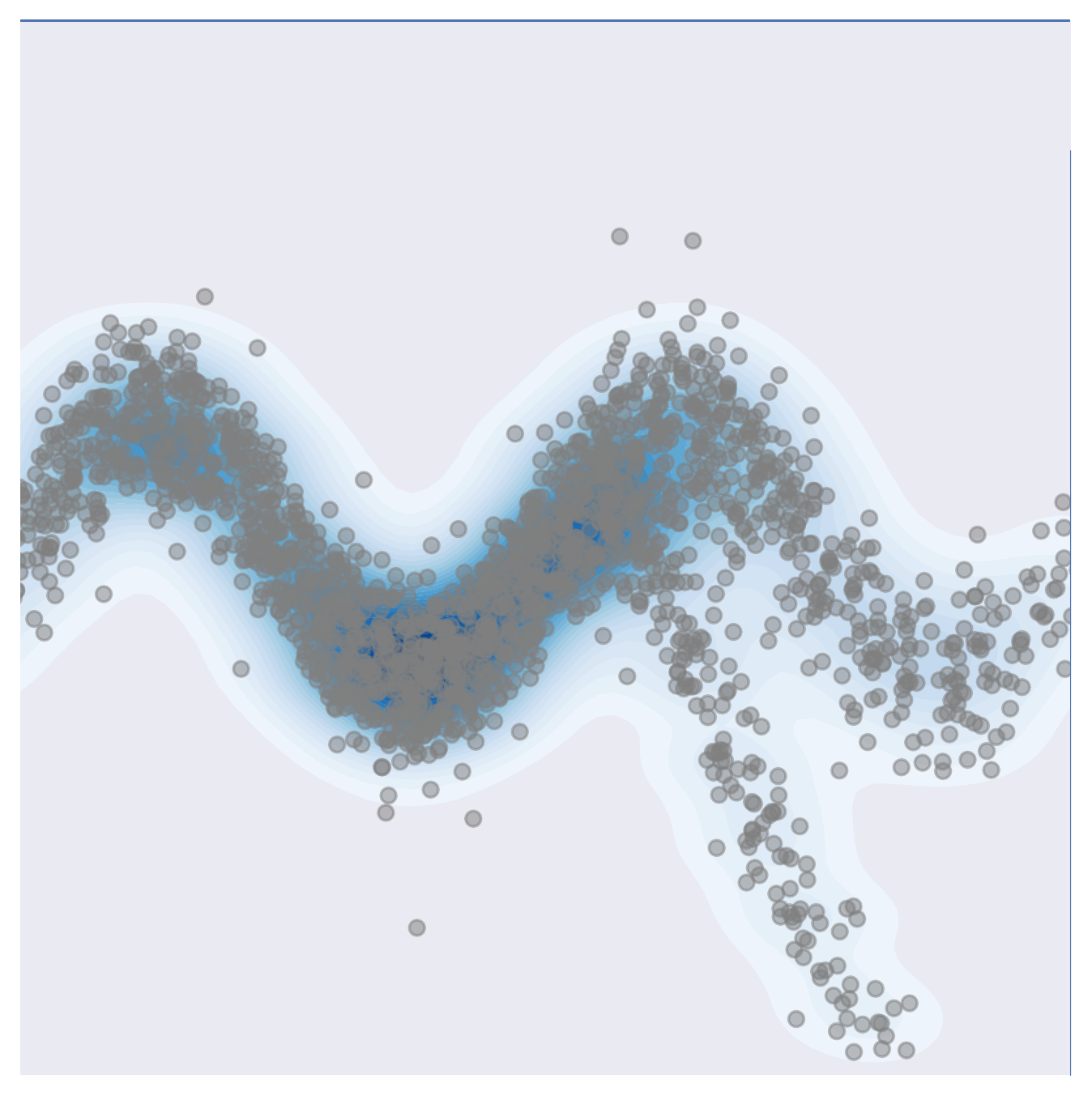} 
		\vspace{-1mm}
		\\
	\end{tabular} \vspace{-5mm}
	\caption{{\small Illustration of different algorithms on toy distributions. Each column is a distribution case. 1st row: Ground truth; 2nd row: standard SGLD; 3rd row: $w$-SGLD; 4th row: $w$-SGLD-B; 5th row: SVGD; 6th row: $\pi$-SGLD}.}
	\label{fig:2000toy}
	\vspace{-4mm}
\end{figure}

\vspace{-0.4cm}
\paragraph{Toy Distributions}
We compare various sampling methods on multi-mode toy examples, {\it i.e.}, SGLD, SVGD, $w$-SGLD, $w$-SGLD-B and $\pi$-SGLD. We aim to sample from four unnormalized 2D densities $p(z) \propto \exp\{U(z)\}$, with detailed functional form provided in the SM. We optimize/sample 2000 particles to approximate target distributions. The results are shown in Figure \ref{fig:2000toy}. It can be seen from Figure~\ref{fig:2000toy} that though SGLD maintains good asymptotic properties, it is inaccurate to approximate distributions with only a few samples; in some case, the samples cannot even cover all the modes. Interestingly, all other particle-optimization-based algorithms successfully find all the modes and fit the distributions well. $w$-SGLD is good at finding modes, but worse at modeling the correct variance due to difficulty of controlling the balance between attractive and repulsive forces between particles. $w$-SGLD-B is better than $w$-SGLD at modeling the distribution variance, performing similarly to SVGD and $\pi$-SGLD. Even though, we note that $w$-SGLD is very useful when the number of particles is small, which fits a distribution better, as shown in Section~\ref{app:ex_exp} of the SM. 

\begin{figure}
	\centering
	\includegraphics[width=\linewidth]{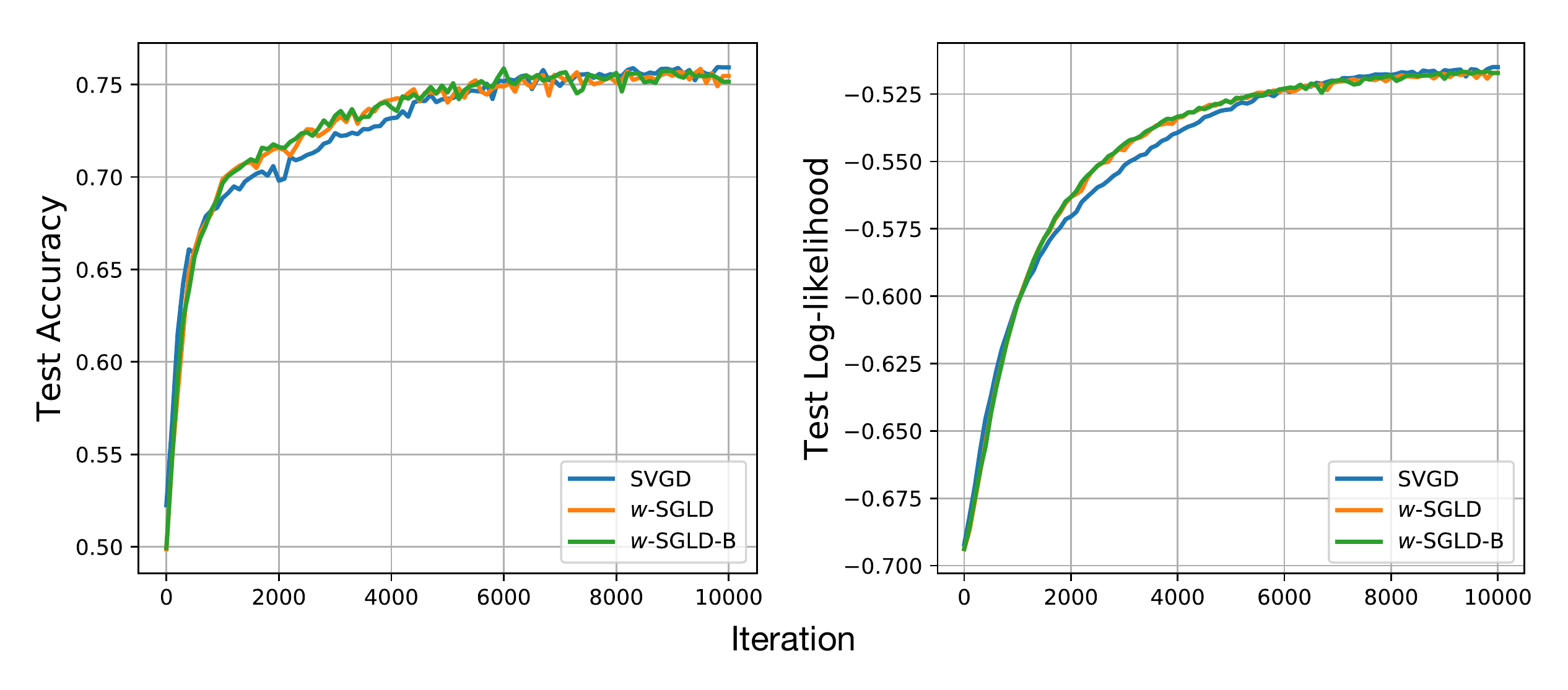}
	\vskip -0.2in
	\caption{Test accuracies (left) and log-likelihoods (right) v.s.\! iterations for SVGD, $w$-SGLD and $w$-SGLD-B.}\label{fig:lr_sgmcmc_svgd}
	\vskip -0.2in
\end{figure}

\vspace{-0.4cm}
\paragraph{Bayesian Logistic regression}
We next compare the three variants of our framework ({\it i.e.}SVGD, $w$-SGLD and $w$-SGLD-B) on a simple logistic-regression task with quantitative evaluations. We use the same model, data and experimental settings as \cite{LiuW:NIPS16}. The Covertype dataset contains 581,012 data points and 54 features. We perform 5 runs for each setting and report the mean of testing accuracies/log-likelihoods. Figure~\ref{fig:lr_sgmcmc_svgd} plots both test accuracies and test log-likelihoods w.r.t.\! the number of training iterations. It is clearly that while all methods converge to the same accuracy/likelihood level, both $w$-SGLD and $w$-SGLD-B converge slightly faster than SVGD. In addition, $w$-SGLD and $w$-SGLD-B have similar convergence behaviors, thus we only use $w$-SGLD in the DNN experiments below.
\begin{wrapfigure}{R}{5.5cm}\vspace{-0.5cm}
	\centering
	\hspace{-0.0cm}\includegraphics[width=\linewidth]{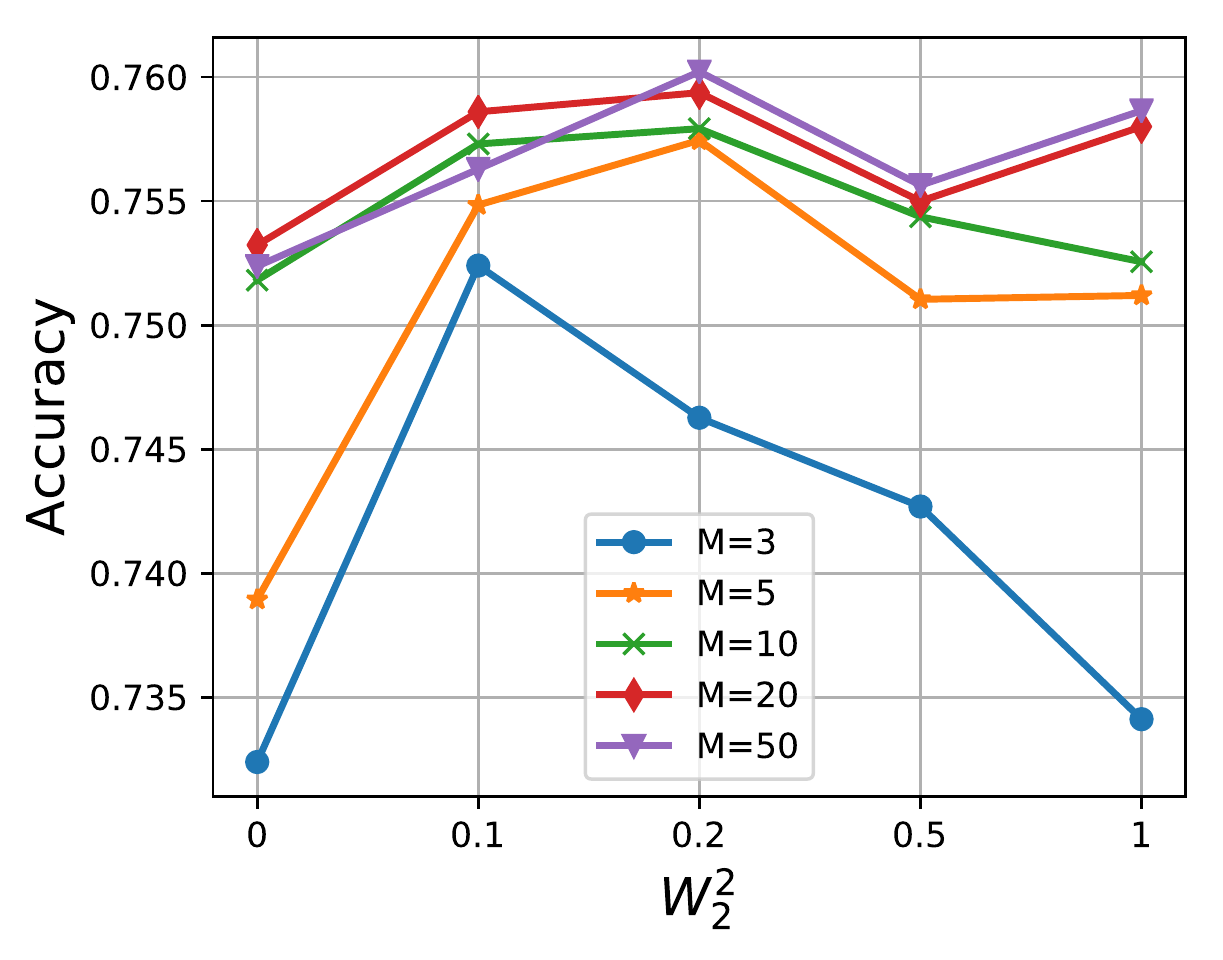}
	\vskip -0.15in
	\caption{\small Impact of $W_2^2$ factor $\gamma$ and particle number $M$.}\label{fig:hyper}
	\vskip -0.2in
\end{wrapfigure}
\vspace{-0.2cm}
\paragraph{Parameter Sensitivity}
Now we study the role of hyperparameters in $\pi$-SGLD: the number of  particles $M$ and the scaling factor $\gamma$ to replace the $u_iv_j$-term in \eqref{eq:gradf2}. We use the same dataset and model as the above experiment. Figure \ref{fig:hyper} plots test accuracies along with different parameter settings. As expected, the best performance is achieved with appropriate scale of $W^2_2$. The performance keep improving with increasing particles. Interestingly, the Wasserstein regularization is more important when the number of particles is small, demonstrating the superiority when approximate distributions with very few particles. 

\vspace{-0.3cm}
\subsection{Applications on deep neural networks}
\vspace{-0.3cm}
We conduct experiments for Bayesian learning of DNNs. Different from traditional optimization for DNNs, we are interested in modeling weight uncertainty of neural networks, an important topic that has been well explored \citep{HernaandezLobatoA:icml15,BlundellCKW:icml15,PSGLD:AAAI16,LouizosW:ICML16}. We assign priors to the weights, which are simple isotropic Gaussian priors in our case, and perform posterior sampling with the proposed particle-optimization-based algorithms, as well as other standard algorithms such as SGLD and SGD. We use the RMSprop optimizer for feed-forward networks (FNN), and Adam for for convolutional neural networks (CNNs) and recurrent neural networks (RNNs).
For all methods, we use a RBF kernel $K(\thetab,\thetab') = \exp(-\|\thetab-\thetab^\prime\|_2^2/h)$, with the bandwidth set to $h=\mathtt{med}^2/\log M$. Here $\mathtt{med}$ is the median of the pairwise distance between particles.
All experiments are conducted on a single TITAN X GPU.
\vspace{-0.4cm}
\paragraph{Feed-forward Neural Networks}
\vspace{-2mm}
We perform the classification tasks on the standard MNIST dataset.
A two-layer model 784-X-X-10 with $\ReLU$ activation function  is used, with X being the number of hidden units for each layer. The training epoch is set to 100. The test errors are reported in Table \ref{Table:FNN}. Not surprisingly, Bayesian methods generally perform better than their optimization counterparts. The new $\pi$-SGLD which combines $w$-SGLD and SVGD improves both methods with little computational overhead. In additional, $w$-SGLD seems to perform better than SVGD in this case, partially due to a better asymptotic property mentioned in \citep{liu2017stein_flow}. Furthermore, standard SGLD which is based on MCMC obtains higher test errors compared to particle-optimization-based algorithms, partially due to the correlated-sample issue discussed in the introduction. See~\citep{blundell2015weight} for details on the other methods in Table~\ref{Table:FNN}.
\begin{table}[h]
	\centering
	\vskip -0.15in
	\caption{Classification error of FNN on MNIST.}
	\vskip 0.1in
	\begin{adjustbox}{scale=1,tabular=llccc,center}
		\specialrule{.1em}{.05em}{.05em} 
		& \multicolumn{2}{c}{Test Error}  \\
		Method & 400-400 & 800-800 \\
		\hline
		$\pi$-SGLD & $\mathbf{1.36\%}$ & $\mathbf{1.30\%}$ \\
		$\w$-SGLD & 1.44\% & 1.37\%\\
		SVGD  & 1.53\% &  1.40\% &   \\
		SGLD	    	 &  1.64\% &    1.41\% \\
		\hline
		RMSprop  	   &  1.59\% &    1.43\%  \\
		RMSspectral 	   		  &  1.65\% &    1.56\%  \\
		SGD 	   		  &   1.72\% &   1.47\%   \\
		\hline
		BPB, Gaussian   &  1.82\% & 1.99\%     \\
		SGD, dropout  & 1.51\%  & 1.33\%  \\
		\specialrule{.1em}{.05em}{.05em} 
	\end{adjustbox}
	\label{Table:FNN}
	\vspace{-0.6cm}
\end{table}
\vspace{-0.4cm}
\paragraph{Convolution Neural Networks}
We use the CIFAR-10 dataset to test our framework on CNNs. We adopt a CNN of three convolution layers, using 3$\times$3 filter size with C64-C128-C256 channels, and 2$\times$2 max-pooling after each convolution layer. 
Our implementation adopts batch normalization, drop out and data augmentation to improve the performance. 
Training losses and test accuracies are presented in Table~\ref{Table:CNN}. Consistently, $\pi$-SGLD outperforms all other algorithms in terms of test accuracy. ADAM obtains a better training loss but worse test accuracy, indicating worse generalization ability of the optimization-based methods compared to Bayesian methods. 
\begin{table}[h]
	\centering
	\vskip -0.15in
	\caption{Classification error of CNN on CIFAR-10.}
	\vskip 0.1in
	\begin{adjustbox}{scale=1,tabular=lccc,center}
		\specialrule{.1em}{.05em}{.05em} 
		Method & Training Loss &Test Accuracy \\
		\hline
		ADAM &  23.80 &    86.76\%\\
		SVGD  & 30.57 &  88.72\% &   \\
		SGLD   &  28.52 &    88.64\% \\
		$\w$-SGLD &  31.26 & 88.80\% \\
		$\pi$-SGLD & 25.06 & $\mathbf{89.52\%}$ \\
		\specialrule{.1em}{.05em}{.05em} 
	\end{adjustbox}
	\label{Table:CNN}
	\vspace{-10pt}
\end{table}
\vspace{-0.5cm}
\paragraph{Recurrent Neural Networks}
For RNNs, we run standard language models. Experiments are presented on three publicly available corpora: APNEWS, IMDB and BNC. APNEWS is a collection of Associated Press news articles from 2009 to 2016. IMDB is a set of movie reviews collected by \cite{maas2011learning}, and BNC~\cite{BNCConsortium2007} is the written portion of the British National Corpus, which contains excerpts from journals, books, letters, essays, memoranda, news and other types of text. These datasets can be downloaded from Github\footnote{https://github.com/jhlau/topically-driven-language-model}.

\begin{table}[!htbp]
	\vspace{-0.4cm}
	\caption{Perplexity of language model on three corpora. }
	\begin{center}
		\scalebox{1}{
			\begin{tabular}{c c c c }
				\specialrule{.1em}{.05em}{.05em} 
				Method& APNEWS & IMDB &BNC \\ \hline
				SGD & 64.13& 72.14 & 102.89 \\
				SGLD & 63.01 & 68.12 & 95.13 \\ 
				SVGD & 61.64 & 69.25& 94.99\\
				$\w$-SGLD & 61.22 & 67.41 & 93.68 \\
				$\pi$-SGLD & $\textbf{59.83}$ & \textbf{67.04} & \textbf{92.33}\\
				\specialrule{.1em}{.05em}{.05em} 
		\end{tabular}}
		\label{Table:RNN}
	\end{center}
	\vspace{-0.4cm}
\end{table} 
We follow the standard set up as \cite{wang2017topic}. Specifically, we lower case all the word tokens and filter out word tokens that occur less than 10 times. All the datasets are divided into training, development and testing sets. For the language model set up, we consider a 1-layer LSTM model with 600 hidden units. The sequence length is fixed to be 30. In order to alleviate overfitting, dropout with a rate of 0.4 is used in each LSTM layer. Results in terms of test perplexities are presented in Table~\ref{Table:RNN}. Again, we see that $\pi$-SGLD performs best among all algorithms, and $w$-SGLD is slightly better than SVGD, both of which are better than other algorithms.

\vspace{-0.5cm}
\section{Conclusion}
\vspace{-0.4cm}
We propose a unified particle-optimization framework for large-scale Bayesian sampling. Our framework defines gradient flows on the space of probability measures, and uses particles to approximate the corresponding densities. Consequently, solving gradient flows reduces to optimizing particles on the parameter space. Our framework includes the standard SVGD as a special case, and also allows us to develop efficient particle-optimization algorithms for SG-MCMC, which is highly related to SVGD. Extensive experiments are conducted, demonstrating the effectiveness and efficiency of our proposed framework. Interesting future work includes designing more practically efficient variants of the proposed particle-optimization framework, and developing theory to study general convergence behaviors of the algorithms, in addition to the asymptotic results presented in \citep{liu2017stein_flow}.

{\small\bibliographystyle{apalike}
\bibliography{reference}}

\clearpage
\appendix

\section{Proof of Proposition~2}
\begin{proof}
	A stationary distribution $\mu$ of \eqref{eq:vlasov}  means $\partial_{t}\mu = 0$. Assuming $\mu = p(\thetab|\Xb) \triangleq p$, then we need to prove that
	\begin{align*}
		\nabla \cdot \left((\Wb*p)p\right) = 0~.
	\end{align*}
	By the definition of $\Wb$ in \eqref{eq:W}, and applying Stein's identity \cite{LiuW:NIPS16}, we have $\Wb*p = 0$. Consequently, we have $\nabla \cdot \left((\Wb*p)p\right) = 0$.
	
	The above argument indicates $p(\thetab|\Xb)$ is a stationary distribution of \eqref{eq:vlasov}. This completes the proof.
\end{proof}

\section{More Details on Lemma~3}\label{app:lemma3}

We first specify the conditions the energy functional $E$ needs to satisfy in Assumption~\ref{ass:energy}.
\begin{assumption}\label{ass:energy}
	The energy functional is assumed to
	\begin{itemize}
		\item {\em proper}: $D(E) \triangleq \{\thetab\in\Omega: E(\thetab) < +\infty\} \neq \emptyset$.
		\item {\em coercive}: There exists $\tau_0 > 0$, $\thetab_{0} \in \Omega$ such that $\inf\left\{\frac{1}{2\tau_0}W_2^2(\thetab_{0}, \vb) + E(\vb): \vb \in\Omega \right\} > -\infty$.
		\item {\em lower semicontinuous}: For all $\thetab_n, \thetab \in \Omega$ such that $\thetab_n\rightarrow\thetab$, $\lim\inf_{n\rightarrow\infty}E(\thetab_n) \geq E(\thetab)$.
		\item {\em convex}: $E$ is convex in the sense that given $\lambda\in\mathbb{R}$ and a curve $\thetab_{\alpha}\in\Omega$,
		\begin{align*}
			E(\thetab_{\alpha}) \leq (1 - \alpha)E(\thetab_{0}) + \alpha E(\thetab_{1})~.
		\end{align*}
	\end{itemize}
\end{assumption}

\begin{proof}[Sketch proof of Lemma~\ref{lem:discrete_error}]
	Our case is just a simplification of Theorem~3.5.1 in \cite{Craig:thesis14}, where we restrict the energy functional to be convex instead the more general case of $\lambda$-convex. For $\lambda$-convex energy functional, \cite{Craig:thesis14} proves that $W_2^2(\tilde{\mu}_{T/h}, \mu_T) \leq \sqrt{3}|\partial E|(\mu)e^{3\lambda^{-}T}\sqrt{Th}$ with $\lambda^{-} \triangleq \max\{0, -\lambda\}$. Our result follows by simply letting $\lambda^{-} = 0$, which is for the case of convex $E$.
\end{proof}


\section{Derivation of (22)}\label{supp:particle_derivation}

The derivation of \eqref{eq:velocity} relies on the following Lemma from \cite{CarrilloCP:arxiv17}.
\begin{lemma}[Proposition~3.12 in \cite{CarrilloCP:arxiv17}]
	Let $F: (0, \infty) \rightarrow\mathbb{R}$ belongs to $C^2(0, +\infty)$ and satisfy $\lim_{s\rightarrow+\infty}F(s) = +\infty$ and $\lim\inf_{s\rightarrow 0}F(s)/s^{\beta} > -\infty$ for some $\beta > -2/(d+2)$. Define
	\begin{align*}
		\mathcal{F}(\mu) \triangleq \int F \circ (K*\mu)\mathrm{d}\mu~,
	\end{align*}
	where $\circ$ denotes function composition, {\it i.e.}, $F$ is evaluated on the output of $K*\mu$. Then we have
	\begin{align}
		\nabla\frac{\delta \mathcal{F}}{\delta \mu} = \nabla\varphi_{\epsilon} * \left(F^\prime \circ (\varphi_{\epsilon} * \mu)\mu\right) + \left(F^\prime\circ(\varphi_{\epsilon} * \mu)\right) \nabla\varphi_{\epsilon} * \mu~.
	\end{align}
\end{lemma}

Now it is ready to derive \eqref{eq:velocity}. In this case, $F = \log(\cdot)$. Let $F_1 = \nabla\varphi_{\epsilon} * \left(F^\prime \circ (\varphi_{\epsilon} * \mu)\mu\right)$, $F_2 = \left(F^\prime\circ(\varphi_{\epsilon} * \mu)\right) \nabla\varphi_{\epsilon} * \mu$. We use particles to approximate $\mu$, {\it e.g.}, $\mu \approx \frac{1}{M}\sum_{i=1}^{M}\delta_{\thetab^{(i)}i}$. We have
\begin{align}\label{eq:gradF1}
F_1(\thetab) &= \left(\nabla K * \frac{\mu}{K * \mu}\right)(\thetab) \nonumber\\
&\approx \left(\nabla K * \left(\frac{1}{M}\sum_{i=1}^M\frac{\delta_{\thetab^{(i)}}}{(K * (\frac{1}{M}\sum_j\delta_{\thetab^{(j)}}))(\thetab^{(i)})}\right)\right)(\thetab) \nonumber\\
&= \left(\nabla K * \left(\sum_{i=1}^M\frac{\delta_{\thetab^{(i)}}}{\sum_j K(\thetab^{(i)}-\thetab^{(j)})}\right)\right)(\thetab) \nonumber\\
&= \sum_{i=1}^M\frac{\nabla K(\thetab - \thetab^{(i)})}{\sum_j K(\thetab^{(i)} - \thetab^{(j)})}~.
\end{align}

\begin{align}\label{eq:gradF2}
F_2(\thetab) &= \left(\frac{\nabla K * \mu}{K * \mu}\right)(\thetab) \nonumber\\
&\approx \left(\frac{\nabla K * \frac{1}{M}\sum_{i=1}^{M}\delta_{\thetab^{(i)}}}{K * \frac{1}{M}\sum_{i=1}^{M}\delta_{\thetab^{(i)}}}\right)(\thetab) \nonumber\\
&= \frac{\sum_{i=1}^M\nabla K(\thetab - \thetab^{(i)})}{\sum_{i=1}^M K(\thetab - \thetab^{(i)})}~.
\end{align}
Combing \eqref{eq:gradF1} and \eqref{eq:gradF2} gives the formula for $\vb$ in \eqref{eq:velocity}.

\begin{table}[htp]\centering \hspace{-0mm}
	\caption{\small Hyper-parameter settings for MNIST on FNN.}
	\vspace{-3mm}
	\label{tab:mnist_para}
	\vskip 0.0in
	\centering
	\small
	\hspace{ 0mm} 	
	\begin{adjustbox}{scale=1.0,tabular=l|cc,center}
		\hline
		{\bf Datasets} &\multicolumn{2}{c}{$\mathtt{MNIST}$}\\
		\hline
		Batch Size  	  &   100  &   100 \\
		Step Size &  $5\!\times10^{-4}$ &  $5\!\times10^{-4}$ \\	
		\#Epoch &$150$ & $150$\\
		RMSProp &0.99 & 0.99\\
		\hline
		Network (hidden layers) & [400, 400]& [800, 800] 	\\					
		\hline
		Variance in prior & 1 & 1   \\
		\hline
	\end{adjustbox}
	\vspace{2mm}
\end{table}

\section{Experimental Setting}
We list some experimental settings in Table~\ref{tab:mnist_para} and Table~\ref{tab:cifar10_para}. For evaluation on BNNs, following a standard Bayesian treatment, we use ensemble of particle predictions to compute the test accuracy. We will need to store all the $M$ particles in order to do particle optimization, thus the time and memory complexity would be proportional to the number of particles. In practice, however, we can reduce the complexity by only treating a small part of the parameters as particles ({\it e.g.}, the parameters of the last layer of a BNN), and leaving others as single values.

\begin{table}[htp]\centering \hspace{-0mm}
	\caption{\small Hyper-parameter settings for CIFAR-10 on CNN.}
	\vspace{-3mm}
	\label{tab:cifar10_para}
	\vskip 0.0in
	\centering
	\small
	\hspace{ 0mm} 	
	\begin{adjustbox}{scale=1.0,tabular=l|c,center}
		\hline
		{\bf Datasets} &{$\mathtt{CIFAR10}$}\\
		\hline
		Batch size  	  &   128  \\
		Step size &    0.01 ($<5e^3$), 0.001 ($<1e^4$)\, 0.0001\\	
		\#Epoch & $200$ \\
		\hline
		Filter size & 3$\times$3	\\	
		Channels & C64-C128-C256\\		
		Network (hidden layers) & [1024]	\\		
		\hline
		Variance in prior & 1  \\
		\hline
	\end{adjustbox}
	\vspace{2mm}
\end{table}

\section{Extra Experiments}\label{app:ex_exp}
\begin{figure*}[h!] \centering
	\begin{tabular}{ccccc}
		\hspace{-4mm}
		\includegraphics[width=4cm]{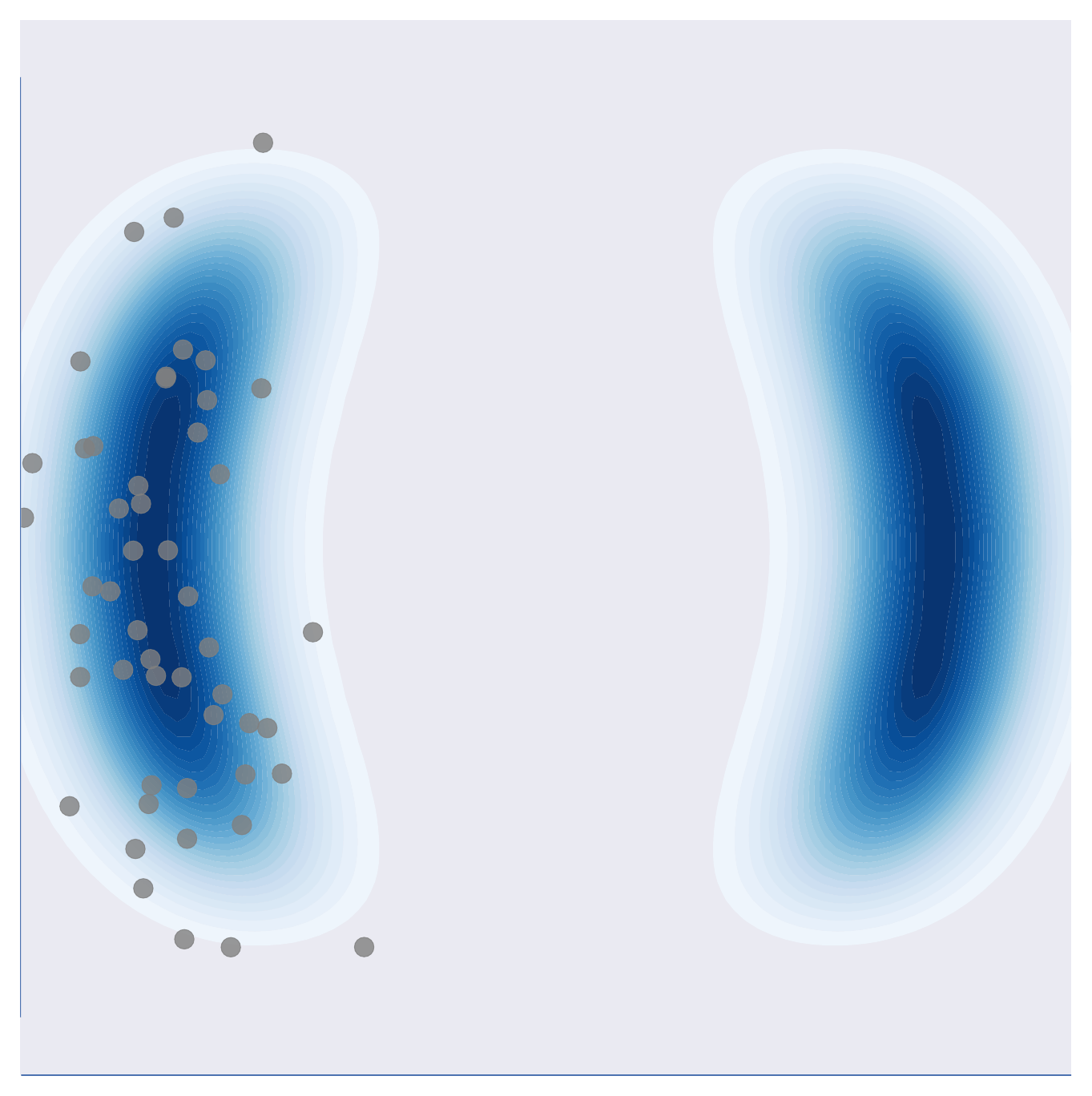}  
		&   \hspace{-5mm}
		\includegraphics[width=4cm]{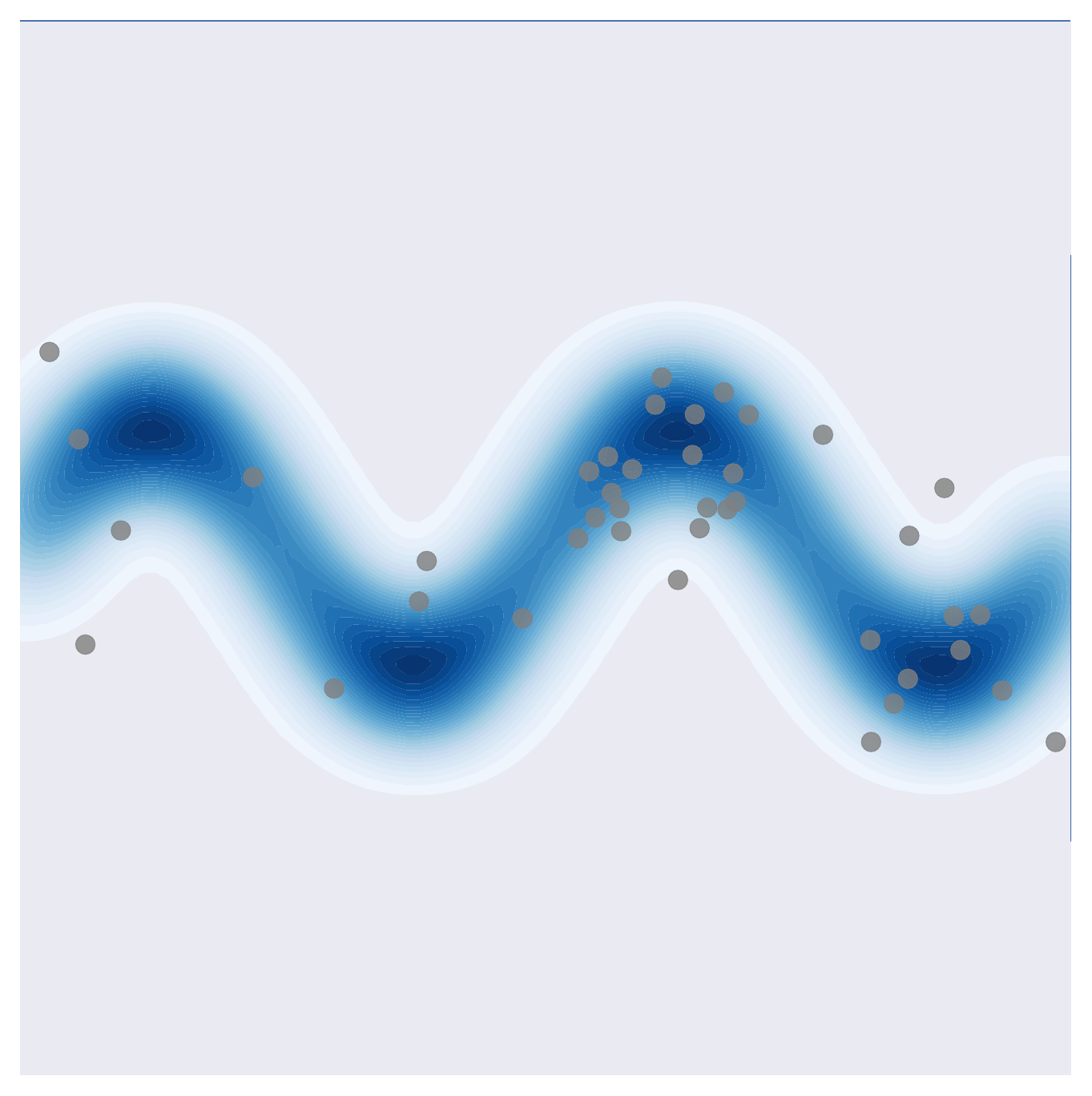} 
		& 		\hspace{-5mm}
		\includegraphics[width=4cm]{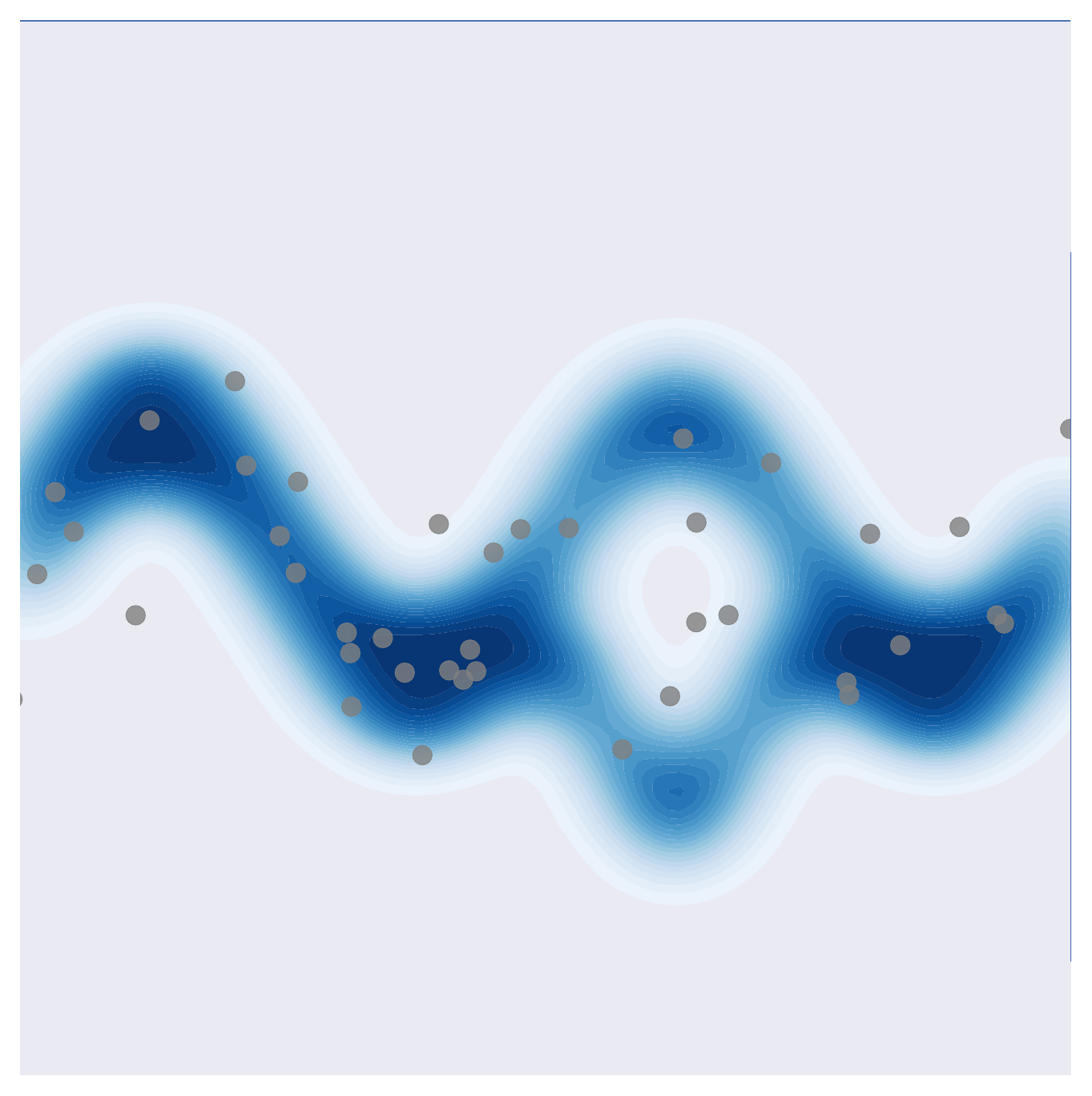}   
		&		\hspace{-5mm}
		\includegraphics[width=4cm]{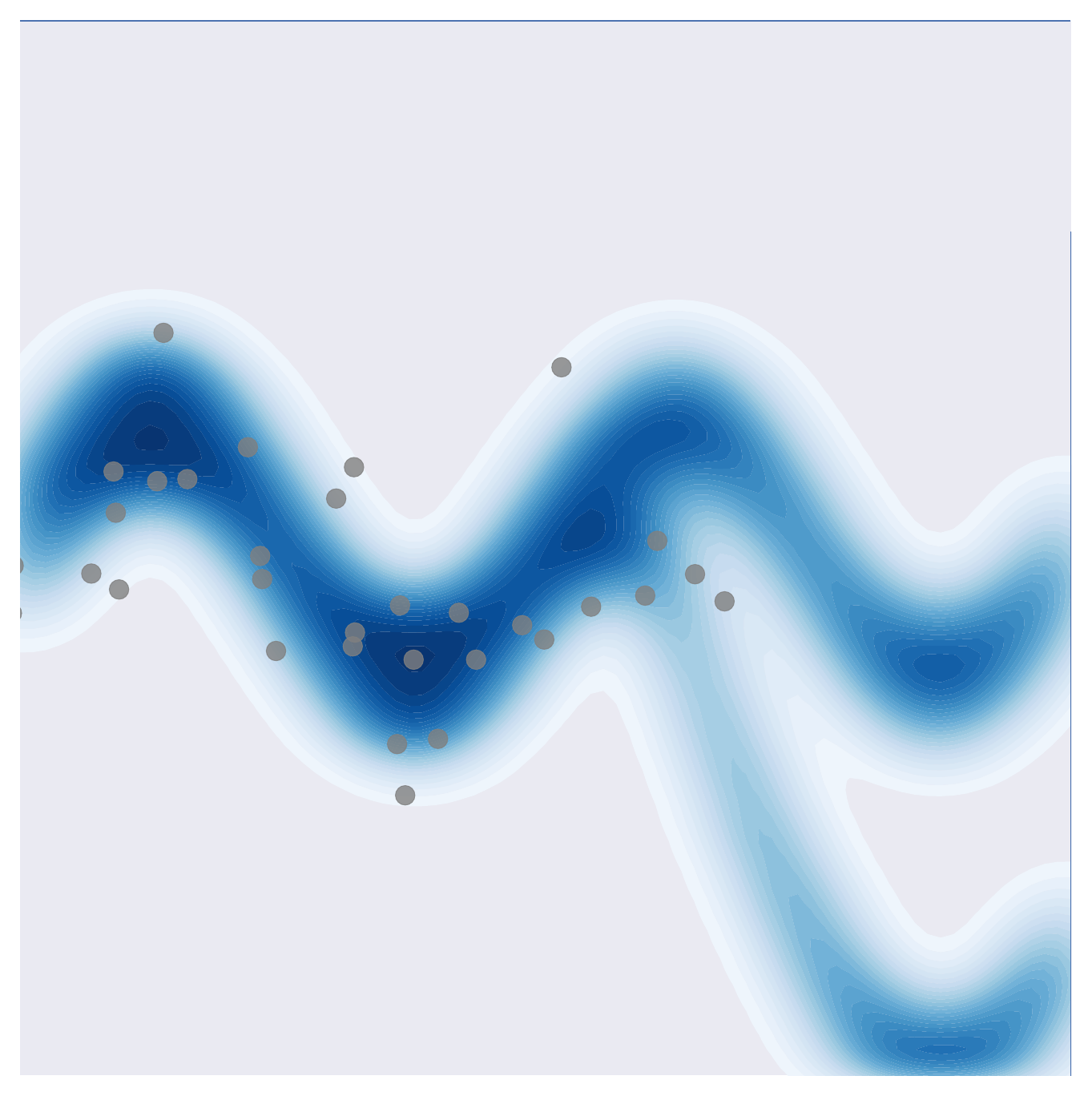}
		\vspace{-1mm}
		\\
		\hspace{-4mm}
		\includegraphics[width=4cm]{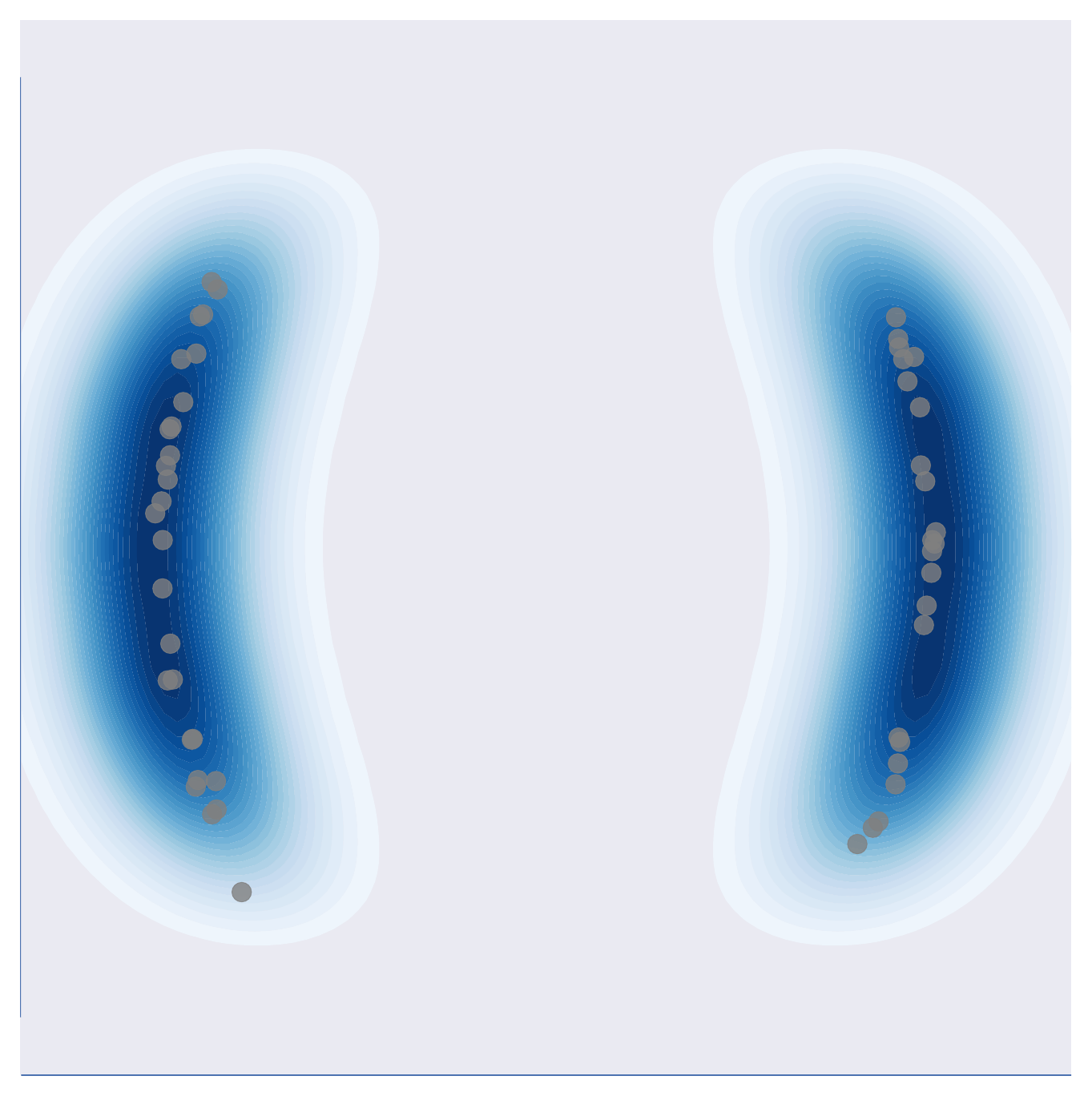}  
		&   \hspace{-5mm}
		\includegraphics[width=4cm]{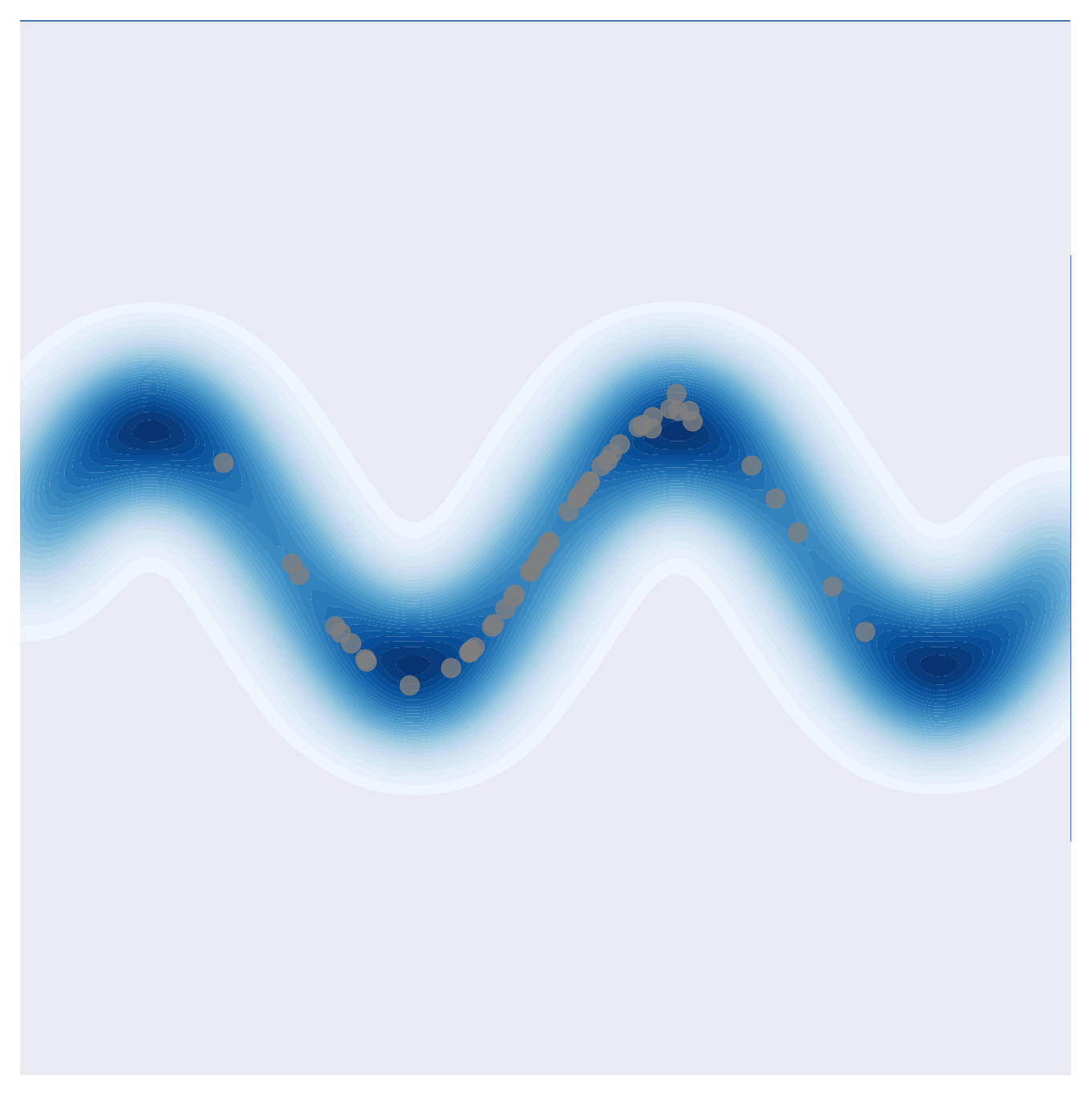} 
		& 		\hspace{-5mm}
		\includegraphics[width=4cm]{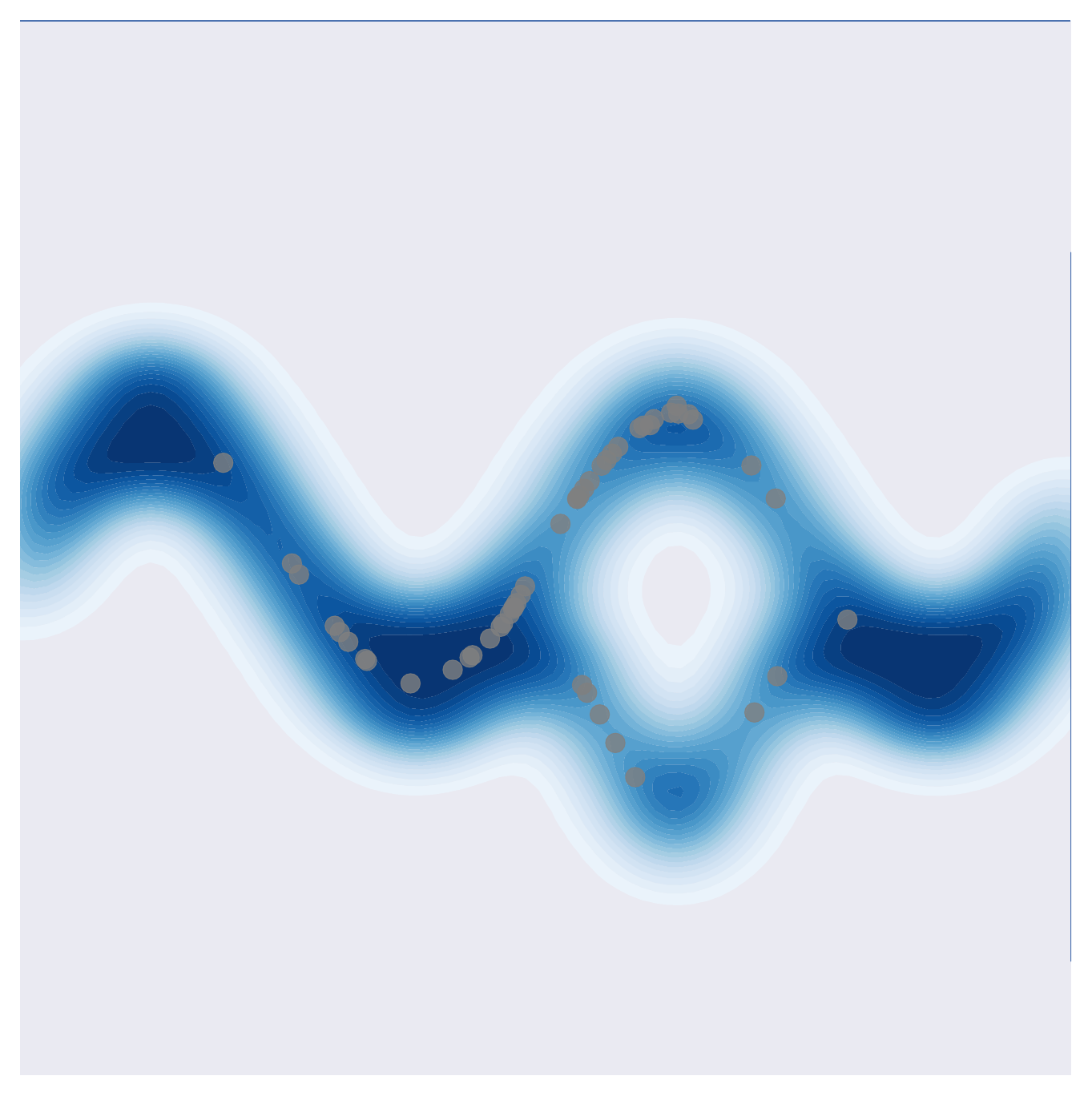}   
		&		\hspace{-5mm}
		\includegraphics[width=4cm]{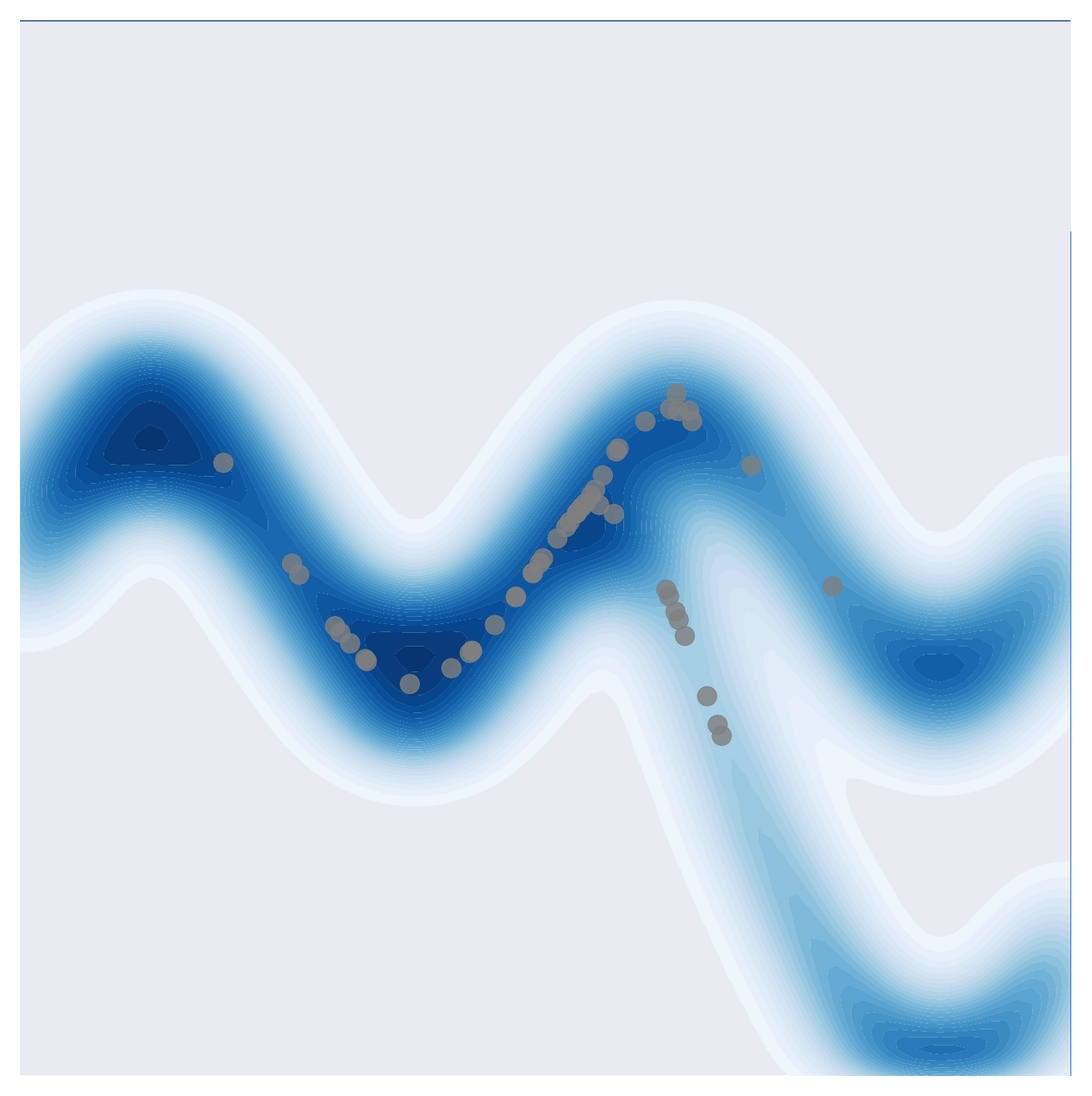}
		\vspace{-1mm}
		\\
		\hspace{-4mm}
		\includegraphics[width=4cm]{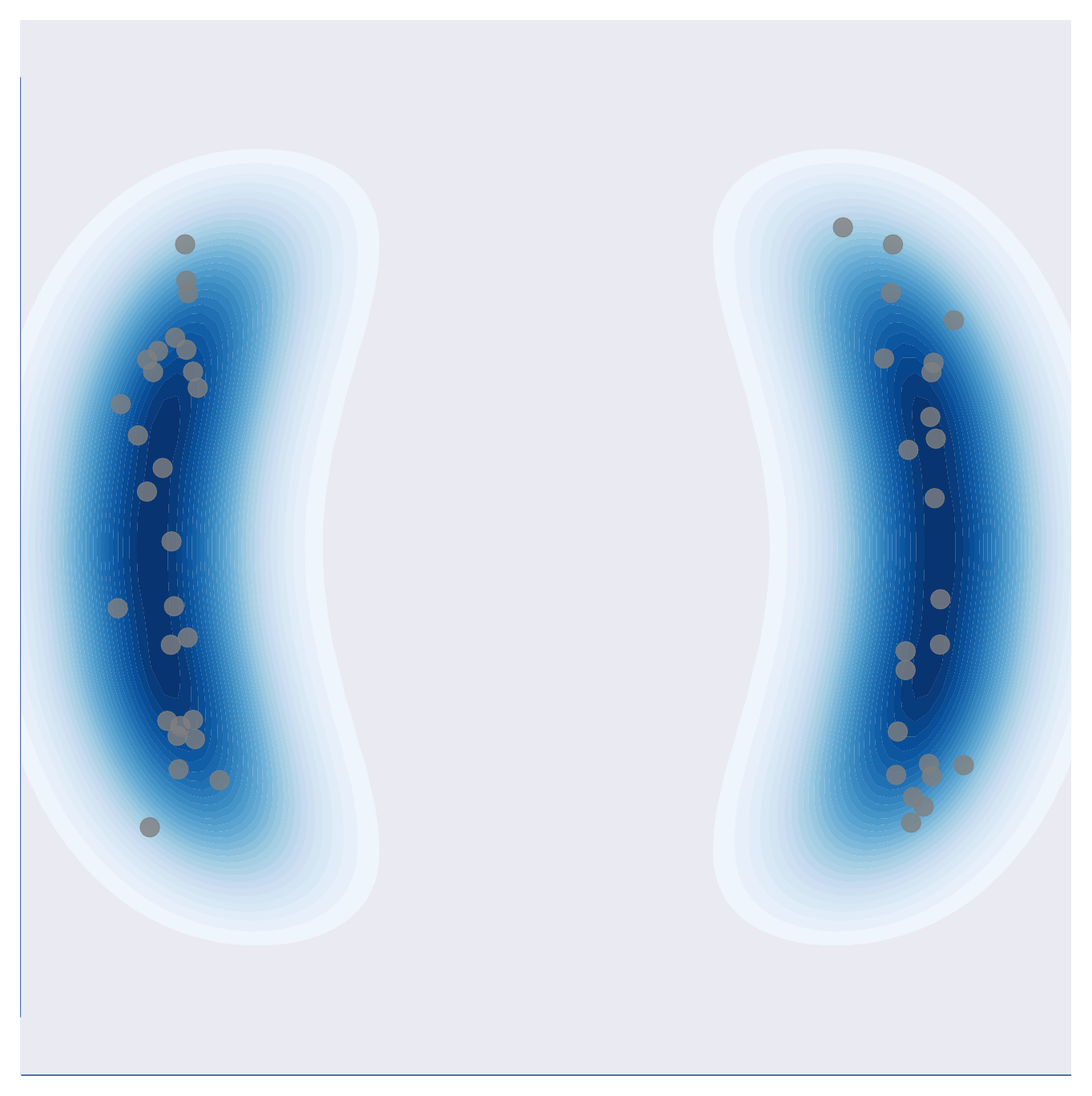}  
		&   \hspace{-5mm}
		\includegraphics[width=4cm]{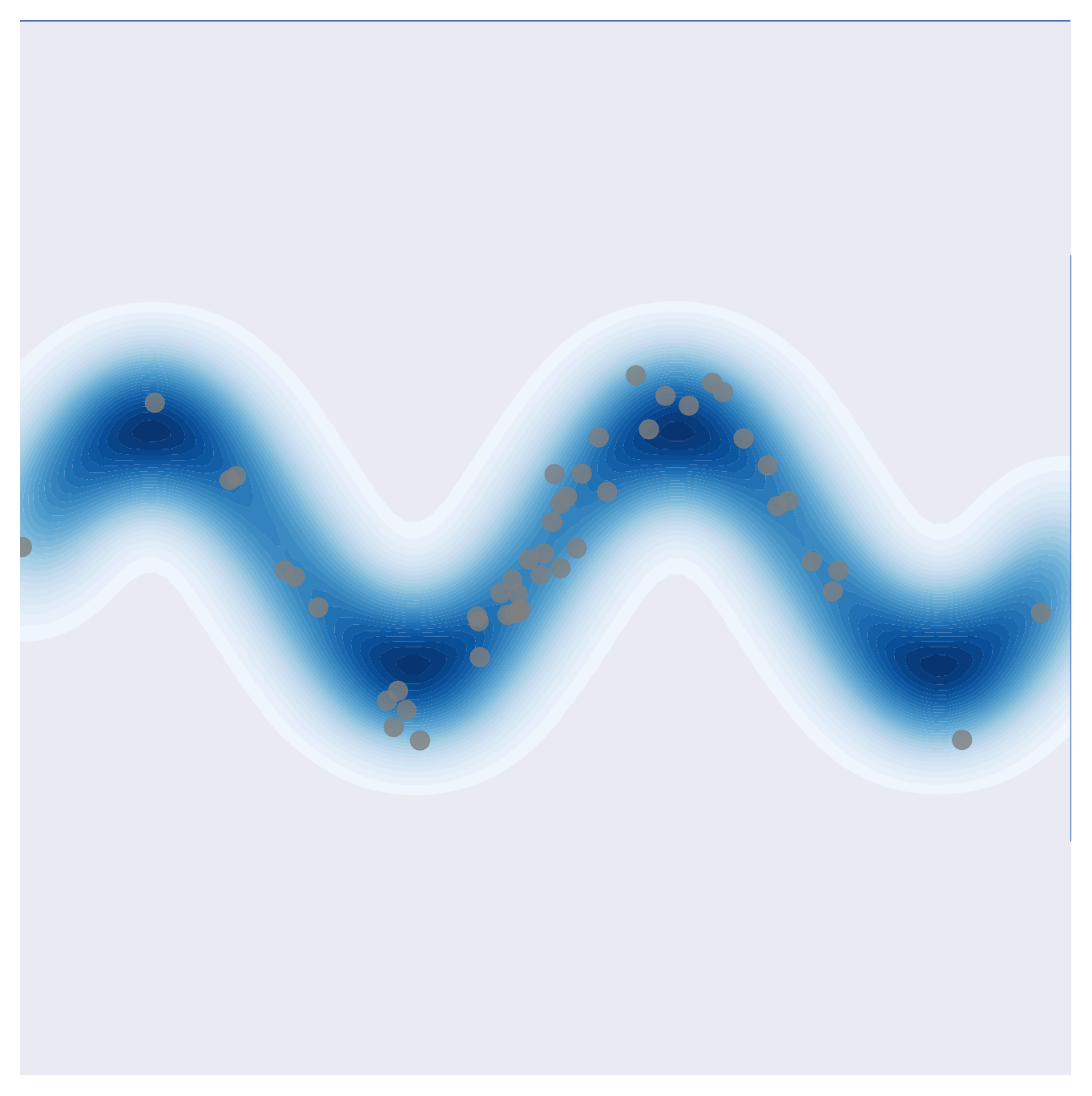} 
		& 		\hspace{-5mm}
		\includegraphics[width=4cm]{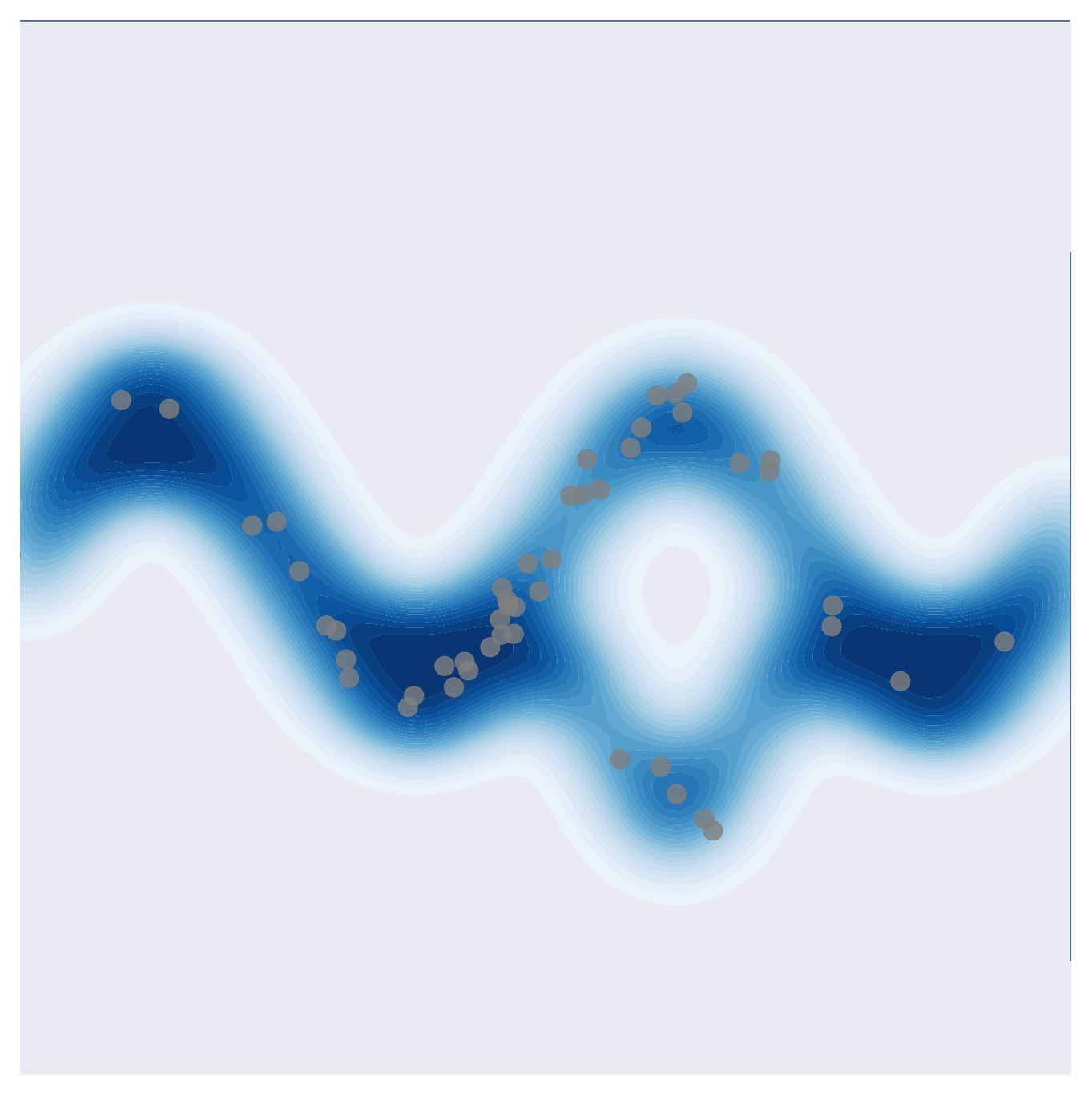}   
		&		\hspace{-5mm}
		\includegraphics[width=4cm]{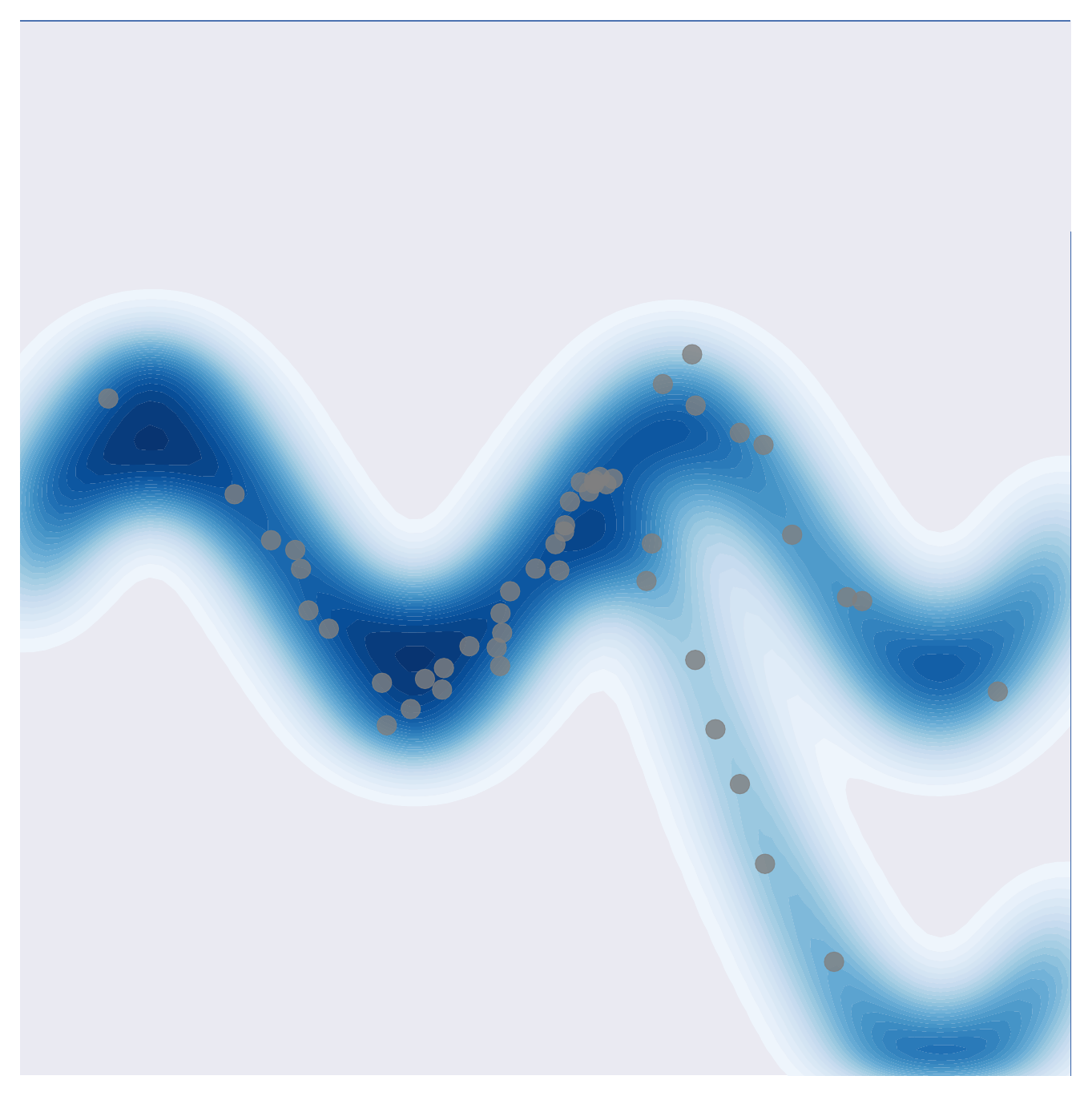}
		\vspace{-1mm}
		\\
		\hspace{-4mm}
		\includegraphics[width=4cm]{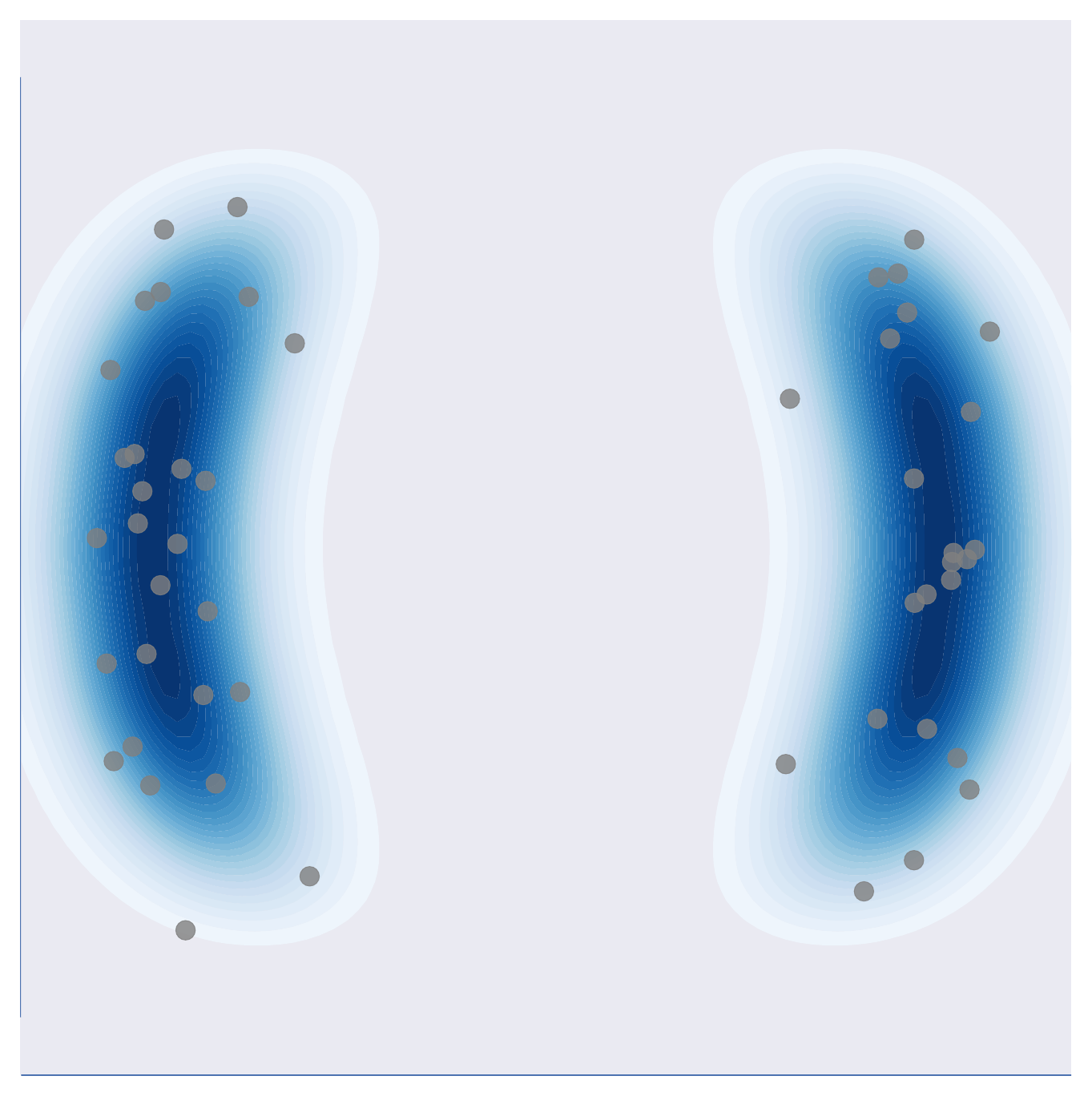}  
		&   \hspace{-5mm}
		\includegraphics[width=4cm]{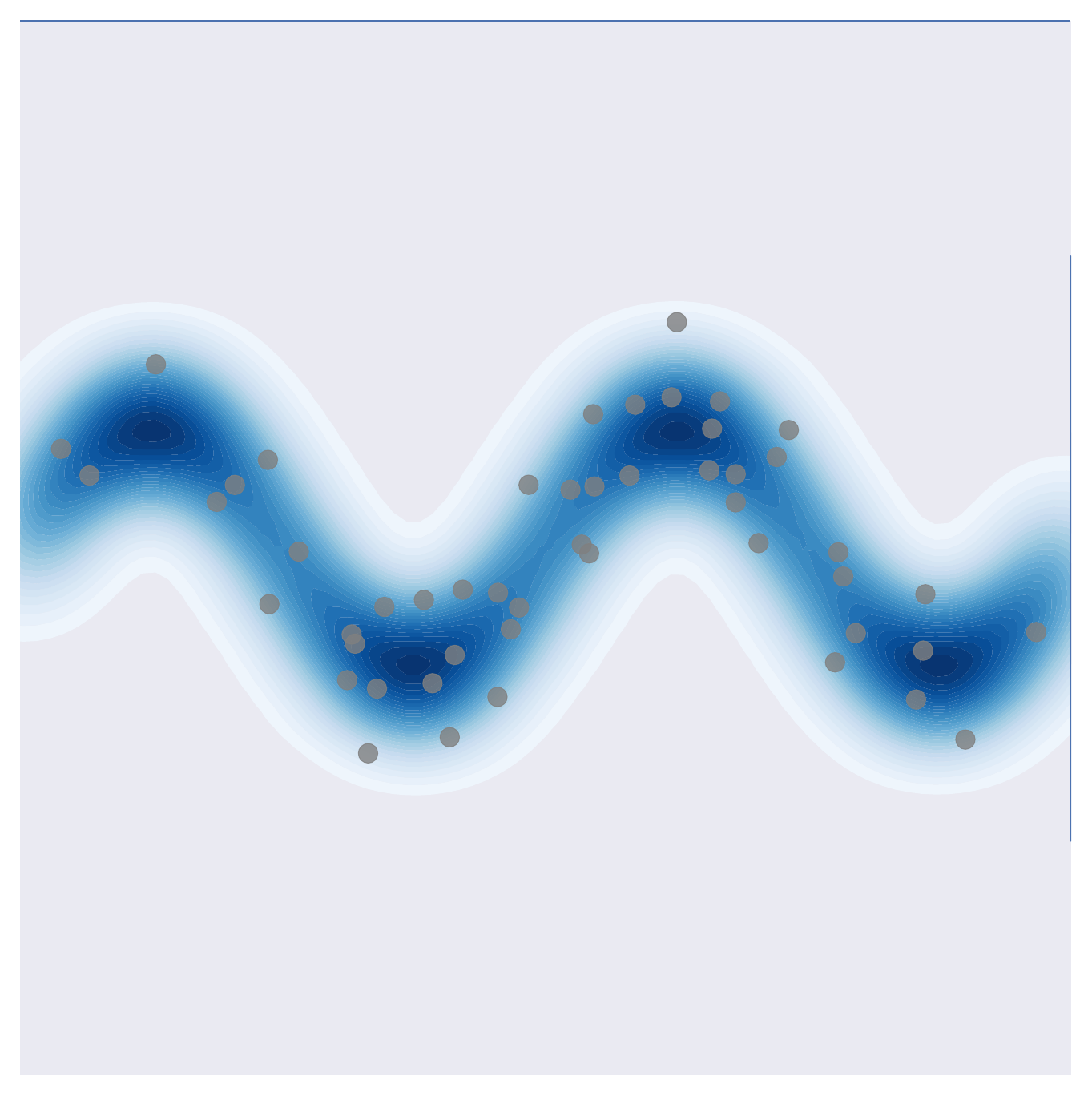} 
		& 		\hspace{-5mm}
		\includegraphics[width=4cm]{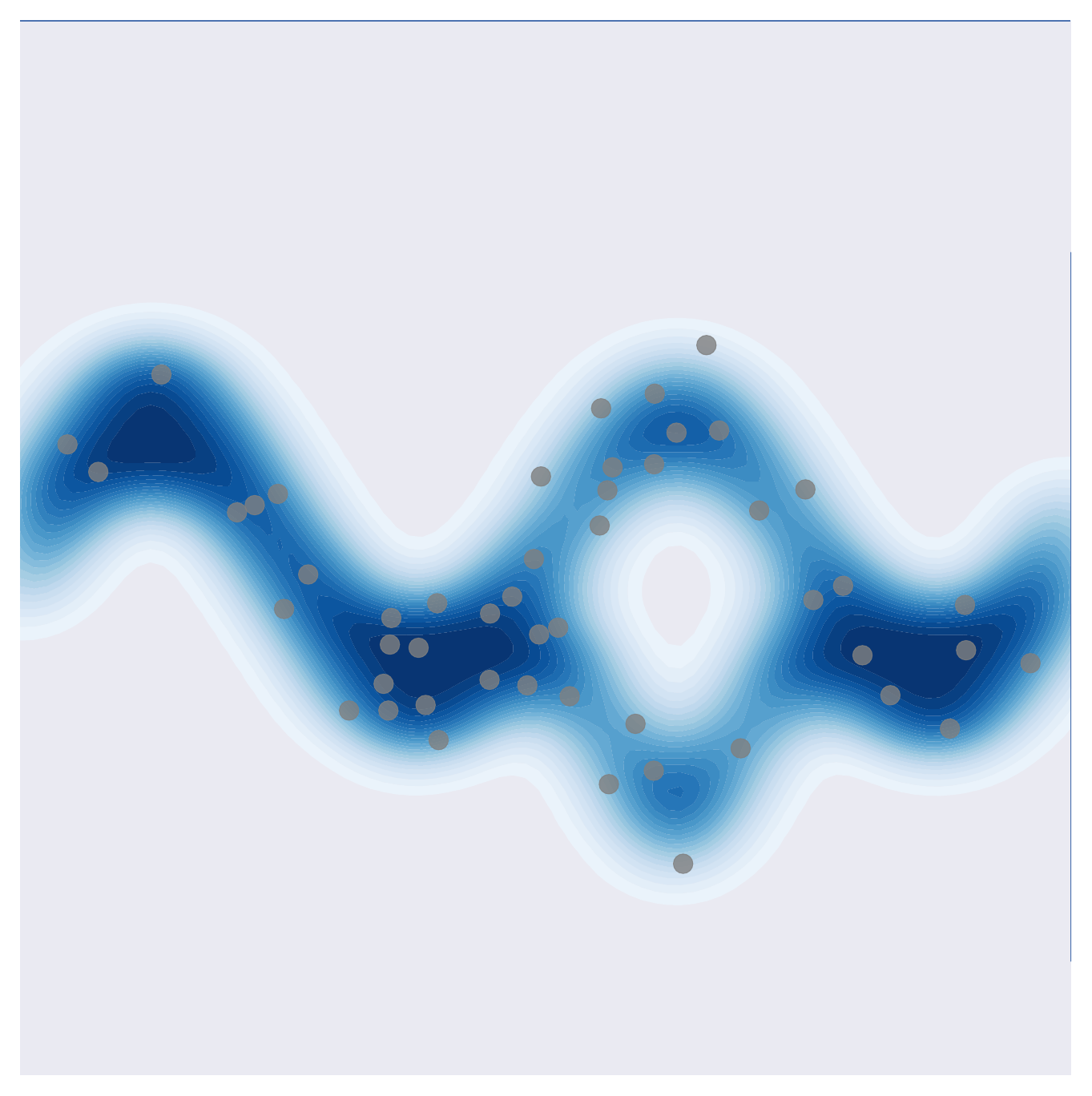}   
		&		\hspace{-5mm}
		\includegraphics[width=4cm]{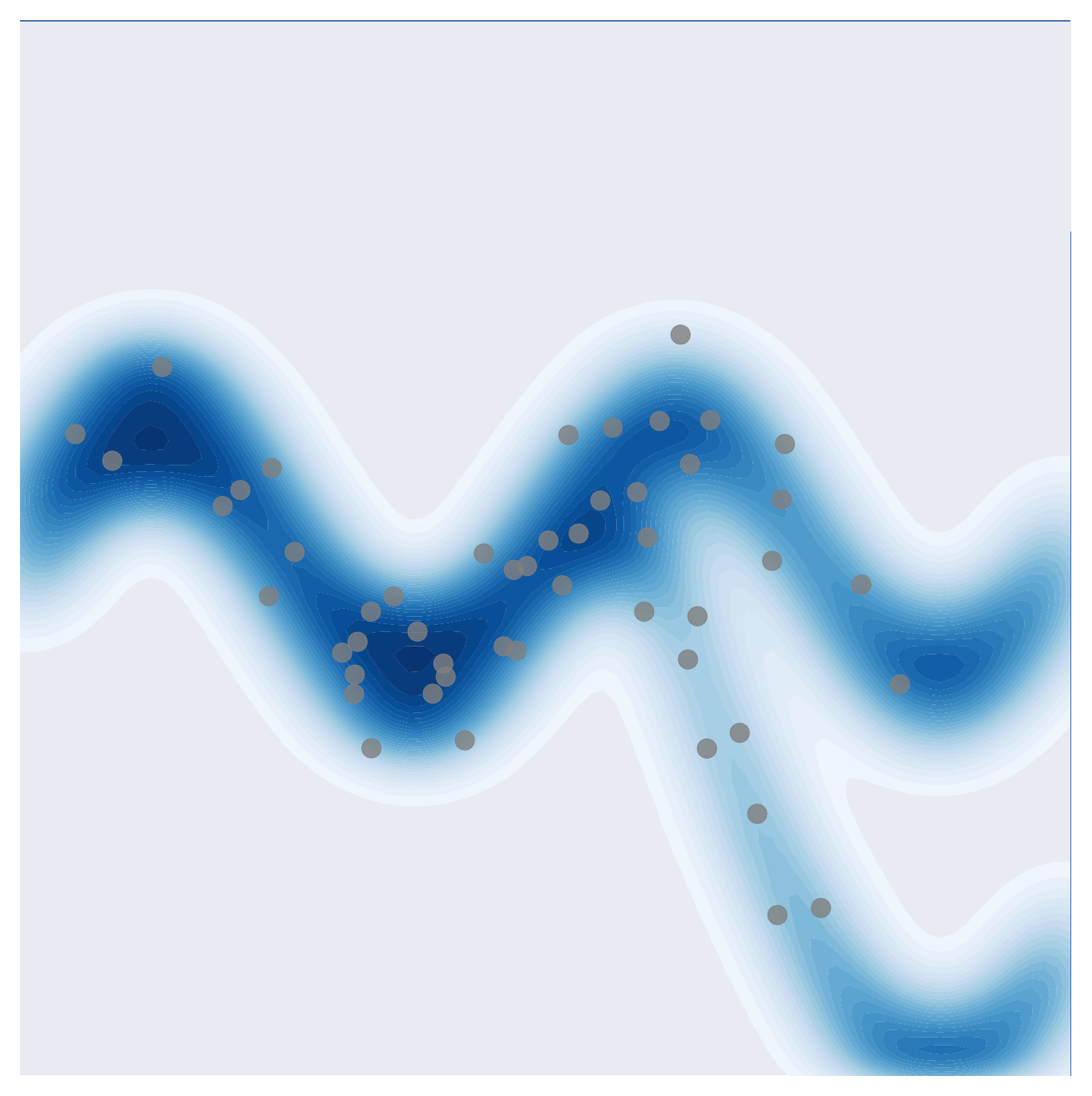}
		\vspace{-1mm}
		\\
		\hspace{-4mm}
		\includegraphics[width=4cm]{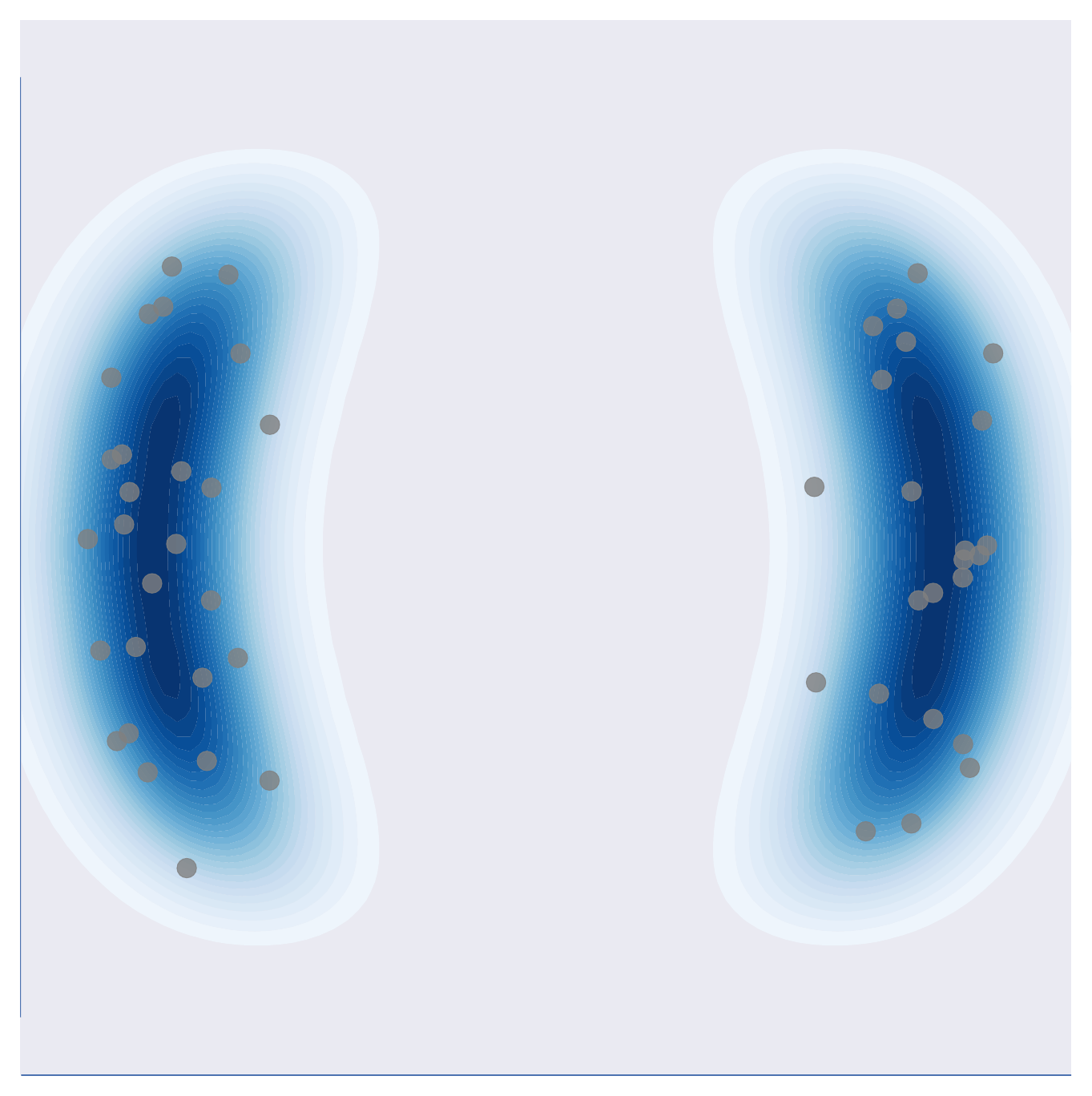}  
		&   \hspace{-5mm}
		\includegraphics[width=4cm]{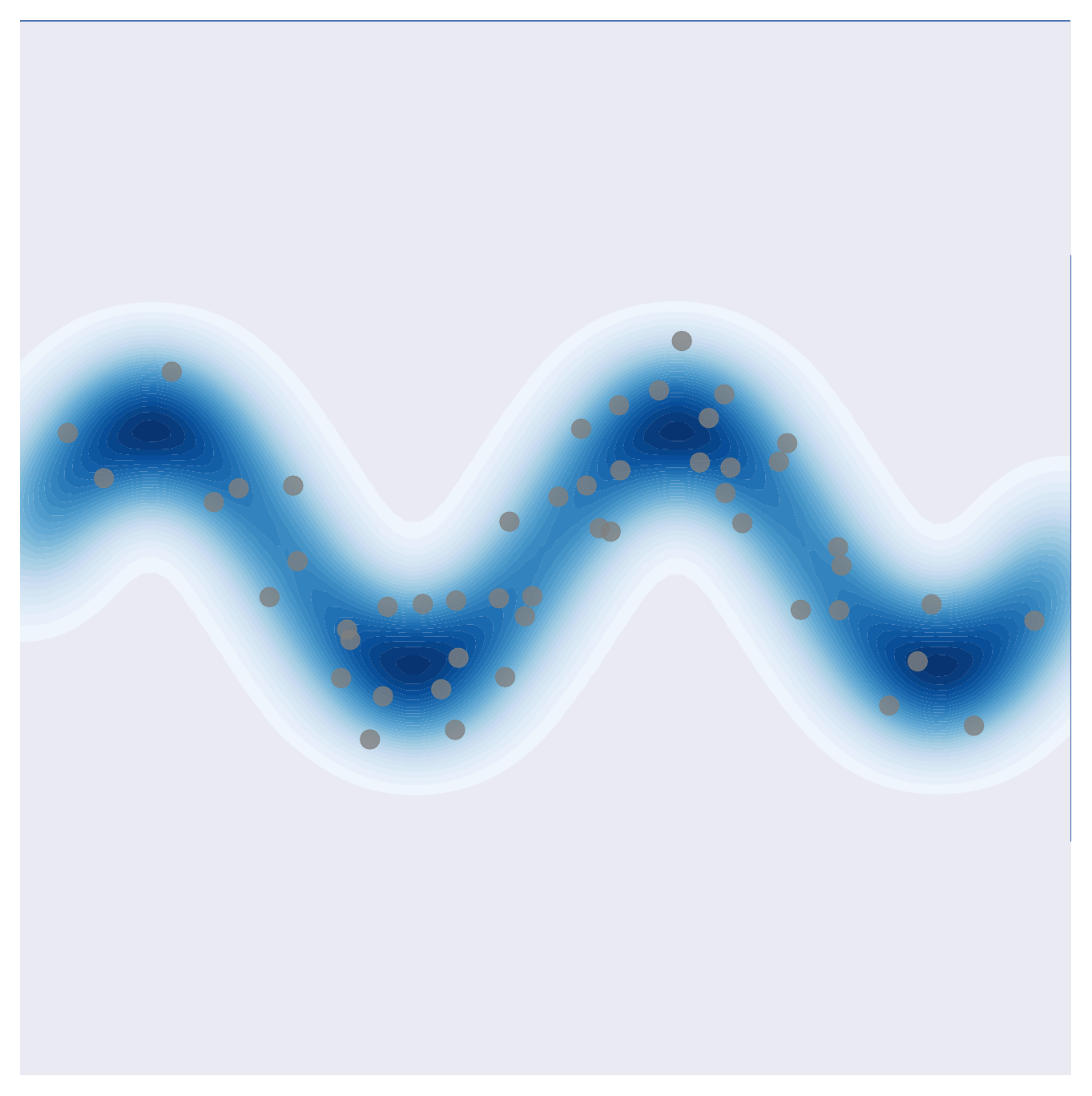} 
		& 		\hspace{-5mm}
		\includegraphics[width=4cm]{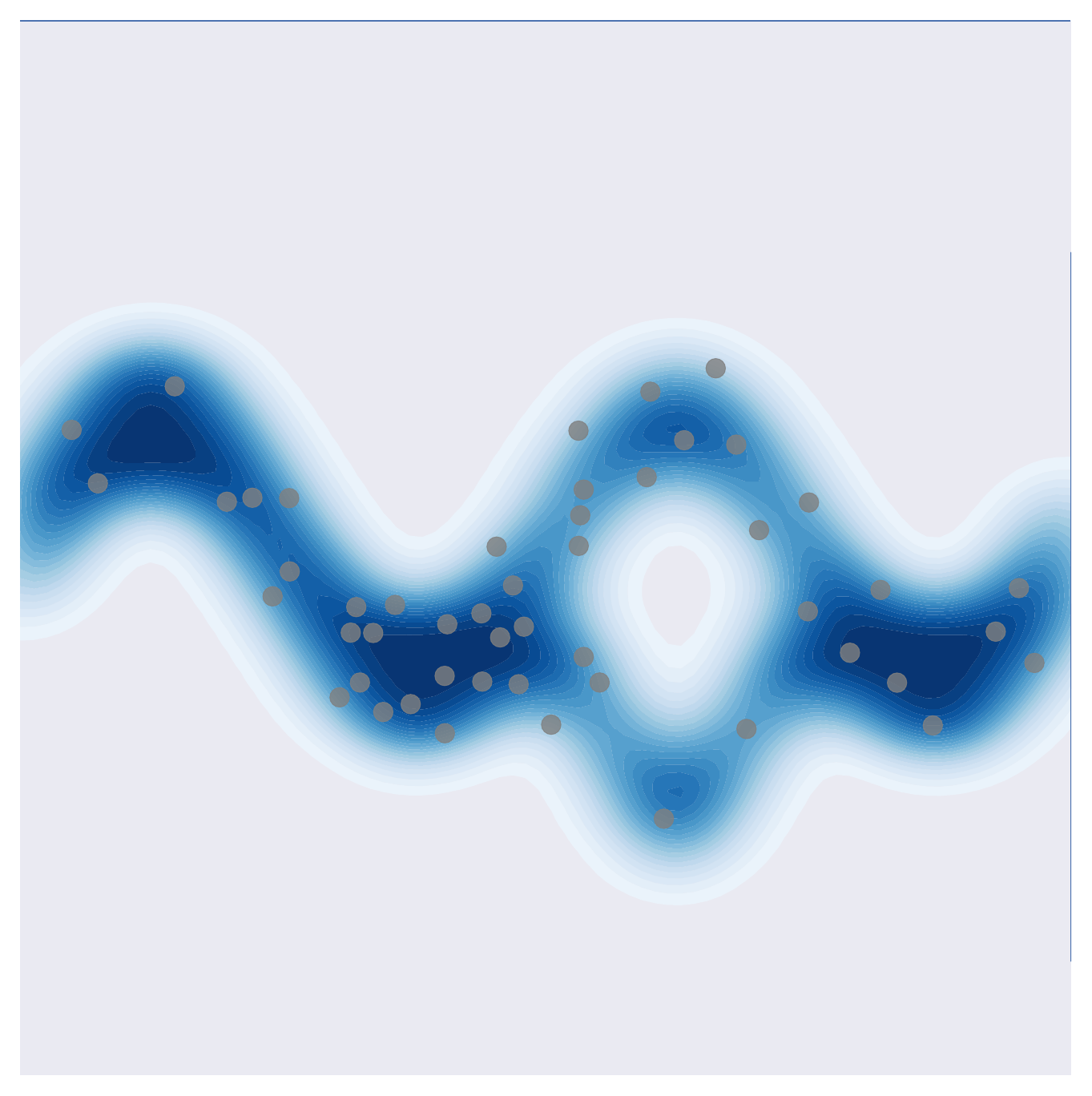}   
		&		\hspace{-5mm}
		\includegraphics[width=4cm]{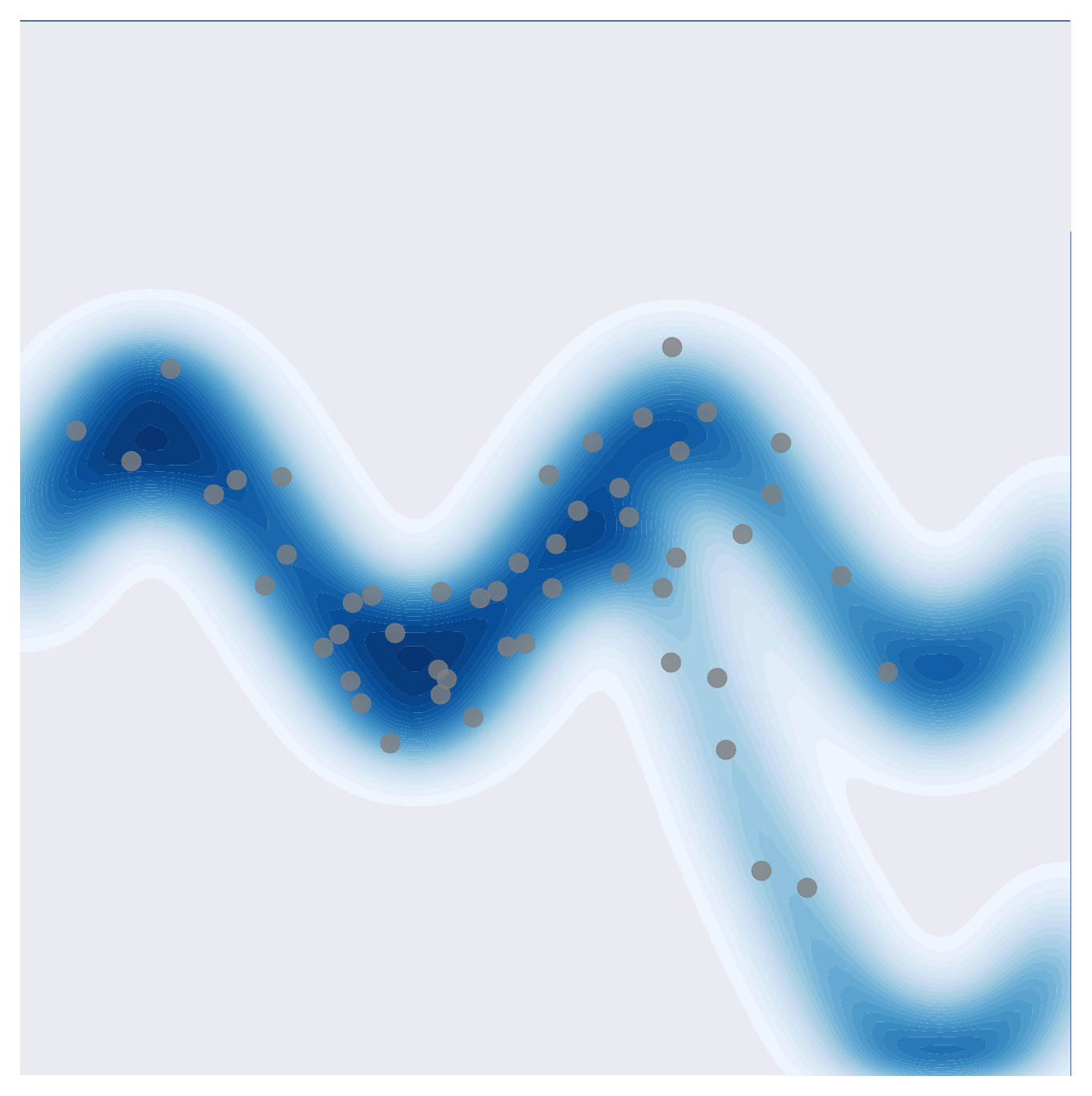} 
		\vspace{-1mm}
		\\
	\end{tabular} 
	\caption{{\small Illustration of different algorithms on toy distributions with 50 particles. Each column is a distribution case. 1st row: standard SGLD; 2nd row: $w$-SGLD; 3rd row: $w$-SGLD-B; 4th row: SVGD; 5th row: $\pi$-SGLD}. The blue shapes are ground true density contours.}
	\label{fig:50toy}
	\vspace{-4mm}
\end{figure*}

We further optimize 50 particles to approximate different distributions. The optimized particles are plotted in Figure~\ref{fig:50toy}, which shows that $w$-SGLD seems to be able to learn better particles due to the concentration property of the Wasserstein regularization term.

\end{document}